\documentclass{article} 
\usepackage[final]{neurips_2022}
 
\usepackage[utf8]{inputenc} 
\usepackage[T1]{fontenc}    
\usepackage{hyperref}       

\hypersetup{
    colorlinks=true,
    linkcolor=red,
    citecolor=cyan,
    filecolor=magenta,      
    urlcolor=cyan,
    }

\usepackage{url}            
\usepackage{booktabs}       
\usepackage{amsfonts}       
\usepackage{nicefrac}       
\usepackage{microtype}      
\usepackage{xcolor}   

\usepackage{times}
\usepackage{epsfig} 
\usepackage{amssymb}
\usepackage{enumerate}  
\usepackage{amsthm}
\usepackage{color}
\usepackage{algpseudocode} 
\usepackage{pifont}
\newcommand{\cmark}{\ding{51}}%
\newcommand{\xmark}{\ding{55}}%

\DeclareMathAlphabet\mathbfcal{OMS}{cmsy}{b}{n}

\def\0{{\bf 0}}
\def\1{{\bf 1}}

\newtheorem{thm}{Theorem}

\def\eg{\emph{e.g.}} 
\def\ie{\emph{i.e.}}

\usepackage{tabu}  

\usepackage{capt-of} 
\usepackage{wrapfig} 
\usepackage{graphicx}
\usepackage{subfigure} 
\usepackage{bm}
\usepackage{comment}
\usepackage{amsfonts}
\usepackage{amsmath}
\usepackage{balance} 
\usepackage{listings} 
\usepackage{xcolor} 
\usepackage{arydshln}
\usepackage{cases}  

\usepackage{algorithm}
\usepackage{multicol,multirow}    
\usepackage{threeparttable} 
\def\ie{\emph{i.e., }}
\def\eg{\emph{e.g., }}

\def\red{\textcolor{red}}

\newcommand{\return}{\textbf{Output:}}
 
\usepackage[capitalize]{cleveref}
\crefname{section}{Sec.}{Secs.}
\Crefname{section}{Section}{Sections}
\Crefname{table}{Table}{Tables}
\crefname{table}{Tab.}{Tabs.}

\title{Self-Supervised Aggregation of Diverse Experts  for Test-Agnostic Long-Tailed Recognition}

\author{%
  Yifan Zhang$^1$ ~~~  Bryan Hooi$^1$ ~~ Lanqing Hong$^2$  ~~ Jiashi Feng$^{3}$ \\ 
  $^1$National University of Singapore ~~~~ $^2$Huawei Noah’s Ark Lab ~~~~  $^3$ByteDance \\
  \texttt{yifan.zhang@u.nus.edu, jshfeng@gmail.com} 
}

\begin{document}

\maketitle 

\begin{abstract} 
Existing long-tailed recognition methods, aiming to train class-balanced models from long-tailed data, generally assume the models would be evaluated on the uniform test class distribution. However, practical test class distributions often violate this assumption (\eg being either long-tailed or even inversely long-tailed), which may lead existing methods to fail in real applications. In this paper, we study a more practical yet challenging task, called \emph{test-agnostic long-tailed recognition}, where the training class distribution is long-tailed while the test class distribution is \emph{agnostic and not necessarily uniform}. In addition to the issue of class imbalance, this task poses another challenge: the class distribution shift between the training and test data is unknown. To tackle this task, we propose a novel approach, called \emph{Self-supervised Aggregation of Diverse Experts}, which consists of two strategies: (i) a new skill-diverse expert learning strategy that trains multiple experts from a single and stationary long-tailed dataset to separately handle different class distributions; (ii) a novel test-time expert aggregation strategy that leverages self-supervision to aggregate the learned multiple experts for handling unknown test class distributions. We theoretically show that our self-supervised strategy has a provable ability to simulate test-agnostic class distributions. Promising empirical results demonstrate the effectiveness of our method on both vanilla and test-agnostic long-tailed recognition. Code is available at  \url{https://github.com/Vanint/SADE-AgnosticLT}.

\end{abstract}     
  
\section{Introduction}  
Real-world  visual recognition datasets typically exhibit a long-tailed distribution, where a few classes contain numerous samples (called head classes), but the others are associated with only a few instances (called tail classes)~\cite{kang2021exploring,menon2020long}. Due to  the class imbalance,   the trained  model is easily biased towards   head   classes and perform poorly on tail  classes~\cite{cai2021ace,zhang2021deep}. To tackle this issue, numerous   studies have   explored  long-tailed recognition for learning well-performing models from imbalanced data~\cite{jamal2020rethinking,zhang2021distribution}.

\begin{figure*}[t]\vspace{-0.15in} 
 \begin{minipage}{0.46\linewidth}
  \centering 
   \hspace{-0.15in} \includegraphics[width=6.2cm]{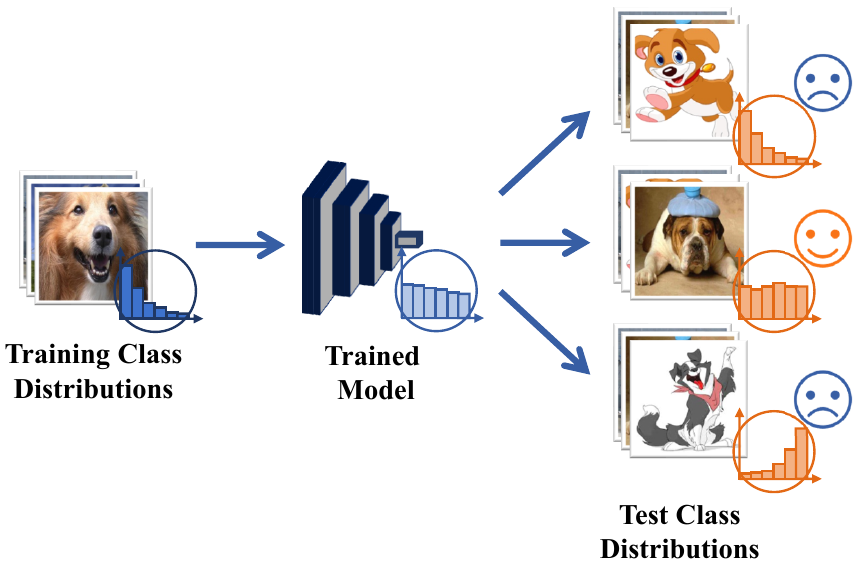} 
  
   {\vspace{-0.05in}\hspace{-0.15in} \centerline{\small{(a) Existing long-tailed recognition methods}}}
  \end{minipage}
   \hfill  
  \begin{minipage}{0.51\linewidth}
    \centering
   \hspace{-0.15in}\includegraphics[width=7.2cm]{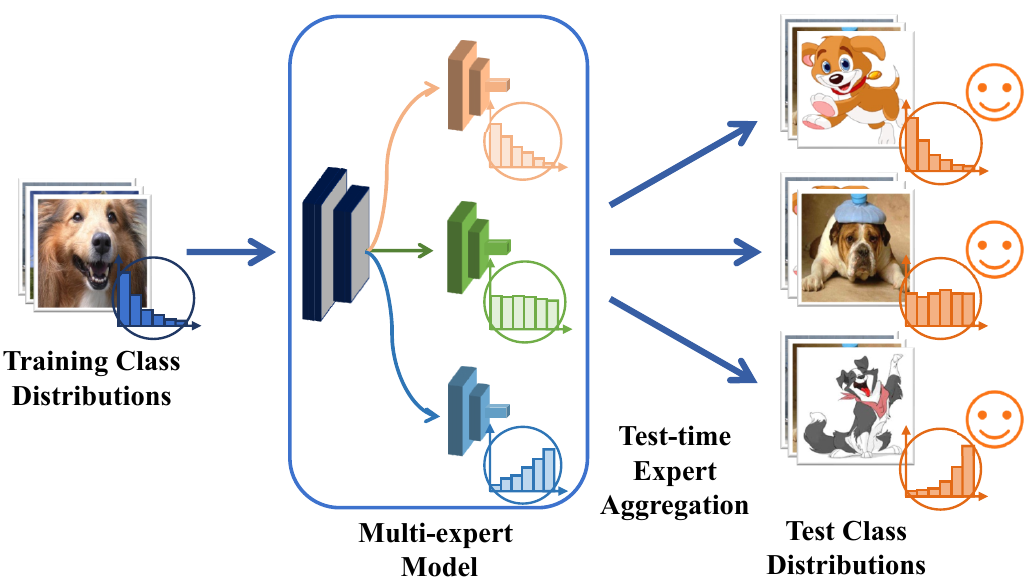}
   
   {\vspace{-0.05in} \hspace{-0.15in}\centerline{\small{(b) The idea of our proposed method}}}
    \end{minipage}
     
  \caption{Illustration of test-agnostic long-tailed recognition. (a) Existing  long-tailed  recognition  methods  aim  to train   models  that perform well on test data with the uniform class distribution. However, the resulting models may fail to handle practical test class distributions that skew arbitrarily. (b) Our method seeks to learn a   multi-expert model from a single long-tailed training set, where  different experts are skilled  in handling different   class distributions, respectively. By reasonably aggregating    these   experts at test time, our method is able to handle       unknown test class distributions.}  
  \label{ideas}\vspace{-0.15in} 
\end{figure*}

Most existing long-tailed   studies~\cite{cao2019learning,cui2019class,deng2021pml,wang2021contrastive,weng2021unsupervised} assume   the test class distribution is uniform, \ie each class  has an equal amount of test data. Therefore, they develop various techniques, \eg class re-sampling~\cite{guo2021long,huang2016learning,kang2019decoupling,zang2021fasa}, cost-sensitive learning~\cite{feng2021exploring,Influence2021Park,tan2020equalization,wang2021seesaw} or ensemble learning~\cite{cai2021ace,guo2021long,li2020overcoming,xiang2020learning},  to re-balance the  model  performance   on different  classes for fitting the uniform class  distribution. However, this assumption does not always hold in real applications, where   actual   test data may follow any kind of class distribution, being either uniform, long-tailed, or even inversely long-tailed to the training data (cf. Figure~\ref{ideas}(a)). For example, one may train a  recognition model for autonomous cars based on the training data collected from city areas, where pedestrians are majority classes and stone obstacles are minority classes. However,  when the model is deployed to mountain areas, the pedestrians   become the minority while the stones become the majority.
In this case, the test class distribution is inverse to the training one, and existing    methods may perform poorly.

To address the issue of varying class distributions, as the first research attempt,  LADE~\cite{hong2020disentangling}  assumes   the test class distribution to be known  and uses the knowledge to post-adjust model predictions. However, the actual test class distribution is usually unknown  a priori, making LADE not applicable in practice.  Therefore, we    study   a more realistic yet challenging    problem,  namely \textit{test-agnostic long-tailed recognition}, where the training class distribution is long-tailed  while the test  distribution is  \emph{agnostic}. 
To tackle this problem,  motivated by the idea of "divide and conquer",  we propose to learn multiple experts with \emph{diverse} skills that excel at handling different class distributions (cf.~Figure~\ref{ideas}(b)). As long as these skill-diverse experts can be aggregated suitably at test time, the  multi-expert model would manage to handle the  unknown test class distribution.  
Following this idea, we develop a novel approach, namely \emph{Self-supervised Aggregation of Diverse Experts} (SADE).

The first challenge for SADE  is how to learn multiple \emph{diverse} experts from a  \emph{single} and  \emph{stationary} long-tailed training dataset.
To handle this challenge,  we   empirically evaluate existing long-tailed methods in this task, and find     that  the  models trained by existing  methods have a \emph{simulation correlation between the learned class distribution and the training loss function}. That is, the models learned by various losses are skilled in handling class distributions with different skewness. For example, the model trained with the conventional softmax loss simulates the long-tailed training class distribution, while the models obtained from existing  long-tailed methods are good at the uniform class distribution. Inspired by this finding, SADE presents a simple but effective skill-diverse expert learning strategy  to generate     experts  with different distribution preferences from   a single long-tailed training distribution. Here, various experts are trained with different   expertise-guided objective functions  to  deal with different  class distributions, respectively. 
As a result, the learned  experts are  more diverse than  previous multi-expert long-tailed methods~\cite{wang2020long,zhou2020bbn}, leading to  better ensemble performance, and  in aggregate   simulate a wide spectrum of possible  class distributions.

The  other challenge  is how to aggregate these   skill-diverse experts for handling test-agnostic class distributions based on only \emph{unlabeled test data}.
To tackle this challenge, we empirically investigate the property of different experts, and   observe that there is \emph{a positive correlation between  expertise and prediction stability}, \ie stronger experts have higher prediction consistency between  different perturbed views of samples from their favorable classes.   
Motivated by this finding, we develop  a novel  self-supervised strategy, namely prediction stability maximization,    to adaptively aggregate  experts  based on only  unlabeled test data. 
We theoretically show that  maximizing the prediction stability  enables SADE to learn an aggregation weight that maximizes  the mutual information between the predicted
label distribution     and the true class distribution. In this way, the resulting  model is able to   simulate  unknown  test class distributions.
  
We empirically verify the superiority of SADE on both vanilla and test-agnostic long-tailed recognition. Specifically,  SADE achieves promising performance on   vanilla long-tailed recognition under all benchmark datasets.   For instance, SADE achieves 58.8$\%$   accuracy  on ImageNet-LT with more than 2$\%$   accuracy gain over previous state-of-the-art  ensemble long-tailed methods, \ie RIDE~\cite{wang2020long} and ACE~\cite{cai2021ace}. 
More importantly,  SADE is the first long-tailed approach that is able to handle various  test-agnostic class distributions without knowing the true class   distribution  of  test data in advance. Note that  SADE even outperforms LADE~\cite{hong2020disentangling} that uses      knowledge of the test class distribution.

Compared to previous long-tailed methods (\eg LADE~\cite{hong2020disentangling} and RIDE~\cite{wang2020long}), our method   offers the following advantages: (i) SADE does not assume the   test class distribution to be  known, and  provides the first practical approach to handling test-agnostic long-tailed recognition; (ii) SADE develops a simple diversity-promoting strategy to learn   skill-diverse experts from a  single and stationary long-tailed dataset;  (iii) SADE presents a novel self-supervised strategy to  aggregate skill-diverse experts at test time, by maximizing prediction consistency  between    unlabeled test samples'   perturbed views; (iv) the presented self-supervised strategy has a provable ability to  simulate     test-agnostic class distributions, which opens the opportunity for tackling unknown  class distribution shifts at test time. 
 
\section{Related Work}  
\textbf{Long-tailed recognition}~~~ 
Existing long-tailed recognition methods, related to our study, can be categorized into three types: class re-balancing, logit adjustment and ensemble learning. Specifically, class  re-balancing resorts to   re-sampling~\cite{chawla2002smote,guo2021long,huang2016learning,kang2019decoupling} or   cost-sensitive learning~\cite{cao2019learning,deng2021pml,he2022relieving,zhao2018adaptive} to balance different classes during model training. Logit adjustment~\cite{hong2020disentangling,menon2020long,peng2021optimal,tian2020posterior}   adjusts models' output logits   via the label frequencies of training data at inference time,  for obtaining a large relative margin between head  and tail classes. Ensemble-based methods~\cite{cai2021ace,guo2021long,xiang2020learning,zhou2020bbn}, \eg   RIDE~\cite{wang2020long}, are based on  multiple experts, which seek to capture heterogeneous knowledge, followed by ensemble aggregation. More discussions on the difference between our method  and RIDE~\cite{wang2020long}  are provided in Appendix~\ref{cls_expert_learning_supp}.
Regarding test-agnostic long-tailed recognition, LADE~\cite{hong2020disentangling} assumes the   test class distribution   is available  and uses it to post-adjust model predictions. However, the true test class distribution is usually unknown a priori, making LADE inapplicable. In contrast, our  method does not rely on the true test  distribution for   handling this problem, but presents   a novel self-supervised  strategy to aggregate skill-diverse experts at test time for    test-agnostic class  distributions. 
Moreover,  some ensemble-based long-tailed methods~\cite{sharma2020long}   aggregate   experts   based on a \emph{labeled} uniform validation set. However, as the test class distribution could be different from the   validation one, simply aggregating experts   on the validation set is unable to handle   test-agnostic long-tailed recognition.

\textbf{Test-time training}~~~
Test-time training~\cite{kamani2020targeted,kim2020learning,liu2021ttt++,sun2020test,wang2021tent} is a transductive learning paradigm for handling    distribution shifts~\cite{lin2022prototype,long2014transfer,niu2022efficient,qiu2021source,varsavsky2020test,zhang2020collaborative} between training and test  data, and  has been applied with  success to   out-of-domain generalization~\cite{iwasawa2021test,pandey2021generalization} and  dynamic scene deblurring~\cite{chi2021test}. In this study, we  explore this  paradigm   to handle test-agnostic long-tailed recognition, where the issue of class distribution shifts is the main challenge.   However, most existing test-time training methods seek to handle covariate distribution shifts instead of  class distribution shifts, so simply leveraging  them   cannot resolve test-agnostic long-tailed recognition, as shown in our experiment (cf. Table~\ref{table_agnostic_imagenet_testime}).

\section{Problem Formulation}\label{sec3}

Long-tailed recognition aims to learn a well-performing classification model from a training dataset  with long-tailed class distribution. Let $\mathcal{D}_s\small{=}\{x_i,y_i\}_{i=1}^{n_s}$ denote the long-tailed training set, where $y_i$ is the class label of the sample $x_i$. The total number of training data over $C$ classes is $n_s \small{=} \sum_{k=1}^C n_k$, where  $n_k$ denotes the  number of samples in   class $k$. Without loss of generality, we follow a common assumption~\cite{hong2020disentangling,kang2019decoupling} that the    
classes are sorted by cardinality in decreasing order (\ie if $i_1 < i_2$, then $n_{i_1}\geq n_{i_2}$), and $n_1 \gg n_C$. The imbalance ratio is defined as $\max(n_k)$/$\min(n_k)=n_1$/$n_C$. The test data $\mathcal{D}_t=\{x_j,y_j\}_{j=1}^{n_t}$ is defined in a similar way. 

Most existing long-tailed recognition methods assume the  test class distribution is uniform (\ie $p_t(y)=1/C$), and    seek to train models from the long-tailed training distribution $p_s(y)$ to perform well on the uniform test distribution. However,  such an assumption does not always hold    in practice. The actual test class distribution in real-world applications may also   be  long-tailed (\ie $p_t(y)=p_s(y)$), or even inversely long-tailed to the training data (\ie $p_t(y)= \text{inv}(p_s(y))$). Here, $\text{inv}(\cdot)$ indicates that  the order of the long tail on  classes is flipped. As a result, the models learned by  existing methods may fail when the actual test class distribution is different from the assumed one.
To address this, we propose to study a more practical yet challenging long-tailed  problem, \ie \textbf{Test-agnostic Long-tailed Recognition}. This task aims to learn a recognition model from long-tailed training data, where the resulting model would be  evaluated on multiple  test sets  that follow different class distributions. This task is   challenging due to the integration of two challenges: (1) the severe  class imbalance in the training data makes  it difficult to train models; (2) unknown class distribution shifts between training and test data (\ie $p_t(y)\neq p_s(y)$) makes   models hard to generalize.

 \begin{table}[t] 
	\caption{Accuracy of existing long-tailed  (LT)  methods on ImageNet-LT  with various test class distributions, including uniform,  forward and backward LT distributions with imbalance ratios of 10 and 50, respectively. The results show that  each method  strives  to simulate a specific   class distribution in terms of  many-shot, medium-shot and few-shot classes, which does not change when the  test class distribution varies. The corresponding visualization results are reported in Figure~\ref{visualization_existing} in Appendix~\ref{cls_testtime_learning_supp}.}  \vspace{-0.05in}
	\label{performance_existing} 
 \begin{center}    
 \begin{threeparttable} 
    \resizebox{0.8\textwidth}{!}{
 	\begin{tabular}{lccccccccccc}\toprule 
        \multirow{2}{*}{Test class  distribution~~}&\multicolumn{3}{c}{Softmax}&&\multicolumn{3}{c}{Balanced Softmax~\cite{jiawei2020balanced}}&&\multicolumn{3}{c}{LADE w/o prior~\cite{hong2020disentangling}}  \cr\cmidrule{2-4}\cmidrule{6-8}\cmidrule{10-12}
        & Many  & Medium & Few  &&  Many  & Medium & Few &&  Many  & Medium & Few \cr
        \midrule
         Forward-LT-50   &67.5	 	& 41.7	& 14.0	&& 63.5	 	& 47.8	& 37.5  && 63.5	 	& 46.4	& 33.1  \\ 
         Forward-LT-10   &68.2	 	& 40.9	& 14.0	&& 64.1	 	& 48.2	& 31.2  &&64.7	 	& 47.1	& 32.2	\\ 
         Uniform    &  68.1         &41.5	 	& 14.0 	&& 64.1         & 48.2	 	& 33.4	&& 64.4 & 47.7 & 34.3 \\ 
        Backward-LT-10   & 67.4	& 41.9 & 13.9 	&& 63.4	& 49.1 & 33.6 && 	64.4	& 48.2 & 34.2 \\ 
         Backward-LT-50   & 70.9	& 41.1 & 13.8 	&& 66.5	& 48.4 & 33.2 && 	66.3	& 47.8 & 34.0 \\ 
        \bottomrule

	\end{tabular}}
	 \end{threeparttable}
	 \end{center}  \vspace{-0.1in} 
\end{table}

\section{Method}
\label{method}
To tackle the above   problem,  inspired by the idea of "divide and conquer", we propose to learn  multiple skill-diverse experts that excel at handling different class distributions. By reasonably fusing these experts at test time, the multi-expert model would manage to handle   unknown class distribution shifts and resolve   test-agnostic long-tailed recognition. Following this idea,  we develop a novel Self-supervised Aggregation of Diverse Experts (SADE) approach. Specifically, SADE consists of two innovative  strategies: (1) \emph{learning skill-diverse experts}   from a single long-tailed training dataset; (2) \emph{test-time aggregating experts   with self-supervision} to handle    test-agnostic class distributions.

\subsection{Skill-diverse Expert Learning}
\label{expert_learning}   

\begin{wrapfigure}{r}{0.5\textwidth} 
  \centering
  \vspace{-3pt}
\includegraphics[width=7cm]{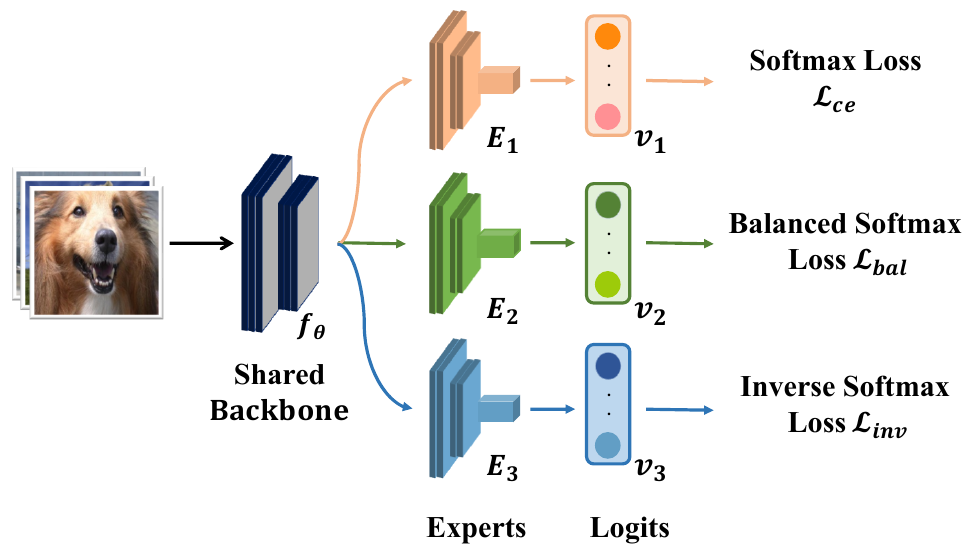}
  \vspace{-10pt}
  \caption{The scheme of SADE with  three experts, where different experts are trained with    different expertise-guided losses.
  }  \vspace{-9pt}
  \label{scheme}  
\end{wrapfigure}

As shown in Figure~\ref{scheme}, SADE builds a three-expert model that comprises two components: (1) an expert-shared  backbone $f_{\theta}$; (2) independent expert networks   $E_1$, $E_2$ and $E_3$. When training the model, the key challenge is how to learn  skill-diverse experts from a single and  stationary long-tailed training dataset.
Existing ensemble-based long-tailed  methods~\cite{guo2021long,wang2020long} seek to  train experts for   the uniform test  class distribution, and hence the trained experts are not  differentiated sufficiently for handling various class distributions (refer to   Table~\ref{table_expert_learning} for an example). To tackle this challenge, we first empirically investigate existing long-tailed methods in this task. From Table~\ref{performance_existing}, we find   that  there is a \emph{simulation correlation between the learned class distribution and the training loss function}. That is, the models learned by different losses are good at dealing with   class distributions with different skewness. For instance, the model trained with the  softmax loss is good at   the long-tailed  distribution, while the models  obtained from  long-tailed methods are skilled in the uniform   distribution.   

Motivated by this finding, we develop a simple  skill-diverse  expert learning strategy   to generate experts with different  distribution preferences. To be specific, the forward  expert $E_1$ seeks to be good at the  long-tailed class distribution and performs well on many-shot classes. The uniform  expert $E_2$ strives to be skilled in the  uniform  distribution. The backward  expert $E_3$ aims at  the inversely long-tailed distribution and performs well on few-shot classes.
Here, the  forward and  backward experts are necessary since they span a wide spectrum of possible   class distributions, while the  uniform expert ensures  retaining high accuracy on the uniform   distribution. 
To this end, we use three different expertise-guided losses to train the three experts, respectively.

\textbf{The forward expert} $E_1$~~~ We  use the softmax cross-entropy loss to train this expert, so that it   directly simulates the original long-tailed training class distribution:
\begin{align}\label{softmax_loss}
    \mathcal{L}_{ce} = \frac{1}{n_s}\sum_{x_i\in \mathcal{D}_s}  -y_i \log \sigma(v_1(x_i)),
\end{align}
where $v_1(\cdot)$ is the output logits of the forward expert $E_1$, and $\sigma(\cdot)$ is the softmax function.

  \textbf{The uniform expert} $E_2$~~~  We aim to train this expert to simulate the uniform class distribution. Inspired by the effectiveness of logit adjusted losses for long-tailed recognition~\cite{menon2020long}, we resort to the balanced softmax loss~\cite{jiawei2020balanced}. Specifically, let $\hat{y}^k=\frac{\exp(v^k)}{\sum_{c=1}^C \exp(v^c)}$ be the  prediction probability. The  balanced softmax    adjusts the prediction probability    by compensating for the long-tailed  class distribution with  the prior of  training label frequencies: 
   $ \hat{y}^k = \frac{\pi^k\exp(v^k)}{\sum_{c=1}^C \pi^c\exp(v^c)}= \frac{\exp(v^k + \log \pi^k)}{\sum_{c=1}^C \exp(v^c+ \log \pi^c)}$, 
where $\pi^k = \frac{n_k}{n} $ denotes the training label frequency of    class $k$. Then, given $v_2(\cdot)$ as  the output logits of the  expert $E_2$,   the balanced softmax loss for the  expert $E_2$ is defined as:
\begin{align}\label{balanced_loss}
    \mathcal{L}_{bal}=\frac{1}{n_s}\sum_{x_i\in \mathcal{D}_s} -y_i \log \sigma(v_2(x_i) + \log \pi).
\end{align} 
Intuitively, by adjusting   logits to compensate for the long-tailed      distribution with the prior $\pi$, this loss enables $E_2$ to output class-balanced predictions that simulate the uniform  distribution.

\textbf{The backward expert} $E_3$~~~  We seek to train this expert to simulate  the inversely long-tailed class distribution. To this end, we propose a new \emph{inverse softmax loss}, based on the same rationale of logit adjusted losses~\cite{jiawei2020balanced,menon2020long}. Specifically, we adjust the prediction probability  by: 
  $  \hat{y}^k = \frac{\exp(v^k + \log \pi^k -  \log \bar{\pi}^k)}{\sum_{c=1}^C \exp(v^c + \log \pi^c  - \log \bar{\pi}^c)}$,   
where  the inverse training prior  $\bar{\pi}$ is obtained by inverting the order of training label frequencies $\pi$. Then, the new inverse softmax loss for the expert $E_3$ is defined as:
\begin{align} \label{inverse_loss}
    \mathcal{L}_{inv} \small{=} \frac{1}{n_s}\sum_{x_i\in \mathcal{D}_s} -y_i \log \sigma(v_3(x_i)  \small{+} \log \pi  \small{-} \lambda \log \bar{\pi}), 
\end{align} 
where $v_3(\cdot)$ denotes  the output logits of $E_3$ and $\lambda$ is a hyper-parameter. Intuitively, this loss adjusts logits to compensate for the long-tailed     distribution with $\pi$, and further applies  reverse adjustment  with~$\bar{\pi}$. This   enables $E_3$ to   simulate the inversely long-tailed distribution   (cf. Table~\ref{table_expert_learning} for  verification).

\subsection{Test-time Self-supervised Aggregation}
\label{Test_Time}
Based on the skill-diverse learning strategy, the  three experts in SADE are skilled   in different class distributions.  The remaining challenge is how to fuse them to deal with unknown test class distributions. 
A basic principle for expert aggregation is that the experts should play a bigger role in situations where they have expertise.  
Nevertheless, how to detect strong experts for  unknown  test class distribution remains unknown.  
Our key insight is that strong experts should be  more stable in predicting the samples from their skilled classes,  even though these samples are perturbed. 

\vspace{0.05in}
\textbf{Empirical observation}~~~ To verify this hypothesis, we estimate the prediction stability of  experts by comparing the cosine similarity between their predictions for a  sample's two augmented views. Here, the data views are generated by  the   data augmentation techniques in MoCo v2~\cite{chen2020improved}. From Table~\ref{cos}, we find  that there is a  \emph{positive correlation between  expertise and prediction stability}, \ie stronger experts have higher prediction similarity between  different views of samples from their favorable classes.  
Following this finding, we propose to explore the relative prediction stability to detect strong experts and weight   experts for the  unknown  test class  distribution. Consequently, we develop a novel self-supervised  strategy, namely prediction stability maximization.

\vspace{0.05in} 
\textbf{Prediction stability maximization}~~~ This strategy learns aggregation weights for  experts (with frozen parameters) by maximizing model  prediction stability  for unlabeled test samples.  As shown in Figure~\ref{ssl}, the method comprises  three  major components as follows.

\emph{Data view generation}~~~ For a given sample $x$, we conduct two stochastic data augmentations to generate the sample's two   views, \ie $x^1$ and $x^2$. 
Here,  we use the same   augmentation techniques  as the advanced contrastive learning method, \ie MoCo v2~\cite{chen2020improved}, which has been shown   effective in self-supervised learning.

 \emph{Learnable aggregation weight}~~~ 
Given the output logits of three experts $(v_1,v_2,v_3) \in \mathbb{R}^{3\times C}$, we aggregate  experts with a  learnable aggregation weight  $w = [w_1,w_2,w_3] \in \mathbb{R}^3$  and obtain the final softmax prediction by $\hat{y} = \sigma(w_1\small{\cdot} v_1 + w_2\small{\cdot} v_2 + w_3\small{\cdot} v_3)$, where $w$ is normalized before aggregation, \ie $w_1+w_2+w_3\small{=}1$.

\emph{Objective function}~~~ Given the view predictions of unlabeled test data,  we maximize the prediction stability based on the cosine similarity between the view predictions: 
\begin{align}\label{sta_loss}
  \max_{w} ~\mathcal{S}, ~~ \text{where} ~~  \mathcal{S}  = \frac{1}{n_t}\sum_{x\in \mathcal{D}_t}  \hat{y}^1 \cdot \hat{y}^2.
\end{align} 
Here, $\hat{y}^1$ and $\hat{y}^2$ are normalized by the softmax function. In test-time training, only the aggregation weight $w$ is updated.  Since stronger experts have higher  prediction similarity for their skilled classes, maximizing the prediction stability $\mathcal{S}$ would learn higher weights for stronger experts regarding the unknown test class distribution. Moreover, the self-supervised aggregation strategy can be conducted in an online manner  for streaming test data. The  pseudo-code of SADE is provided in Appendix~\ref{App_B}.

\begin{table}[t]
	\begin{minipage}{0.485\textwidth}
	\caption{Prediction stability of experts in terms of the cosine similarity between their predictions of a sample's two views. Note that  expert $E_1$ is good at many-shot classes and expert $E_3$ is skilled in few-shot classes.   The experts tend to have better   prediction consistency for the samples from their skilled classes. Here, the  imbalance ratio  of CIFAR100-LT is 100.}	\label{cos} 
	 \vspace{3pt}
 \begin{center}
 \begin{threeparttable} 
    \resizebox{0.95\textwidth}{!}{
 	\begin{tabular}{lccccccc}\toprule 
 	
        \multirow{4}{*}{Model}&\multicolumn{7}{c}{Cosine similarity between view predictions}\cr\cmidrule{2-8}
        &\multicolumn{3}{c}{ImageNet-LT}&&\multicolumn{3}{c}{CIFAR100-LT}\cr\cmidrule{2-4}\cmidrule{6-8}
        & Many  & Med. & Few  &&  Many  & Med. & Few  \cr
        \midrule
        Expert  $E_1$    & 0.60	 	&  0.48	& 0.43	&&  0.28	 	& 0.22	& 0.20   \\
        Expert  $E_2$  & 0.56	 	&  0.50	& 0.45	&&  0.25	 	& 0.21	& 0.19   \\
        Expert  $E_3$    &  0.52        & 0.53	 	& 0.58 	&& 0.22        & 0.23	 	& 0.25	  \\
        \bottomrule

	\end{tabular}}
	 \end{threeparttable}
	 \end{center}
	\end{minipage}
	~~
	\begin{minipage}{0.485\textwidth}
 \centering
  \vspace{-7pt}
\includegraphics[width=5.6cm]{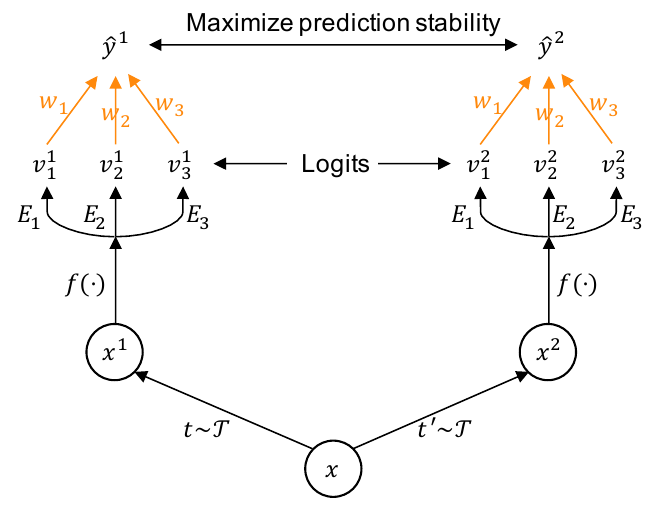} 
\captionof{figure}{The scheme of test-time self-supervised aggregation. Two data augmentations sampled  from the same family of augmentations ($t\small{\sim}\mathcal{T}$ and $t'\small{\sim}\mathcal{T}$)  are applied  to obtain two data views.} \label{ssl} 
    \vspace{-10pt}
    	\end{minipage} 
    	 \vspace{-5pt}
    \end{table}
  
\vspace{0.05in}
\textbf{Theoretical Analysis}~~~ 
We then theoretically analyze the  prediction stability maximization strategy to conceptually understand why it works. To this end, we first define the random variables of predictions  and labels as $\hat{Y}\small{\sim}p(\hat{y})$  and $Y\small{\sim}p_t(y)$. We have the following result:
\begin{thm} \label{thm1} 
The prediction stability  $\mathcal{S}$ is positive proportional  to  the mutual information between the predicted
label   distribution    and the test class distribution $I(\hat{Y};Y)$, and negative proportional  to the prediction entropy $H(\hat{Y})$: 
\begin{align} 
    \mathcal{S}~\propto~ I(\hat{Y};Y) - H(\hat{Y}).  \nonumber
\end{align}
\end{thm} 
Please refer to   Appendix~\ref{App_A} for proofs. According to Theorem~\ref{thm1}, maximizing the  prediction stability  $\mathcal{S}$ enables SADE to learn an aggregation weight that maximizes the mutual information between the predicted label distribution   $p(\hat{y})$   and the test class distribution $p_t(y)$, as well as minimizing the prediction entropy. Since  minimizing  entropy helps to improve the confidence of the classifier output~\cite{grandvalet2005semi}, the  aggregation weight is learned to simulate the test class distribution $p_t(y)$ and increase  the prediction confidence. This property intuitively explains why   our method has the potential to tackle the challenging task of test-agnostic long-tailed recognition at test time.

\section{Experiments}
In this section, we first evaluate the superiority of SADE  on both  vanilla  and test-agnostic  long-tailed recognition.  We  then   verify the effectiveness of SADE in terms of its  two strategies, \ie skill-diverse expert learning  and  test-time self-supervised aggregation. More ablation studies are reported in  appendices. Here, we begin with the experimental settings.

\subsection{Experimental Setups}
\label{exp_setup}  
 
 \textbf{Datasets}~~~
We use four benchmark  datasets (\ie ImageNet-LT~\cite{liu2019large}, CIFAR100-LT~\cite{cao2019learning}, Places-LT~\cite{liu2019large}, and iNaturalist 2018~\cite{van2018inaturalist}) to simulate real-world long-tailed class distributions. Their data  statistics and imbalance ratios are summarized   in Appendix~\ref{App_C1}. The imbalance ratio is defined as $\max{n_j}$/$\min{n_j}$, where $n_j$ denotes the data number of class $j$.  Note that CIFAR100-LT has three variants with different imbalance ratios.

\textbf{Baselines}~~~
We  compare SADE with   state-of-the-art long-tailed methods, including two-stage methods (Decouple~\cite{kang2019decoupling},  MiSLAS~\cite{zhong2021improving}), logit-adjusted training (Balanced Softmax~\cite{jiawei2020balanced}, LADE~\cite{hong2020disentangling}),  ensemble learning (BBN~\cite{zhou2020bbn}, ACE~\cite{cai2021ace}, RIDE~\cite{wang2020long}), classifier design (Causal~\cite{tang2020long}),  and representation learning (PaCo~\cite{cui2021parametric}). Note that LADE uses  the prior of test class distribution    for post-adjustment (although it is unavailable in practice), while all other methods do not use this   prior.
  
\textbf{Evaluation protocols}~~~
In  test-agnostic long-tailed recognition, following LADE~\cite{hong2020disentangling}, the  models are  evaluated on multiple sets of test data that  follow  different class distributions,  in terms of micro accuracy.  Same as LADE~\cite{hong2020disentangling}, we construct three kinds of test class distributions, \ie the uniform  distribution, forward long-tailed distributions as training data, and backward long-tailed distributions. In the backward ones, the order of the long tail on  classes  is flipped. More details of test data construction are provided in Appendix~\ref{App_C2}.  
Besides, we also evaluate   methods on  vanilla long-tailed recognition~\cite{kang2019decoupling,liu2019large}, where the  models are evaluated on the uniform test class distribution. 
Here, the   accuracy on three class sub-groups is also   reported, \ie many-shot classes (more than 100 training images), medium-shot classes (20$\sim$100 images) and few-shot classes (less than 20 images).

 \textbf{Implementation details}~~~
 We use the same setup for all the baselines and our method.  Specifically, following~\cite{hong2020disentangling,wang2020long}, we use ResNeXt-50 for ImageNet-LT, ResNet-32 for CIFAR100-LT, ResNet-152 for Places-LT and ResNet-50 for iNaturalist 2018 as   backbones, respectively. Moreover, we adopt the cosine classifier for prediction on all datasets. If not specified, we use the SGD optimizer with the momentum of 0.9 for training 200 epochs and set the initial
learning rate as 0.1 with linear decay. We set $\lambda\small{=}2$ for ImageNet-LT and CIFAR100-LT, and $\lambda\small{=}1$  for the remaining datasets.  During  test-time training, we train the aggregation weights for 5 epochs with the batch size 128, where we  use the same  optimizer and learning rate as the training phase. More implementation details and the hyper-parameter statistics  are reported in Appendix~\ref{App_C3}.

\begin{table*}[t]  
 \caption{Top-1 accuracy on CIFAR100-LT, Places-LT and iNaturalist 2018, where the test class distribution is uniform. More results on three class sub-groups    are reported in Appendix~\ref{App_D3}.}  
  \label{Result_all_uniform} 
  \begin{center}   
  \begin{minipage}{0.32\linewidth}
 \centerline{\small{(a) CIFAR100-LT}}
 \begin{center}
 \begin{threeparttable} 
    \resizebox{1\textwidth}{!}{
 	\begin{tabular}{lccc}\toprule 
 	
        \multirow{1}{*}{Imbalance Ratio}& 
         10 &50 &100\cr 
        \midrule
        Softmax  & 59.1	 	&  45.6	& 41.4   \\
        BBN~\cite{zhou2020bbn} & 59.8	 	&  49.3	& 44.7\\
        Causal~\cite{tang2020long}  & 59.4	 	& 48.8 & 45.0\\
        Balanced Softmax~\cite{jiawei2020balanced}    &  61.0	 	& 50.9	& 46.1	 \\
        MiSLAS~\cite{zhong2021improving}   & 62.5     & 51.5 	& 46.8  \\ 
        LADE~\cite{hong2020disentangling} & 61.6      & 50.1	 	& 45.6 \\
        RIDE~\cite{wang2020long} & 61.8       &51.7	 	& 48.0 \\\midrule
        SADE (ours)  & \textbf{63.6}      & \textbf{53.9}	 	& \textbf{49.8} \\
        \bottomrule
	\end{tabular}}
	 \end{threeparttable}
	 \end{center}
 \end{minipage}
 \hfill 
 \begin{minipage}{0.326\linewidth}
  \centerline{\small{(b) Places-LT}}
  \begin{center}
 \begin{threeparttable} 
    \resizebox{1\textwidth}{!}{
 	\begin{tabular}{lc}\toprule 
        \multirow{1}{*}{Method}& 
         Top-1 accuracy  \cr 
        \midrule
        Softmax  & 31.4	  \\
        Causal~\cite{tang2020long}  &  32.2  \\
        Balanced Softmax~\cite{jiawei2020balanced}    & 39.4	 	 \\
        MiSLAS~\cite{zhong2021improving}   & 38.3      \\ 
        LADE~\cite{hong2020disentangling} & 39.2   \\
        RIDE~\cite{wang2020long} &40.3     \\\midrule
        SADE (ours)  & \textbf{40.9}      \\
        \bottomrule
	\end{tabular}}
	
	 \end{threeparttable}
	 \end{center}
  \end{minipage}
\hfill 
 \begin{minipage}{0.326\linewidth}
 
  \centerline{\small{(c) iNaturalist 2018}}
  \begin{center}
 \begin{threeparttable} 
    \resizebox{1\textwidth}{!}{
 	\begin{tabular}{lccccccccc}\toprule 
 	
        \multirow{1}{*}{Method}& 
         Top-1 accuracy  \cr 
        \midrule
        Softmax  & 64.7 	  \\
        Causal~\cite{tang2020long}  &  64.4	  \\
        Balanced Softmax~\cite{jiawei2020balanced}    &  70.6	 	 \\
        MiSLAS~\cite{zhong2021improving}   & 70.7        \\ 
        LADE~\cite{hong2020disentangling} & 69.3  \\
        RIDE~\cite{wang2020long} & 71.8      \\\midrule
        SADE (ours) & \textbf{72.9}      \\
        \bottomrule
	\end{tabular}}
	 \end{threeparttable} 
	 \end{center}
  \end{minipage}
  \end{center} 
 \vspace{-0.1in}
  
\end{table*}

  \begin{wraptable}{r}{0.41\textwidth} 
  \vspace{-0.2in}
	\caption{Top-1 accuracy on ImageNet-LT.} 
	\label{table_imagenet_uniform}  
  \begin{center} 
 \begin{threeparttable} 
    \resizebox{0.4\textwidth}{!}{
 	\begin{tabular}{lcccc}\toprule 
 	
        \multirow{1}{*}{Method} & Many  & Med. & Few & All\cr
        \midrule
        Softmax  & 68.1  	&  41.5	& 14.0  & 48.0  \\ 
        Decouple-LWS~\cite{kang2019decoupling} &61.8	 	&  47.6	& 30.9 & 50.8 \\
        Causal~\cite{tang2020long}  & 64.1 	 	&  45.8 & 27.2  & 50.3\\
        Balanced Softmax~\cite{jiawei2020balanced}    &  64.1	 	& 48.2	& 33.4	& 52.3 \\
        MiSLAS~\cite{zhong2021improving}   & 62.0     & 49.1	& 32.8 & 51.4 \\ 
        LADE~\cite{hong2020disentangling} & 64.4     & 47.7	 	& 34.3 & 52.3 \\
        PaCo~\cite{cui2021parametric}& 63.2 	& 51.6 & 39.2  & 54.4 \\
        ACE~\cite{cai2021ace}&   \textbf{71.7}	& 54.6 & 23.5  & 56.6 \\
        RIDE~\cite{wang2020long} & 68.0       & 52.9	 	& 35.1 & 56.3\\\midrule
        SADE (ours)  & 66.5      & \textbf{57.0}	 	& \textbf{43.5} & \textbf{58.8} \\
        \bottomrule

  \vspace{-0.2in}
	\end{tabular}}
	 \end{threeparttable}
	 \end{center} 
\end{wraptable}

\subsection{Superiority on Vanilla Long-tailed Recognition}
This subsection compares   SADE with state-of-the-art long-tailed   methods   on vanilla long-tailed recognition. Specifically, as shown in Tables~\ref{Result_all_uniform}-\ref{table_imagenet_uniform},   Softmax   trains the model with only cross-entropy, so it simulates the long-tailed training   distribution and performs   well on many-shot classes. However, it performs poorly on medium-shot  and few-shot classes, leading to   worse overall performance. In contrast, existing long-tailed  methods (\eg Decouple, Causal) seek to simulate the  uniform   class distribution, so their performance  is more class-balanced, leading to better overall performance. However,  as these methods mainly seek    balanced performance, they inevitably sacrifice the performance on   many-shot classes. To address this, RIDE and ACE explore ensemble learning for long-tailed recognition and achieve  better performance on tail classes without sacrificing the head-class performance. In comparison, based on the increasing expert diversity derived from   skill-diverse expert learning, our method performs  the best on all datasets, \eg  with more than 2$\%$ 
accuracy gain on  ImageNet-LT    compared to RIDE and ACE. These results  demonstrate  the superiority  of SADE over the compared methods that are particularly designed  for the uniform test  class distribution. Note that SADE also outperforms baselines  in experiments with stronger data augmentation (\ie RandAugment~\cite{cubuk2020randaugment}) and other architectures, as reported in Appendix~\ref{App_D3}.

 \newpage 
\subsection{Superiority on Test-agnostic Long-tailed Recognition}\label{sec_5.5}
 
In this subsection, we evaluate   SADE on test-agnostic long-tailed recognition.  The   results  on    various  test class distributions are reported in Table~\ref{table_agnostic_imagenet}. 
Specifically, since Softmax  seeks to simulate  the long-tailed training    distribution,  it performs well on forward long-tailed test   distributions. However, its performance  on the uniform and backward long-tailed distributions is poor. In contrast, existing long-tailed methods   show more  balanced performance among classes, leading to better overall accuracy. However, the resulting models by these methods   suffer from a simulation bias, \ie performing similarly among classes on various  class distributions (c.f. Table~\ref{performance_existing}). As a result, they cannot adapt to diverse test  class distributions well.
To handle this task, LADE assumes the test class distribution to be known and 
uses this information to adjust its predictions,  leading to better performance on various   test class distributions. However, since  obtaining the actual test class distribution is difficult in real applications,   the methods requiring such knowledge  may be not applicable in practice. 
Moreover, in some specific cases like Forward-LT-3 and Backward-LT-3 distributions of iNaturalist 2018,  the  number of test samples on some   classes becomes zero.  In such cases, the test prior cannot be used in LADE, since adjusting logits with $\log0$   results in   biased predictions. 
In contrast,  without   relying on the knowledge of test class distributions, our  SADE presents an innovative  self-supervised strategy to deal with unknown class distributions, and    obtains even   better performance than LADE that uses the test class prior (c.f. Table~\ref{table_agnostic_imagenet}). 
The promising  results demonstrate the effectiveness and  practicality of our  method on test-agnostic long-tailed recognition.  Note that  the performance advantages of SADE    become larger   as the test data  get  more imbalanced.   Due to the page limitation, the   results on more datasets  are reported in Appendix~\ref{App_D4}.

\begin{table}[t]  \vspace{-0.15in} 
	\caption{Top-1 accuracy  on long-tailed datasets with various unknown test class distributions. ``Prior" indicates that the test class distribution is used as prior knowledge.  ``Uni." denotes the uniform distribution. ``IR" indicates the imbalance ratio. ``BS" denotes the balanced softmax~\cite{jiawei2020balanced}.}\vspace{-0.1in} 
	\label{table_agnostic_imagenet} 
 \begin{center}
 \begin{threeparttable} 
    \resizebox{0.99\textwidth}{!}{
 	\begin{tabular}{lcccccccccccccccccccccccccccc}\toprule  
   	     \multirow{4}{*}{Method} &\multirow{4}{*}{Prior}&  \multicolumn{13}{c}{(a) ImageNet-LT} && \multicolumn{13}{c}{(b) CIFAR100-LT (IR100)}\cr \cmidrule{3-15}  \cmidrule{17-29}  
       && \multicolumn{5}{c}{Forward-LT} && Uni.  && \multicolumn{5}{c}{Backward-LT} && \multicolumn{5}{c}{Forward-LT} && Uni. && \multicolumn{5}{c}{Backward-LT} \cr  \cmidrule{3-7} \cmidrule{9-9}\cmidrule{11-15} \cmidrule{17-21} \cmidrule{23-23}\cmidrule{25-29} 
         && 50 &25 &10&5&2 && 1 && 2& 5& 10 &25 &50& &50 &25 &10&5&2 && 1 && 2& 5& 10 &25 &50 \cr  
        \midrule
        Softmax  & \xmark & 66.1 & 63.8 & 	  60.3& 56.6&52.0 && 48.0 && 43.9 & 38.6 & 34.9 & 30.9 & 27.6 && 63.3&62.0&56.2 & 52.5 & 46.4 && 41.4 && 36.5 & 30.5 & 25.8 & 21.7 & 17.5  \\  
        BS    & \xmark & 63.2	 	& 61.9	& 59.5	& 57.2 & 54.4&&52.3 && 50.0 & 47.0 & 45.0 & 42.3 & 40.8 && 57.8 & 55.5 & 54.2 & 52.0 & 48.7 && 46.1 && 43.6 & 40.8 & 38.4 & 36.3 & 33.7  \\ 
        MiSLAS    & \xmark& 61.6 & 60.4 & 58.0 & 56.3 & 53.7 && 51.4 && 49.2 & 46.1& 44.0 & 41.5 & 39.5 && 58.8 & 57.2 & 55.2 & 53.0 & 49.6 && 46.8 && 43.6 & 40.1 & 37.7 & 33.9 & 32.1 \\ 
        LADE  &\xmark & 63.4 & 62.1 & 59.9 & 57.4 & 54.6 && 52.3 && 49.9 & 46.8 & 44.9 & 42.7 &40.7 && 56.0 & 55.5 &52.8 &  51.0 & 48.0 && 45.6 && 43.2 & 40.0 & 38.3 & 35.5 & 34.0  \\
        LADE  & \cmark&   65.8 & 63.8 & 60.6 & 57.5 & 54.5 && 52.3 &&   50.4 & 48.8 & 48.6 & 49.0 &49.2 && 62.6 & 60.2 & 55.6 & 52.7 & 48.2 && 45.6 && 43.8 & 41.1 & 41.5 & 40.7 & 41.6 \\
        RIDE  & \xmark &67.6 & 66.3      & 64.0 & 61.7 & 58.9 && 56.3 && 54.0 & 51.0 & 48.7 & 46.2 & 44.0&& 63.0 & 59.9 & 57.0 & 53.6 & 49.4 && 48.0 && 42.5 & 38.1 & 35.4 & 31.6 & 29.2  \\\midrule
        SADE  & \xmark & \textbf{69.4} &  \textbf{67.4}    & \textbf{65.4} 	& \textbf{63.0}  & \textbf{60.6} && \textbf{58.8} && \textbf{57.1} & \textbf{55.5} & \textbf{54.5} & \textbf{53.7} & \textbf{53.1} && \textbf{65.9} & \textbf{62.5} & \textbf{58.3} & \textbf{54.8} & \textbf{51.1} && \textbf{49.8} && \textbf{46.2} & \textbf{44.7} & \textbf{43.9} & \textbf{42.5} & \textbf{42.4}\\     
        \midrule\midrule

       \multirow{4}{*}{Method} &\multirow{4}{*}{Prior}&  \multicolumn{13}{c}{(c) Places-LT} && \multicolumn{13}{c}{(d) iNaturalist 2018}\cr \cmidrule{3-15}  \cmidrule{17-29}  
       && \multicolumn{5}{c}{Forward-LT} && Uni. && \multicolumn{5}{c}{Backward-LT} && \multicolumn{5}{c}{Forward-LT} && Uni. && \multicolumn{5}{c}{Backward-LT} \cr  \cmidrule{3-7} \cmidrule{9-9}\cmidrule{11-15} \cmidrule{17-21} \cmidrule{23-23}\cmidrule{25-29} 
         && 50 &25 &10&5&2 && 1 && 2& 5& 10 &25 &50& &  &3 & &2&  && 1 &&  &2&   &3 &  \cr  
        \midrule
        Softmax  & \xmark & 45.6 & 42.7 & 40.2 & 38.0 & 34.1 &&  31.4 && 28.4 & 25.4 & 23.4 & 20.8 & 19.4 &&  &65.4 &&    65.5 &&&64.7 &&&64.0 &&63.4 &  \\  
        BS   & \xmark & 42.7 & 41.7 & 41.3 & 41.0 & 40.0 & & 39.4 && 38.5 & 37.8 & 37.1 & 36.2 & 35.6 &&      &70.3 &&    70.5 &&& 70.6 &&& 70.6 &&70.8 &\\ 
        MiSLAS   & \xmark&  40.9 & 39.7 & 39.5 & 39.6 & 38.8 && 38.3 && 37.3 & 36.7 & 35.8 & 34.7 & 34.4 &&  & 70.8 &&   70.8 &&&70.7 &&& 70.7 &&70.2 & \\ 
        LADE &\xmark & 42.8 & 41.5 & 41.2 & 40.8 & 39.8 && 39.2 && 38.1 & 37.6 & 36.9 & 36.0 & 35.7 &&  & 68.4 &&   69.0 &&& 69.3 &&& 69.6 && 69.5 & \\
        LADE & \cmark&   46.3  & 44.2  & 42.2 & 41.2 & 39.7 && 39.4 && 39.2 & 39.9 &  40.9  & \textbf{42.4} & \textbf{43.0}  && & \xmark &&   69.1 &&& 69.3 &&& 70.2 && \xmark  &\\
        RIDE  & \xmark & 43.1 & 41.8 & 41.6 & 42.0 & 41.0 && 40.3 && 39.6 & 38.7 & 38.2 & 37.0 & 36.9 &&& 71.5 &&   71.9 &&& 71.8 &&& 71.9 && 71.8  & \\\midrule
         SADE   & \xmark &  \textbf{46.4} & \textbf{44.9} & \textbf{43.3} & \textbf{42.6} & \textbf{41.3} & & \textbf{40.9} && \textbf{40.6} & \textbf{41.1} & \textbf{41.4} & 42.0 & 41.6 &&     &\textbf{72.3}  &&  \textbf{72.5} &&& \textbf{72.9}  &&& \textbf{73.5} && \textbf{73.3} & \\    
 
        \bottomrule
    
	\end{tabular}}
	 \end{threeparttable}
	 \end{center}  
 \vspace{-0.1in} 
\end{table}

\begin{table}[t]   
	\caption{Performance of each expert on the uniform test distribution, where the  imbalance ratio  of CIFAR100-LT is 100. The results show that our proposed method   learns multiple experts with higher skill diversity, which  leads to better ensemble performance.} \vspace{-0.1in}
	\label{table_expert_learning} 
 \begin{center}
 \begin{threeparttable} 
    \resizebox{0.85\textwidth}{!}{
 	\begin{tabular}{lccccccccccccccccccc}\toprule 
 	
        \multirow{4}{*}{Model}&\multicolumn{9}{c}{RIDE~\cite{wang2020long}}&&\multicolumn{9}{c}{SADE (ours)}\cr\cmidrule{2-10}\cmidrule{12-20}
        &\multicolumn{4}{c}{ImageNet-LT}&&\multicolumn{4}{c}{CIFAR100-LT}& &\multicolumn{4}{c}{ImageNet-LT}&&\multicolumn{4}{c}{CIFAR100-LT}\cr\cmidrule{2-5}\cmidrule{7-10}\cmidrule{12-15}\cmidrule{17-20}
        & Many  & Med. & Few & All &&  Many  & Med. & Few & All&&  Many  & Med. & Few & All &&  Many  & Med. & Few & All \cr
        \midrule
         Expert  $E_1$   & 64.3	 	&  49.0	& 31.9 & 52.6 &&  63.5	 	& 44.8 & 20.3 & 44.0 & & \textbf{68.8}	 	&  43.7	& 17.2	& 49.8 &&  \textbf{67.6}	 	& 36.3	& 6.8  & 38.4\\
         Expert  $E_2$  &  64.7	 	& 49.4	& 31.2	& 52.8 && 63.1	 	& 44.7	& 20.2 	& 43.8 & & 65.5	 	&  50.5	& 33.3	& \textbf{53.9}&& 61.2	 	& 44.7	& 23.5 & \textbf{44.2}\\
         Expert  $E_3$    & 64.3       & 48.9	 	& 31.8 &	52.5 && 63.9      & 45.1	 	& 20.5 & 44.3 && 43.4         & 48.6   	& \textbf{53.9}  &	 47.3  && 14.0         & 27.6	 	& \textbf{41.2} &	25.8  \\ \cmidrule{1-10}\cmidrule{12-20}
         Ensemble & 68.0      & 52.9	 	& 35.1 &	56.3 && 67.4       & 49.5	 	&  23.7   & 48.0  && 67.0       & 56.7 	 	& 42.6 &	\textbf{58.8} && 61.6        & 50.5	 	&  33.9  & \textbf{49.4} \\  
  
        \bottomrule

	\end{tabular}}
	 \end{threeparttable}
	 \end{center} 
    \vspace{-0.2in}
\end{table}

\subsection{Effectiveness of Skill-diverse Expert Learning}
\label{cls_expert_learning}

We next examine   our  skill-diverse expert learning strategy. The results are  reported in Table~\ref{table_expert_learning}, where RIDE~\cite{wang2020long} is a state-of-the-art ensemble-based method.  RIDE trains each expert with cross-entropy independently  and uses  KL-Divergence     to improve expert  diversity. However, simply maximizing the divergence of expert predictions cannot learn visibly diverse experts (cf. Table~\ref{table_expert_learning}).  In contrast,
the three experts learned by our strategy have significantly diverse expertise, excelling at   many-shot classes,  the uniform  distribution (with higher overall  performance), and few-shot classes, respectively. As a result, the  increasing expert diversity leads to a non-trivial gain for the ensemble performance of  SADE compared to RIDE. 
Moreover,   consistent results on more datasets are reported  in Appendix~\ref{cls_expert_learning_supp}, while the ablation studies of the expert learning  strategy are provided in Appendix~\ref{App_E}.

\begin{table}[t]
 \vspace{-0.15in}
	\begin{minipage}{0.485\textwidth}
	\caption{The expert  weights learned by our self-supervised    strategy on ImageNet-LT with various test class distributions.  Our method   learns suitable   weights for various unknown     distributions.}
	\label{table_test_time}  \vspace{-0.1in}
 \begin{center}
 \begin{threeparttable} 
    \resizebox{0.95\textwidth}{!}{
 	\begin{tabular}{lccc}\toprule 
 	
        \multirow{1}{*}{Test Dist.} & Expert $E_1$ ($w_1$)  & Expert $E_2$ ($w_2$) & Expert $E_3$  ($w_3$)   \cr
        \midrule
        Forward-LT-50   &  0.52	 	&  0.35	& 0.13	  \\
         Forward-LT-10  &  0.46	 	&  0.36	& 0.18 \\
         Uniform    &  0.33       & 0.33 	 &	 0.34 \\
          Backward-LT-10  &  0.21 	& 0.29	& 0.50\\
         Backward-LT-50    &  0.17       &0.27	 &	0.56 \\ 
        \bottomrule

	\end{tabular}}
	 \end{threeparttable}
	 \end{center}   
	\end{minipage}
	~~
	\begin{minipage}{0.485\textwidth}
 	\caption{The performance improvement  by  our test-time  self-supervised    strategy  on ImageNet-LT with various test class distributions.} 
	\label{table_expert_learning_performance} \vspace{-0.05in}
 \begin{center} 
 \begin{threeparttable}  
    \resizebox{0.99\textwidth}{!}{
 	\begin{tabular}{lccccccccc}\toprule  
        \multirow{4}{*}{Test Dist.}&\multicolumn{9}{c}{ImageNet-LT}\cr\cmidrule{2-10}
        &\multicolumn{4}{c}{Ours w/o test-time   strategy}&&\multicolumn{4}{c}{Ours w/ test-time   strategy}\cr\cmidrule{2-5}\cmidrule{7-10}
        & Many  & Med. & Few & All &&  Many  & Med. & Few & All \cr
        \midrule
         Forward-LT-50   & 65.6	 	&  55.7	& 44.1 &   {65.5}  && 70.0 	& 53.2	& 33.1 &  {69.4 (\red{+3.9})} \\
         Forward-LT-10  &  66.5	 	& 56.8 & 44.2	&  {63.6}  && 69.9	 	& 54.3	& 34.7 	&  {65.4 (\red{+1.8})} \\
         Uniform    &  67.0	 	& 56.7	& 42.6	& {58.8} && 66.5	 	& 57.0	& 43.5 	&  {58.8 (+0.0)} \\
         Backward-LT-10    & 65.0      & 57.6 	& 43.1 &  {53.1} && 60.9      & 57.5	 	& 50.1 &  {54.5 (\red{+1.4})}  \\ 
          Backward-LT-50 & 69.1       & 57.0	 	& 42.9 &	 {49.8} && 60.7      & 56.2	 	&  50.7 &  {53.1 (\red{+3.3})}\\ 
        \bottomrule

	\end{tabular}}
	 \end{threeparttable}
	 \end{center} 

    	\end{minipage} \vspace{-0.1in}
    \end{table}

\begin{table}[H]   
	\caption{Comparison among different test-time training strategies for handling class distribution shifts on ImageNet-LT with various unknown test class distributions.}\vspace{-0.1in} 
	\label{table_agnostic_imagenet_testime} 
 \begin{center}
 \begin{threeparttable} 
    \resizebox{0.95\textwidth}{!}{
	\begin{tabular}{lcccccccccccccc}\toprule  
   	     \multirow{2}{*}{Backbone} &\multirow{2}{*}{Test-time strategy}&   \multicolumn{5}{c}{Forward} && Uniform  && \multicolumn{5}{c}{Backward} \cr  \cmidrule{3-7} \cmidrule{9-9}\cmidrule{11-15} 
         && 50 &25 &10&5&2 && 1 && 2& 5& 10 &25 &50\cr  
        \midrule
         \multirow{5}{*}{SADE}   &  No test-time adaptation & 65.5 & 64.4 &63.6 &62.0 & 60.0 &&58.8 && 56.8 &54.7 &53.1 &51.5 &49.8  \\         
         &  Test-time pseudo-labeling & 67.1  & 66.1  & 64.7 & 63.0 & 60.1  & & 57.7 && 54.7 & 51.1 &  48.1  & 45.0 & 42.4 \\          
         & Test class distribution estimation~\cite{lipton2018detecting} & 69.1 & 66.6 & 63.7 & 60.5 & 56.5 &&  53.3 && 49.9  & 45.6 & 42.7 & 39.5 & 36.8 \\         
         & Entropy minimization with Tent~\cite{wang2021tent} & 68.0  & 67.0 &  \textbf{65.6}  & 62.8 &  60.5  & & 58.6  & & 56.0 & 53.2 & 50.6 & 48.1 &  45.7 \\    
     
          & Self-supervised expert aggregation (ours) & \textbf{69.4} &  \textbf{67.4}    & 65.4  	& \textbf{63.0}  & \textbf{60.6} && \textbf{58.8} && \textbf{57.1} & \textbf{55.5} & \textbf{54.5} & \textbf{53.7} & \textbf{53.1} \\    
 
        \bottomrule
    
	\end{tabular}}
	 \end{threeparttable}
	 \end{center} \vspace{-0.15in}
\end{table}

\subsection{Effectiveness of Test-time Self-supervised Aggregation}
\label{cls_testtime_learning}
This subsection evaluates our test-time self-supervised aggregation strategy.

\textbf{Effectiveness in expert aggregation.} As shown in Table~\ref{table_test_time}, our self-supervised   strategy    learns suitable expert   weights for  various unknown   test class distributions.  For   forward long-tailed   distributions, the weight of  the forward expert $E_1$ is higher; while for     backward long-tailed   ones,  the weight of  the  backward expert $E_3$ is relatively high. This enables our multi-expert model to boost the performance  on   dominant  classes  for   unknown test   distributions, leading to   better ensemble performance (cf.~Table~\ref{table_expert_learning_performance}), particularly as test data get  more skewed.    The      results on more datasets  are reported in Appendix~\ref{cls_testtime_learning_supp}, while more ablation studies of our strategy are shown in Appendix~\ref{App_F}.

\textbf{Superiority over test-time training methods.} We then verify the superiority of our self-supervised   strategy over existing test-time training approaches  on various test class distributions. Specifically, we adopt three non-trivial baselines: (i) \emph{Test-time pseudo-labeling}  uses the multi-expert  model to iteratively generate  pseudo  labels  for unlabeled test data and  uses them to fine-tune the    model; (ii) \emph{Test class distribution estimation}  leverages BBSE~\cite{lipton2018detecting} to estimate the test class distribution and   uses it to pose-adjust model predictions;
  (iii) \emph{Tent}~\cite{wang2021tent}  fine-tunes the batch normalization layers  of models through entropy minimization  on unlabeled test data. The results in Table~\ref{table_agnostic_imagenet_testime} show  that directly   applying  existing  test-time training methods cannot handle well the class distribution shifts, particularly on the inversely 
long-tailed class distribution. In comparison, our self-supervised  strategy is able to   aggregate multiple experts appropriately for the unknown test class distribution (cf. Table~\ref{table_test_time}), leading to  promising performance gains on various    test class  distributions (cf. Table~\ref{table_agnostic_imagenet_testime}).  

   \begin{wraptable}{r}{0.5\textwidth} 
  \vspace{-0.1in}
	\caption{The effectiveness of our   self-supervised aggregation strategy in dealing with (unknown) partial test class distributions on ImageNet-LT.} 
	\label{table_imagenet_partial}  
 \begin{center} 
 \begin{threeparttable}  
    \resizebox{0.49\textwidth}{!}{
 	\begin{tabular}{lcccc}\toprule  
        \multirow{2}{*}{Method}&\multicolumn{3}{c}{ImageNet-LT}\cr\cmidrule{2-5} 
        & Only many  & Only medium & Only few   \cr
        \midrule
        SADE w/o test-time  strategy   & 67.4	 	&  56.9	& 42.5      \\
      SADE w/ test-time   strategy  &  71.0  	 	& 57.2 & 53.6	   \\
      Accuracy gain  &   (\red{+3.6}) 	 	& (\red{+0.3})  &  (\red{+11.1})	   \\
        \bottomrule

	\end{tabular}}
	 \end{threeparttable}\vspace{-0.14in}
	 \end{center} 

\end{wraptable} 

\textbf{Effectiveness on partial class distributions.} Real-world test data  may follow any  type  of class distribution, including partial class distributions (\ie not all of the classes appear in the test data). Motivated by this, we further evaluate SADE on three partial  class distributions: only many-shot classes, only medium-shot classes, and only few-shot classes. The results in Table~\ref{table_imagenet_partial} demonstrate the effectiveness of  SADE in  tackling more complex test class distributions.

\section{Conclusion}
In this paper, we have  explored a practical yet challenging  task of \emph{test-agnostic long-tailed recognition}, where the test class distribution is unknown and not necessarily uniform. To tackle  this task, we present a novel  approach, namely \emph{Self-supervised Aggregation of Diverse Experts} (SADE), which consists of two innovative  strategies, \ie skill-diverse expert learning and test-time self-supervised   aggregation. We theoretically analyze    our proposed  method and also empirically show that   SADE achieves new  state-of-the-art performance on both vanilla  and test-agnostic long-tailed recognition.

\clearpage
\subsubsection*{Acknowledgments}
This work was partially supported by  NUS ODPRT Grant R252-000-A81-133 and NUS Advanced Research and Technology Innovation Centre (ARTIC) Project Reference (ECT-RP2).  We also gratefully appreciate the support of MindSpore, CANN (Compute Architecture for Neural Networks) and Ascend AI Processor used for this research.

{  
\bibliographystyle{plain}
\bibliography{sade}

\begin{thebibliography}{10}

\bibitem{boudiaf2020unifying}
Malik Boudiaf, J{\'e}r{\^o}me Rony, et~al.
\newblock A unifying mutual information view of metric learning: cross-entropy
  vs. pairwise losses.
\newblock In {\em European Conference on Computer Vision}, 2020.

\bibitem{cai2021ace}
Jiarui Cai, Yizhou Wang, and Jenq-Neng Hwang.
\newblock Ace: Ally complementary experts for solving long-tailed recognition
  in one-shot.
\newblock In {\em International Conference on Computer Vision}, 2021.

\bibitem{cao2019learning}
Kaidi Cao, Colin Wei, Adrien Gaidon, Nikos Arechiga, and Tengyu Ma.
\newblock Learning imbalanced datasets with label-distribution-aware margin
  loss.
\newblock In {\em Advances in Neural Information Processing Systems}, 2019.

\bibitem{chawla2002smote}
Nitesh~V Chawla, Kevin~W Bowyer, et~al.
\newblock Smote: synthetic minority over-sampling technique.
\newblock {\em Journal of Artificial Intelligence Research}, 2002.

\bibitem{chen2020improved}
Xinlei Chen, Haoqi Fan, Ross Girshick, and Kaiming He.
\newblock Improved baselines with momentum contrastive learning.
\newblock {\em arXiv preprint arXiv:2003.04297}, 2020.

\bibitem{chi2021test}
Zhixiang Chi, Yang Wang, et~al.
\newblock Test-time fast adaptation for dynamic scene deblurring via
  meta-auxiliary learning.
\newblock In {\em Computer Vision and Pattern Recognition}, 2021.

\bibitem{cubuk2020randaugment}
Ekin~Dogus Cubuk, Barret Zoph, Jon Shlens, and Quoc Le.
\newblock Randaugment: Practical automated data augmentation with a reduced
  search space.
\newblock In {\em Advances in Neural Information Processing Systems},
  volume~33, 2020.

\bibitem{cui2021parametric}
Jiequan Cui, Zhisheng Zhong, Shu Liu, Bei Yu, and Jiaya Jia.
\newblock Parametric contrastive learning.
\newblock In {\em International Conference on Computer Vision}, 2021.

\bibitem{cui2019class}
Yin Cui, Menglin Jia, et~al.
\newblock Class-balanced loss based on effective number of samples.
\newblock In {\em Computer Vision and Pattern Recognition}, 2019.

\bibitem{deng2021pml}
Zongyong Deng, Hao Liu, Yaoxing Wang, Chenyang Wang, Zekuan Yu, and Xuehong
  Sun.
\newblock Pml: Progressive margin loss for long-tailed age classification.
\newblock In {\em Computer Vision and Pattern Recognition}, pages 10503--10512,
  2021.

\bibitem{feng2021exploring}
Chengjian Feng, Yujie Zhong, and Weilin Huang.
\newblock Exploring classification equilibrium in long-tailed object detection.
\newblock In {\em International Conference on Computer Vision}, 2021.

\bibitem{grandvalet2005semi}
Yves Grandvalet, Yoshua Bengio, et~al.
\newblock Semi-supervised learning by entropy minimization.
\newblock In {\em CAP}, 2005.

\bibitem{guo2021long}
Hao Guo and Song Wang.
\newblock Long-tailed multi-label visual recognition by collaborative training
  on uniform and re-balanced samplings.
\newblock In {\em Computer Vision and Pattern Recognition}, 2021.

\bibitem{havasi2020training}
Marton Havasi, Rodolphe Jenatton, et~al.
\newblock Training independent subnetworks for robust prediction.
\newblock In {\em International Conference on Learning Representations}, 2021.

\bibitem{he2016deep}
Kaiming He, Xiangyu Zhang, Shaoqing Ren, and Jian Sun.
\newblock Deep residual learning for image recognition.
\newblock In {\em Computer Vision and Pattern Recognition}, pages 770--778,
  2016.

\bibitem{he2022relieving}
Yin-Yin He, Peizhen Zhang, Xiu-Shen Wei, Xiangyu Zhang, and Jian Sun.
\newblock Relieving long-tailed instance segmentation via pairwise class
  balance.
\newblock {\em arXiv preprint arXiv:2201.02784}, 2022.

\bibitem{hong2020disentangling}
Youngkyu Hong, Seungju Han, Kwanghee Choi, Seokjun Seo, Beomsu Kim, and Buru
  Chang.
\newblock Disentangling label distribution for long-tailed visual recognition.
\newblock In {\em Computer Vision and Pattern Recognition}, 2021.

\bibitem{huang2016learning}
Chen Huang, Yining Li, Chen~Change Loy, and Xiaoou Tang.
\newblock Learning deep representation for imbalanced classification.
\newblock In {\em Computer Vision and Pattern Recognition}, 2016.

\bibitem{iwasawa2021test}
Yusuke Iwasawa and Yutaka Matsuo.
\newblock Test-time classifier adjustment module for model-agnostic domain
  generalization.
\newblock In {\em Advances in Neural Information Processing Systems},
  volume~34, 2021.

\bibitem{jamal2020rethinking}
Muhammad~Abdullah Jamal, Matthew Brown, Ming-Hsuan Yang, Liqiang Wang, and
  Boqing Gong.
\newblock Rethinking class-balanced methods for long-tailed visual recognition
  from a domain adaptation perspective.
\newblock In {\em Computer Vision and Pattern Recognition}, pages 7610--7619,
  2020.

\bibitem{jiawei2020balanced}
Ren Jiawei, Cunjun Yu, Xiao Ma, Haiyu Zhao, Shuai Yi, et~al.
\newblock Balanced meta-softmax for long-tailed visual recognition.
\newblock In {\em Advances in Neural Information Processing Systems}, 2020.

\bibitem{johnson2019survey}
Justin~M Johnson and Taghi~M Khoshgoftaar.
\newblock Survey on deep learning with class imbalance.
\newblock {\em Journal of Big Data}, 6(1):1--54, 2019.

\bibitem{kamani2020targeted}
Mohammad~Mahdi Kamani, Sadegh Farhang, Mehrdad Mahdavi, and James~Z Wang.
\newblock Targeted data-driven regularization for out-of-distribution
  generalization.
\newblock In {\em ACM SIGKDD International Conference on Knowledge Discovery \&
  Data Mining}, pages 882--891, 2020.

\bibitem{kang2021exploring}
Bingyi Kang, Yu~Li, Sa~Xie, Zehuan Yuan, and Jiashi Feng.
\newblock Exploring balanced feature spaces for representation learning.
\newblock In {\em International Conference on Learning Representations}, 2021.

\bibitem{kang2019decoupling}
Bingyi Kang, Saining Xie, Marcus Rohrbach, Zhicheng Yan, Albert Gordo, Jiashi
  Feng, and Yannis Kalantidis.
\newblock Decoupling representation and classifier for long-tailed recognition.
\newblock In {\em International Conference on Learning Representations}, 2020.

\bibitem{kim2020learning}
Ildoo Kim, Younghoon Kim, and Sungwoong Kim.
\newblock Learning loss for test-time augmentation.
\newblock {\em Advances in Neural Information Processing Systems},
  33:4163--4174, 2020.

\bibitem{li2020overcoming}
Yu~Li, Tao Wang, Bingyi Kang, Sheng Tang, Chunfeng Wang, Jintao Li, and Jiashi
  Feng.
\newblock Overcoming classifier imbalance for long-tail object detection with
  balanced group softmax.
\newblock In {\em Computer Vision and Pattern Recognition}, 2020.

\bibitem{lin2022prototype}
Hongbin Lin, Yifan Zhang, Zhen Qiu, Shuaicheng Niu, Chuang Gan, Yanxia Liu, and
  Mingkui Tan.
\newblock Prototype-guided continual adaptation for class-incremental
  unsupervised domain adaptation.
\newblock In {\em European Conference on Computer Vision}, 2022.

\bibitem{lipton2018detecting}
Zachary Lipton, Yu-Xiang Wang, and Alexander Smola.
\newblock Detecting and correcting for label shift with black box predictors.
\newblock In {\em International conference on machine learning}, pages
  3122--3130. PMLR, 2018.

\bibitem{liu2021ttt++}
Yuejiang Liu, Parth Kothari, Bastien van Delft, Baptiste Bellot-Gurlet, Taylor
  Mordan, and Alexandre Alahi.
\newblock Ttt++: When does self-supervised test-time training fail or thrive?
\newblock In {\em Advances in Neural Information Processing Systems},
  volume~34, 2021.

\bibitem{liu2019large}
Ziwei Liu, Zhongqi Miao, Xiaohang Zhan, Jiayun Wang, Boqing Gong, and Stella~X
  Yu.
\newblock Large-scale long-tailed recognition in an open world.
\newblock In {\em Computer Vision and Pattern Recognition}, pages 2537--2546,
  2019.

\bibitem{long2014transfer}
Mingsheng Long, Jianmin Wang, Guiguang Ding, Jiaguang Sun, and Philip~S Yu.
\newblock Transfer joint matching for unsupervised domain adaptation.
\newblock In {\em Computer Vision and Pattern Recognition}, pages 1410--1417,
  2014.

\bibitem{menon2020long}
Aditya~Krishna Menon, Sadeep Jayasumana, Ankit~Singh Rawat, Himanshu Jain,
  Andreas Veit, and Sanjiv Kumar.
\newblock Long-tail learning via logit adjustment.
\newblock In {\em International Conference on Learning Representations}, 2021.

\bibitem{niu2022efficient}
Shuaicheng Niu, Jiaxiang Wu, Yifan Zhang, Yaofo Chen, Shijian Zheng, Peilin
  Zhao, and Mingkui Tan.
\newblock Efficient test-time model adaptation without forgetting.
\newblock In {\em International conference on machine learning}, 2022.

\bibitem{pandey2021generalization}
Prashant Pandey, Mrigank Raman, Sumanth Varambally, and Prathosh AP.
\newblock Generalization on unseen domains via inference-time label-preserving
  target projections.
\newblock In {\em Computer Vision and Pattern Recognition}, pages 12924--12933,
  2021.

\bibitem{Influence2021Park}
Seulki Park, Jongin Lim, Younghan Jeon, and Jin~Young Choi.
\newblock Influence-balanced loss for imbalanced visual classification.
\newblock In {\em International Conference on Computer Vision}, 2021.

\bibitem{peng2021optimal}
Hanyu Peng, Mingming Sun, and Ping Li.
\newblock Optimal transport for long-tailed recognition with learnable cost
  matrix.
\newblock In {\em International Conference on Learning Representations}, 2022.

\bibitem{qiu2021source}
Zhen Qiu, Yifan Zhang, Hongbin Lin, Shuaicheng Niu, Yanxia Liu, Qing Du, and
  Mingkui Tan.
\newblock Source-free domain adaptation via avatar prototype generation and
  adaptation.
\newblock In {\em International Joint Conference on Artificial Intelligence},
  2021.

\bibitem{sharma2020long}
Saurabh Sharma, Ning Yu, Mario Fritz, and Bernt Schiele.
\newblock Long-tailed recognition using class-balanced experts.
\newblock In {\em German Conference on Pattern Recognition}, pages 86--100,
  2020.

\bibitem{sun2020test}
Yu~Sun, Xiaolong Wang, Zhuang Liu, John Miller, Alexei Efros, and Moritz Hardt.
\newblock Test-time training with self-supervision for generalization under
  distribution shifts.
\newblock In {\em International Conference on Machine Learning}, 2020.

\bibitem{tan2020equalization}
Jingru Tan, Changbao Wang, Buyu Li, Quanquan Li, Wanli Ouyang, Changqing Yin,
  and Junjie Yan.
\newblock Equalization loss for long-tailed object recognition.
\newblock In {\em Computer Vision and Pattern Recognition}, pages 11662--11671,
  2020.

\bibitem{tang2020long}
Kaihua Tang, Jianqiang Huang, and Hanwang Zhang.
\newblock Long-tailed classification by keeping the good and removing the bad
  momentum causal effect.
\newblock In {\em Advances in Neural Information Processing Systems},
  volume~33, 2020.

\bibitem{tian2020posterior}
Junjiao Tian, Yen-Cheng Liu, et~al.
\newblock Posterior re-calibration for imbalanced datasets.
\newblock In {\em Advances in Neural Information Processing Systems}, 2020.

\bibitem{van2018inaturalist}
Grant Van~Horn, Oisinand Mac~Aodha, et~al.
\newblock The inaturalist species classification and detection dataset.
\newblock In {\em Computer Vision and Pattern Recognition}, 2018.

\bibitem{varsavsky2020test}
Thomas Varsavsky, Mauricio Orbes-Arteaga, et~al.
\newblock Test-time unsupervised domain adaptation.
\newblock In {\em International Conference on Medical Image Computing and
  Computer-Assisted Intervention}, pages 428--436, 2020.

\bibitem{wang2021tent}
Dequan Wang, Evan Shelhamer, Shaoteng Liu, Bruno Olshausen, and Trevor Darrell.
\newblock Tent: Fully test-time adaptation by entropy minimization.
\newblock In {\em International Conference on Learning Representations}, 2021.

\bibitem{wang2021seesaw}
Jiaqi Wang, Wenwei Zhang, Yuhang Zang, Yuhang Cao, Jiangmiao Pang, Tao Gong,
  Kai Chen, Ziwei Liu, Chen~Change Loy, and Dahua Lin.
\newblock Seesaw loss for long-tailed instance segmentation.
\newblock In {\em Computer Vision and Pattern Recognition}, pages 9695--9704,
  2021.

\bibitem{wang2021contrastive}
Peng Wang, Kai Han, et~al.
\newblock Contrastive learning based hybrid networks for long-tailed image
  classification.
\newblock In {\em Computer Vision and Pattern Recognition}, 2021.

\bibitem{wang2020long}
Xudong Wang, Long Lian, Zhongqi Miao, Ziwei Liu, and Stella~X Yu.
\newblock Long-tailed recognition by routing diverse distribution-aware
  experts.
\newblock In {\em International Conference on Learning Representations}, 2021.

\bibitem{wen2016discriminative}
Yandong Wen, Kaipeng Zhang, et~al.
\newblock A discriminative feature learning approach for deep face recognition.
\newblock In {\em European Conference on Computer Vision}, 2016.

\bibitem{wen2019batchensemble}
Yeming Wen, Dustin Tran, and Jimmy Ba.
\newblock Batchensemble: an alternative approach to efficient ensemble and
  lifelong learning.
\newblock In {\em International Conference on Learning Representations}, 2020.

\bibitem{weng2021unsupervised}
Zhenzhen Weng, Mehmet~Giray Ogut, Shai Limonchik, and Serena Yeung.
\newblock Unsupervised discovery of the long-tail in instance segmentation
  using hierarchical self-supervision.
\newblock In {\em Computer Vision and Pattern Recognition}, 2021.

\bibitem{xiang2020learning}
Liuyu Xiang, Guiguang Ding, and Jungong Han.
\newblock Learning from multiple experts: Self-paced knowledge distillation for
  long-tailed classification.
\newblock In {\em European Conference on Computer Vision}, 2020.

\bibitem{yosinski2014transferable}
Jason Yosinski, Jeff Clune, Yoshua Bengio, and Hod Lipson.
\newblock How transferable are features in deep neural networks?
\newblock In {\em Advances in Neural Information Processing Systems},
  volume~27, pages 3320--3328, 2014.

\bibitem{zang2021fasa}
Yuhang Zang, Chen Huang, and Chen~Change Loy.
\newblock Fasa: Feature augmentation and sampling adaptation for long-tailed
  instance segmentation.
\newblock In {\em International Conference on Computer Vision}, 2021.

\bibitem{zhang2021distribution}
Songyang Zhang, Zeming Li, Shipeng Yan, Xuming He, and Jian Sun.
\newblock Distribution alignment: A unified framework for long-tail visual
  recognition.
\newblock In {\em Computer Vision and Pattern Recognition}, pages 2361--2370,
  2021.

\bibitem{Zhang2021UnleashingTP}
Yifan Zhang, Bryan Hooi, Lanqing Hong, and Jiashi Feng.
\newblock Unleashing the power of contrastive self-supervised visual models via
  contrast-regularized fine-tuning.
\newblock In {\em Advances in Neural Information Processing Systems}, 2021.

\bibitem{zhang2021deep}
Yifan Zhang, Bingyi Kang, Bryan Hooi, Shuicheng Yan, and Jiashi Feng.
\newblock Deep long-tailed learning: A survey.
\newblock {\em arXiv preprint arXiv:2110.04596}, 2021.

\bibitem{zhang2020collaborative}
Yifan Zhang, Ying Wei, et~al.
\newblock Collaborative unsupervised domain adaptation for medical image
  diagnosis.
\newblock {\em IEEE Transactions on Image Processing}, 2020.

\bibitem{zhang2018online}
Yifan Zhang, Peilin Zhao, Jiezhang Cao, Wenye Ma, Junzhou Huang, Qingyao Wu,
  and Mingkui Tan.
\newblock Online adaptive asymmetric active learning for budgeted imbalanced
  data.
\newblock In {\em ACM SIGKDD International Conference on Knowledge Discovery \&
  Data Mining}, pages 2768--2777, 2018.

\bibitem{zhao2018adaptive}
Peilin Zhao, Yifan Zhang, Min Wu, Steven~CH Hoi, Mingkui Tan, and Junzhou
  Huang.
\newblock Adaptive cost-sensitive online classification.
\newblock {\em IEEE Transactions on Knowledge and Data Engineering},
  31(2):214--228, 2018.

\bibitem{zhong2021improving}
Zhisheng Zhong, Jiequan Cui, Shu Liu, and Jiaya Jia.
\newblock Improving calibration for long-tailed recognition.
\newblock In {\em Computer Vision and Pattern Recognition}, 2021.

\bibitem{zhou2020bbn}
Boyan Zhou, Quan Cui, Xiu-Shen Wei, and Zhao-Min Chen.
\newblock Bbn: Bilateral-branch network with cumulative learning for
  long-tailed visual recognition.
\newblock In {\em Computer Vision and Pattern Recognition}, pages 9719--9728,
  2020.

\end{thebibliography}
}  
 
\clearpage 
\section*{Checklist}

\begin{enumerate}

\item For all authors...
\begin{enumerate}
  \item Do the main claims made in the abstract and introduction accurately reflect the paper's contributions and scope?
        \answerYes{}
  \item Did you describe the limitations of your work?
    \answerYes{Please refer to Appendix~\ref{App_H}.}
  \item Did you discuss any potential negative societal impacts of your work?
     \answerNA{This is a fundamental research that does not have particular negative social impacts.}
  \item Have you read the ethics review guidelines and ensured that your paper conforms to them?
      \answerYes{}
\end{enumerate}

\item If you are including theoretical results...
\begin{enumerate}
  \item Did you state the full set of assumptions of all theoretical results?
   \answerYes{}
        \item Did you include complete proofs of all theoretical results?
     \answerYes{Please refer to Appendix~\ref{App_A}.}
\end{enumerate}

\item If you ran experiments...
\begin{enumerate}
  \item Did you include the code, data, and instructions needed to reproduce the main experimental results (either in the supplemental material or as a URL)?
     \answerYes{Please refer to the supplemental material.}
  \item Did you specify all the training details (e.g., data splits, hyperparameters, how they were chosen)?
     \answerYes{Please refer to Section~\ref{exp_setup} and Appendix~\ref{App_C}.}
        \item Did you report error bars (e.g., with respect to the random seed after running experiments multiple times)?
    \answerNA{The common practice in long-tailed recognition does not report error bars, so we follow the previous papers and do not report them.}
        \item Did you include the total amount of compute and the type of resources used (e.g., type of GPUs, internal cluster, or cloud provider)?
        \answerYes{Please refer to  Appendix~\ref{App_C3} for details on different datasets.}
\end{enumerate}

\item If you are using existing assets (e.g., code, data, models) or curating/releasing new assets...
\begin{enumerate}
  \item If your work uses existing assets, did you cite the creators?
       \answerYes{}
  \item Did you mention the license of the assets?
   \answerNA{All the used benchmark datasets are publicly available.}
  \item Did you include any new assets either in the supplemental material or as a URL?
     \answerYes{We submitted the source codes of our method as an anonymized zip file.}
  \item Did you discuss whether and how consent was obtained from people whose data you're using/curating?
      \answerNA{These datasets are open-source benchmark datasets.}
  \item Did you discuss whether the data you are using/curating contains personally identifiable information or offensive content?
       \answerNA{These datasets are open-source benchmark datasets.}
\end{enumerate}

\item If you used crowdsourcing or conducted research with human subjects...
\begin{enumerate}
  \item Did you include the full text of instructions given to participants and screenshots, if applicable?
   \answerNA{}
  \item Did you describe any potential participant risks, with links to Institutional Review Board (IRB) approvals, if applicable?
   \answerNA{}
  \item Did you include the estimated hourly wage paid to participants and the total amount spent on participant compensation?
   \answerNA{}
\end{enumerate}

\end{enumerate}

\clearpage

\appendix 

\begin{table}
	\setlength{\tabcolsep}{0.2cm}
	\begin{tabular}{p{0.97\columnwidth}}
		\nipstophline 
		\vspace{-2pt}
		\centering
		\textbf{\Large{Supplementary Materials}}
		\vspace{-5pt}
		\nipsbottomhline   
	\end{tabular}
\end{table}\vspace{-5pt}

We organize the supplementary materials as follows:

$\bullet$ Appendix~\ref{App_A}: the proofs for Theorem~\ref{thm1}.

$\bullet$ Appendix~\ref{App_B}: the pseudo-code of the proposed method. 

$\bullet$ Appendix~\ref{App_C}: more details of experimental settings. 

$\bullet$ Appendix~\ref{App_D}:    more empirical results on vanilla long-tailed recognition, test-agnostic long-tailed recognition, skill-diverse expert learning, and test-time self-supervised aggregation.

$\bullet$ Appendix~\ref{App_E}:    more ablation studies   on  expert learning and the proposed inverse softmax loss. 
  
$\bullet$ Appendix~\ref{App_F}:    more ablation studies   on  test-time self-supervised aggregation.  

$\bullet$ Appendix~\ref{App_G}:    more discussion   on  model complexity.  

$\bullet$ Appendix~\ref{App_H}:    discussion on potential limitations.  
 
\section{Proofs for Theorem~\ref{thm1}}\label{App_A}

\begin{proof}
We first recall several  key notations   and   define some new notations. The random variables of model predictions  and ground-truth labels are defined as $\hat{Y} \sim p(\hat{y})$  and $Y \sim p(y)$, respectively. The  number of classes is denoted by $C$. Moreover, we further denote the test sample set of the  class $k$ by $\mathcal{Z}_k$, in which the total number of samples in this class is denoted by $|\mathcal{Z}_k|$.   
Let $c_k  = \frac{1}{|\mathcal{Z}_k|}\sum_{\hat{y}\in\mathcal{Z}_k}\hat{y}$ represent the hard mean of all  predictions of samples from the class $k$,  and let  $\small{\overset{c}{=}}$ indicate  equality up to a multiplicative and/or additive constant. 
 
As shown in Eq.(\ref{sta_loss}), the optimization objective of our test-time self-supervised aggregation method is to maximize  $\mathcal{S}= \sum_{j=1}^{n_t} \hat{y}_j^1 \small{\cdot} \hat{y}_j^2$, where $n_t$ denotes the number of test samples. For convenience, we simplify the first data view to be the original data, so the objective function becomes $\sum_{j=1}^{n_t} \hat{y}_{j} \small{\cdot}\hat{y}_{j}^1$.
Maximizing such an objective is equivalent to minimizing $ \sum_{j=1}^{n_t} -\hat{y}_{j} \small{\cdot}\hat{y}_{j}^1$. Here, we assume the data augmentations are strong enough to generate representative data views that can simulate  the test data from the same class. In this sense, the new data  view can be regarded  as an independent sample from the same class.  
Following this, we    analyze our method   by connecting $- \hat{y}_{j} \small{\cdot}\hat{y}_{j}^1$ to $\sum_{\hat{y}_{j}\in \mathcal{Z}_k}\|\hat{y}_{j}\small{-}c_k\|^2$, which is similar to the tightness term  in the center loss~\cite{wen2016discriminative}:
\begin{align} 
   \sum_{\hat{y}_{j},\hat{y}_{j}^1\in \mathcal{Z}_k}   - \hat{y}_{j}\small{\cdot}\hat{y}_{j}^1  
   & \overset{c}{=}  \frac{1}{|\mathcal{Z}_k|} \sum_{\hat{y}_{j},\hat{y}_{j}^1\in \mathcal{Z}_k}  - \hat{y}_{j}\small{\cdot}\hat{y}_{j}^1  ~~ \overset{c}{=} ~~ \frac{1}{|\mathcal{Z}_k|} \sum_{\hat{y}_{j},\hat{y}_{j}^1\in \mathcal{Z}_k}  \|\hat{y}_{j}\|^2 - \hat{y}_{j}\small{\cdot}\hat{y}_{j}^1   \nonumber \\
     &=  \sum_{\hat{y}_{j}\in \mathcal{Z}_k} \|\hat{y}_{j}\|^2 -  \frac{1}{|\mathcal{Z}_k|} \sum_{\hat{y}_{j}\in \mathcal{Z}_k} \sum_{\hat{y}_{j}^1\in \mathcal{Z}_k}    \hat{y}_{j}\small{\cdot}\hat{y}_{j}^1 \nonumber \\
    &  = \sum_{\hat{y}_{j}\in \mathcal{Z}_k} \|\hat{y}_{j}\|^2 - 2 \frac{1}{|\mathcal{Z}_k|} \sum_{\hat{y}_{j}\in \mathcal{Z}_k} \sum_{\hat{y}_{j}^1\in \mathcal{Z}_k}    \hat{y}_{j}\small{\cdot}\hat{y}_{j}^1   +   \frac{1}{|\mathcal{Z}_k|} \sum_{\hat{y}_{j}\in \mathcal{Z}_k} \sum_{\hat{y}_{j}^1\in \mathcal{Z}_k}    \hat{y}_{j}\small{\cdot}\hat{y}_{j}^1 \nonumber \\
     &=  \sum_{\hat{y}_{j}\in \mathcal{Z}_k} \|\hat{y}_{j}\|^2 - 2 \hat{y}_{j}\small{\cdot}c_k  +   \|c_k\|^2\nonumber \\
      &=\sum_{\hat{y}_{j}\in \mathcal{Z}_k}\|\hat{y}_{j}\small{-}c_k\|^2,  \nonumber  
\end{align}
where we use the property of the normalized predictions, \ie $\|\hat{y}_{j}\|^2 = \|\hat{y}_{j}^1\|^2=1$, and the definition of the class hard mean $c_k= \frac{1}{|\mathcal{Z}_k|}\sum_{\hat{y}\in\mathcal{Z}_k}\hat{y}$.

By summing over all classes $k$, we obtain:
\begin{align}
   \sum_{j=1}^{n_t} - \hat{y}_{j}\small{\cdot}\hat{y}_{j}^1 ~~ {\overset{c}{=}} ~~ \sum_{j=1}^{n_t} \|\hat{y}_{j}\small{-}c_{y_i}\|^2. \nonumber
\end{align}

Based on this equation, following~\cite{boudiaf2020unifying,Zhang2021UnleashingTP}, we can interpret $\sum_{j=1}^{n_t}- \hat{y}_{j}\small{\cdot}\hat{y}_{j}^1$ as a conditional cross-entropy between $\hat{Y}$ and another random variable $\bar{Y}$, whose conditional distribution given $Y$ is a standard Gaussian centered around $c_Y\small{:}\bar{Y}|Y\small{\sim}\mathcal{N}(c_y,i)$:
\begin{align}
    \sum_{j=1}^{n_t} - \hat{y}_{j}\small{\cdot}\hat{y}_{j}^1    \overset{c}{=}  \mathcal{H}(\hat{Y};\bar{Y}|Y) = \mathcal{H}(\hat{Y}|Y) \small{+}\mathcal{D}_{KL}(\hat{Y}||\bar{Y}|Y). \nonumber
\end{align}
Hence, we know that $\sum_{j=1}^{n_t} - \hat{y}_{j}\small{\cdot}\hat{y}_{j}^1$ is an upper bound on the conditional entropy of predictions $\hat{Y}$ given labels $Y$:
\begin{align}
 \sum_{j=1}^{n_t} - \hat{y}_{j}\small{\cdot}\hat{y}_{j}^1   \overset{c}{\geq} \mathcal{H}(\hat{Y}|Y),   \nonumber
\end{align} 
where the symbol $\small{\overset{c}{\geq}}$ represents  ``larger than" up to a multiplicative and/or an additive constant. Moreover, when $\hat{Y}|Y\small{\sim}\mathcal{N}(c_y,i)$, the bound is tight.  As a result, minimizing $\sum_{j=1}^{n_t} - \hat{y}_{j}\small{\cdot}\hat{y}_{j}^1$  is equivalent  to minimizing $\mathcal{H}(\hat{Y}|Y)$:
\begin{align}\label{eq_first_term}
  \sum_{j=1}^{n_t} - \hat{y}_{j}\small{\cdot}\hat{y}_{j}^1  \propto \mathcal{H}(\hat{Y}|Y).   
\end{align}

Meanwhile,  the mutual information between  predictions $\hat{Y}$ and  labels $Y$ can be represented by:
\begin{align}\label{eq_theorem_proof1}
    \mathcal{I}(\hat{Y};Y)=  \mathcal{H}(\hat{Y})-\mathcal{H}(\hat{Y}|Y).
\end{align}

Combining Eqs.(\ref{eq_first_term}-\ref{eq_theorem_proof1}), we have:
\begin{align} 
  \sum_{j=1}^{n_t} - \hat{y}_{j}\small{\cdot}\hat{y}_{j}^1  \propto - \mathcal{I}(\hat{Y};Y) + \mathcal{H}(\hat{Y}).   \nonumber
\end{align}

Since $\mathcal{S} = \sum_{j=1}^{n_t} \hat{y}_{j}\small{\cdot}\hat{y}_{j}^1$, we obtain:
\begin{align} 
  \mathcal{S} \propto  \mathcal{I}(\hat{Y};Y) - \mathcal{H}(\hat{Y}),   \nonumber
\end{align}
which concludes the proof for  Theorem~\ref{thm1}.
\end{proof}
 
\clearpage
\section{Pseudo-code}\label{App_B}
This appendix provides the pseudo-code\footnote{The source code is  provided in the supplementary material.} of  SADE, which consists of  skill-diverse expert learning and test-time  self-supervised aggregation. Here, the skill-diverse expert learning strategy is summarized in Algorithm~\ref{algo_expert}. For simplicity, we depict the pseudo-code based on batch size 1, but we conduct batch gradient descent in practice.

\begin{algorithm} 
\caption{Skill-diverse Expert Learning}
\label{algo_expert} 
\begin{algorithmic}[1]
\small
\Require{Epochs $T$;  Hyper-parameters $\lambda$ for $\mathcal{L}_{inv}$ }
\Ensure{Network backbone $f_{\theta}$;  Experts  $E_1, E_2, E_3$ } 
\For{e=1,...,$T$} 
\For{$x \in \mathcal{D}_s$}  ~ // batch sampling in practice 
\State Obtain logits $v_1$ based on $f_{\theta}$ and $E_1$;    
\State Obtain logits $v_2$ based on $f_{\theta}$ and $E_2$;    
\State Obtain logits $v_3$ based on $f_{\theta}$ and $E_3$;    
\State Compute   loss $\mathcal{L}_{ce}$ with $v_1$ for Expert $E_1$; ~ // Eq.(\ref{softmax_loss})
\State Compute loss $\mathcal{L}_{bal}$ with $v_2$ for Expert $E_2$; ~ // Eq.(\ref{balanced_loss})
\State Compute loss $\mathcal{L}_{inv}$ with $v_3$ for Expert $E_3$; ~ // Eq.(\ref{inverse_loss})
\State Train the model with $\mathcal{L}_{ce}+\mathcal{L}_{bal}+\mathcal{L}_{inv}$.
\EndFor
\EndFor 

\hspace{-0.5in}\return ~ The trained model  $\{f_{\theta},E_1, E_2, E_3\}$ 
\end{algorithmic} 
\end{algorithm}

After training the multiple skill-diverse experts with Algorithm~\ref{algo_expert},  the final prediction of the multi-expert model for vanilla long-tailed recognition is the arithmetic mean  of the prediction logits of these experts, followed by a softmax function. 

When it comes to test-agnostic long-tailed recognition, we need to aggregate these skill-diverse experts to handle the unknown test class distribution based on Algorithm~\ref{algo_SADE}.
Here, to avoid  the learned weights of some weak experts becoming zero, we give a stopping condition in Algorithm~\ref{algo_SADE}: if the weight for one expert is less than 0.05, we stop test-time training. Retaining a small amount of weight for each expert is sufficient to ensure the effect of  ensemble learning.  
 
\begin{algorithm}
\caption{Test-time Self-supervised Aggregation}
\label{algo_SADE} 
\begin{algorithmic}[1]
\small
\Require{Epochs $T'$; The trained backbone $f_{\theta}$; The trained experts  $E_1, E_2, E_3$ }
\Ensure{Expert aggregation weights $w$ // uniform initialization} 
\For{e=1,...,$T'$} 
\For{$x \in \mathcal{D}_t$}  ~ // batch sampling in practice 
\State Draw two data augmentation functions $t\small{\sim}\mathcal{T}$, $t'\small{\sim}\mathcal{T}$;
\State Generate data views $x^{1}\small{=}t(x)$, $x^{2}\small{=}t'(x)$;
\State Obtain logits $v^1_1$,$v^1_2$,$v^1_3$ for the view  $x^{1}$;
\State Obtain logits $v^2_1$,$v^2_2$,$v^2_3$ for the view $x^{2}$;
\State Normalize expert weights $w$ via softmax function;
\State Conduct predictions $\hat{y}^{1}$,$\hat{y}^{2}$ based on $\hat{y}\small{=}wv$;  
\State Compute prediction stability   $\mathcal{S}$; ~ // Eq.~(\ref{sta_loss})
\State Maximize  $\mathcal{S}$ to update $w$;
\EndFor
\State If $w_i\leq 0.05$ for any $w_i\in w$, then stop training. 
\EndFor 

\hspace{-0.5in}\return ~ Expert aggregation weights $w$ 
\end{algorithmic} 
\end{algorithm}

Note that,  in test-agnostic long-tailed recognition,  each model is only trained once on long-tailed training data and then directly evaluated on multiple test sets.
Our test-time self-supervised strategy  adapts the trained multi-expert model using only unlabeled test data during testing.

\clearpage
\section{More Experimental Settings}\label{App_C}
In this appendix, we provide more details on experimental settings.

\subsection{Benchmark Datasets} \label{App_C1}
We use four benchmark  datasets (\ie ImageNet-LT~\cite{liu2019large}, CIFAR100-LT~\cite{cao2019learning}, Places-LT~\cite{liu2019large}, and iNaturalist 2018~\cite{van2018inaturalist}) to simulate real-world long-tailed class distributions. These datasets suffer from severe class imbalance~\cite{johnson2019survey,zhang2018online}.Their data  statistics are summarized   in Table~\ref{dataset}, where CIFAR100-LT has three variants with different imbalance ratios. The imbalance ratio is defined as $\max{n_j}$/$\min{n_j}$, where $n_j$ denotes the data number of class $j$. 
\begin{table}[h]   
     \caption{Statistics of datasets.}\label{dataset}   \vspace{-0.1in}
    \begin{center} 
    \scalebox{0.8}{  
    \begin{threeparttable} 
	\begin{tabular}{lcccc}\toprule
        Dataset & $\#$ classes &  $\#$ training data & $\#$ test data &   imbalance ratio  \\ \midrule    
        ImageNet-LT~\cite{liu2019large} 	& 1,000 & 115,846 &50,000 	& 256  \\
         
        CIFAR100-LT~\cite{cao2019learning} & 100 & 50,000 & 10,000 & \{10,50,100\}\\
        Places-LT~\cite{liu2019large} & 365 & 62,500 & 36,500 & 996 \\
        iNaturalist 2018~\cite{van2018inaturalist} & 8,142 & 437,513 & 24,426 & 500 \\
        \bottomrule
	\end{tabular}  
    \end{threeparttable}}
    \end{center}   \vspace{-0.1in} 
\end{table}
 
\subsection{Construction of Test-agnostic Long-tailed Datasets} \label{App_C2}
Following LADE~\cite{hong2020disentangling},  we construct three kinds of test class distributions, \ie the uniform  distribution, forward long-tailed distributions and backward long-tailed distributions. In the backward ones, the long-tailed class order is flipped.  Here, the forward and backward  long-tailed test distributions contain multiple different imbalance ratios, \ie $\rho \in \{2,5,10,25,50\}$. Note that LADE~\cite{hong2020disentangling} only constructed multiple  distribution-agnostic test datasets  for ImageNet-LT; while in this study, we use the same way to construct distribution-agnostic test datasets  for the remaining benchmark datasets, \ie CIFAR100-LT, Places-LT and iNaturalist 2018, as illustrated below.

Considering the long-tailed training classes are sorted in a decreasing order, the various test datasets are constructed as follows: (1) Forward long-tailed distribution: the number of the $j$-th class is $n_j=N\cdot \rho^{(j-1)/C}$, where $N$ indicates the sample number  per class in the original  uniform test dataset and $C$ is the number of classes.  (2) Backward long-tailed distribution: the number of the $j$-th class is $n_j=N\cdot \rho^{(C-j)/C}$.  In the backward long-tailed distributions,  the  order of  the long tail on classes is flipped, so the distribution shift between training and test data is large, especially when    the imbalance ratio gets higher.

For ImageNet-LT, CIFAR100-LT and Places-LT, since there are enough test samples per class,  we follow the setting in LADE~\cite{hong2020disentangling} and construct the imbalance ratio set by $\rho \in \{2,5,10,25,50\}$.  For    iNaturalist 2018,  since each class only contains three test samples,   we adjust   the imbalance ratio set to $\rho \in \{2,3\}$. Note that when we set $\rho=3$, there are some classes  in iNaturalist 2018 containing no test sample. All   these constructed distribution-agnostic long-tailed datasets will be publicly available along with our code.

\subsection{More Implementation  Details of Our Method} \label{App_C3}
We implement our   method  in PyTorch. Following~\cite{hong2020disentangling,wang2020long}, we use ResNeXt-50 for ImageNet-LT, ResNet-32 for CIFAR100-LT, ResNet-152 for Places-LT and ResNet-50 for iNaturalist 2018 as  backbones, respectively. Moreover,   we adopt the cosine classifier for prediction on all datasets. 

Although we have depicted the skill-diverse multi-expert framework in Section~\ref{expert_learning},   we give more details about it here.
Without loss of generality, we take ResNet~\cite{he2016deep} as an example to illustrate the  multi-expert model.   Since the shallow layers  extract more general features and deeper layers extract more task-specific features~\cite{yosinski2014transferable}, the three-expert model  uses the first two stages of ResNet as the expert-shared backbone, while the later stages of ResNet and the fully-connected layer constitute independent components  of each expert. 
To be more specific, the number of convolutional  filters in each expert is reduced by 1/4, since by sharing the backbone and using fewer filters in each expert~\cite{wang2020long, zhou2020bbn}, the computational complexity of the model is reduced compared to the model with     independent experts.  The final prediction is the arithmetic mean  of the prediction logits of these experts, followed by a softmax function. 

In the training phase,  the data augmentations are the same as previous long-tailed studies~\cite{hong2020disentangling,kang2019decoupling}. If not specified, we use the SGD optimizer with the momentum of 0.9 and set the initial learning rate as 0.1 with linear  decay. More specifically, for ImageNet-LT, we train models for 180 epochs with batch size 64 and a learning rate of 0.025 (cosine  decay). For CIFAR100-LT,  the  training epoch is 200 and the batch size is 128. For Places-LT, following~\cite{liu2019large}, we use  ImageNet pre-trained ResNet-152     as the backbone, while the batch size is set to 128 and the training epoch  is 30. Besides, the learning rate  is   0.01 for the classifier and   0.001 for all other layers. For iNaturalist 2018, we set   the  training epoch to 200, the batch size to 512 and the learning rate to 0.2.
In our inverse softmax loss, we set $\lambda\small{=}2$ for ImageNet-LT and CIFAR100-LT, and   $\lambda\small{=}1$  for the remaining    datasets. 

In the test-time training, we use the same augmentations as MoCo v2~\cite{chen2020improved} to generate different data views, \ie random resized crop, color jitter, gray scale, Gaussian  blur and horizontal flip. If not specified, we train the aggregation weights for 5 epochs with the batch size 128, where we adopt  the same optimizer and learning rate as the training phase.  

More detailed statistics of network architectures and  hyper-parameters are reported in Table~\ref{statistic_parameter}. 
Based on these hyper-parameters, we conduct   experiments on 1  TITAN RTX 2080  GPU for CIFAR100-LT, 4 GPUs for iNaturalist18, and 2 GPUs for ImageNet-LT and Places-LT, respectively. The source code of our method is available in the supplementary material.  
  
  \begin{table*}[h]    
	\caption{Statistics of the used network architectures and hyper-parameters in our study.}  
    \label{statistic_parameter}        
    \begin{center}
     \scalebox{0.85}{  
    \begin{threeparttable} 
	\begin{tabular}{l|c|c|c|c}\hline
        Items  &    ImageNet-LT &  CIFAR100LT & Places-LT &  iNarutalist 2018 \\ \hline\hline
         \multicolumn{5}{c}{Network Architectures} \\\hline\hline
        network backbone & ResNeXt-50 & ResNet-32 & ResNet-152 & ResNet-50  \\  \hline 
        classifier &   \multicolumn{4}{c}{cosine classifier}\\     \hline\hline
          \multicolumn{5}{c}{Training Phase} \\\hline\hline
        epochs  & 180 & 200 & 30 &  200 \\     \hline
         batch size     & 64  & 128 & 128  & 512   \\  \hline   
        learning rate (lr)      &  0.025 & 0.1 &0.01 &  0.2  \\  \hline
        lr schedule &       cosine decay& \multicolumn{3}{c}{linear decay}\\ \hline 
        $\lambda$ in inverse softmax loss    & \multicolumn{2}{c|}{2} &   \multicolumn{2}{c}{1}    \\ \hline
        weight decay factor     & $5 \times 10^{-4}$  & $5 \times 10^{-4}$  & $4 \times 10^{-4}$  & $2 \times 10^{-4}$\\  \hline
          momentum factor &     \multicolumn{4}{c}{0.9}\\  \hline
        optimizer &   \multicolumn{4}{c}{SGD optimizer with nesterov}\\ \hline
        \hline 
          \multicolumn{5}{c}{Test-time Training} \\\hline\hline
        epochs  & \multicolumn{4}{c}{5} \\     \hline
         batch size     &  \multicolumn{4}{c}{128}  \\  \hline   
        learning rate (lr)      &  0.025 & 0.1 &0.01 &  0.1  \\  \hline
	\end{tabular} 
    \end{threeparttable}} 
    \end{center}     
\end{table*}

\subsection{Discussions on Evaluation Metric} \label{App_C4}
As mentioned in Section~\ref{exp_setup}, we follow LADE~\cite{hong2020disentangling} and use micro accuracy to evaluate model performance on test-agnostic long-tailed recognition. In this appendix, we   explain why  micro accuracy is a better metric than  macro accuracy when the test dataset exhibits a non-uniform class distribution. 
For instance, in  the test scenario with a backward long-tailed class     distribution, the tail classes are more frequently encountered than the head classes, and thus should have larger weights in evaluation.  However, simply using macro accuracy treats all the categories equally and cannot differentiate classes of different frequencies. 

For example,  one may train a  recognition model for autonomous cars based on the training data collected from city areas, where pedestrians are majority classes and stone obstacles are minority classes. Assume the model accuracy is 60\%  on pedestrians and 40\%  on stones. If deploying the model to   city areas, where pedestrians/stones are assumed to have 500/50 test data,  then the macro accuracy is  50\% and the micro accuracy is $\frac{500\times 0.6+50\times 0.4}{500+50}\small{\approx}58\%$. In contrast,  when deploying the model to mountain areas, the pedestrians become the minority, while stones become the majority. 
Assuming the test data numbers are changed to 50/500 on pedestrians/stones, the micro accuracy is adjusted  to $\frac{50\times 0.6+500\times 0.4}{50+500}\small{\approx}42\%$, but the macro accuracy is still 50\%. In this case,   macro accuracy is less informative than micro accuracy for measuring model performance. Therefore, micro accuracy  is a better metric to evaluate the performance of test-agnostic long-tailed recognition.

\clearpage
\section{More Empirical Results}\label{App_D}
 
\subsection{More Results on Vanilla Long-tailed Recognition}\label{App_D3}
\textbf{Accuracy on class subsets}~~~ In the main paper, we have provided the average performance over all classes on the uniform test class distribution. In this appendix, we further report the accuracy  regarding various class subsets (c.f.~ Table~\ref{table_uniform_supp}),   making the results more complete.

\begin{table*}[h]   
      \vspace{0.1in}
	\caption{Top-1 accuracy of long-tailed recognition methods on the uniform test distribution. }
	\label{table_uniform_supp} 
 \begin{center}
 \begin{threeparttable} 
    \resizebox{0.95\textwidth}{!}{
 	\begin{tabular}{lcccccccccccccc}\toprule 
 	
        \multirow{2}{*}{Method}
        &\multicolumn{4}{c}{ImageNet-LT}&&\multicolumn{4}{c}{CIFAR100-LT(IR10)} &&\multicolumn{4}{c}{CIFAR100-LT(IR50)} \cr\cmidrule{2-5}\cmidrule{7-10}\cmidrule{12-15}
        & Many  & Med. & Few & All &&  Many  & Med. & Few & All &&  Many  & Med. & Few & All\cr
        \midrule
        Softmax  & \textbf{68.1} 	&  41.5	& 14.0  & 48.0  && \textbf{66.0}& 42.7& -& 59.1 & & \textbf{66.8} & 37.4 & 15.5 & 45.6   \\
 
        Causal~\cite{tang2020long}  & 64.1 	 	&  45.8 & 27.2  & 50.3 && 63.3& 49.9& - & 59.4 && 62.9 & 44.9 & 26.2 & 48.8   \\
        Balanced Softmax~\cite{jiawei2020balanced}    &  64.1	 	& 48.2	& 33.4	& 52.3 && 63.4 & 55.7 & - & 61.0 && 62.1& 45.6&36.7 &50.9  \\
       
        MiSLAS~\cite{zhong2021improving}   & 62.0     & 49.1	& 32.8 & 51.4 && 64.9 & 56.6 & - & 62.5 &&  61.8 & 48.9 & 33.9 & 51.5 \\ 
        LADE~\cite{hong2020disentangling} & 64.4     & 47.7	 	& 34.3 & 52.3 && 63.8 & 56.0 & - & 61.6 &&  60.2 & 46.2 &35.6 & 50.1 \\
        RIDE~\cite{wang2020long} & 68.0       & 52.9	 	& 35.1 & 56.3   && 65.7 & 53.3 & -& 61.8 &&  66.6 & 46.2 & 30.3 & 51.7 \\\midrule
          SADE (ours)  & 66.5      & \textbf{57.0}	 	& \textbf{43.5} & \textbf{58.8} && 65.8 & \textbf{58.8} & -& \textbf{63.6} && 61.5 & \textbf{50.2} & \textbf{45.0} & \textbf{53.9}\\
         
         \midrule \midrule
        \multirow{2}{*}{Method}
        &\multicolumn{4}{c}{CIFAR100-LT(IR100)}&&\multicolumn{4}{c}{Places-LT} &&\multicolumn{4}{c}{iNaturalist 2018} \cr\cmidrule{2-5}\cmidrule{7-10}\cmidrule{12-15}
        & Many  & Med. & Few & All &&  Many  & Med. & Few & All &&  Many  & Med. & Few & All\cr
        \midrule
        Softmax  &   \textbf{68.6} &41.1 & 9.6 & 41.4 && \textbf{46.2} & 27.5 & 12.7& 31.4 && \textbf{74.7} & 66.3 & 60.0 & 64.7  \\
 
        Causal~\cite{tang2020long}  & 64.1& 46.8& 19.9& 45.0  &&   23.8 & 35.7 & \textbf{39.8} & 32.2 &&  71.0 & 66.7 & 59.7 & 64.4 \\
        Balanced Softmax~\cite{jiawei2020balanced}    & 59.5 & 45.4 & \textbf{30.7} & 46.1 &&42.6 & 39.8& 32.7&39.4 &&  70.9 & 70.7 & 70.4 & 70.6    \\
       
        MiSLAS~\cite{zhong2021improving}   & 60.4 & \textbf{49.6} & 26.6 & 46.8&&   41.6 & 39.3 & 27.5 & 37.6 &&  71.7 & 71.5 & 69.7 & 70.7 \\ 
        LADE~\cite{hong2020disentangling} &  58.7 & 45.8& 29.8& 45.6 && 42.6 & 39.4 & 32.3 & 39.2 && 68.9 & 68.7 & 70.2 &69.3 \\
        RIDE~\cite{wang2020long} & 67.4 & 49.5 & 23.7 & 48.0 &&   43.1 & 41.0 & 33.0 & 40.3 && 71.5 & 70.0 & 71.6 & 71.8 \\\midrule
          SADE (ours)   & 65.4 & 49.3  & 29.3 & \textbf{49.8} && 40.4 & \textbf{43.2} &36.8 & \textbf{40.9} && 74.5 & \textbf{72.5} & \textbf{73.0} & \textbf{72.9} \\ 
        \bottomrule

	\end{tabular}}
	 \end{threeparttable} 
	 \end{center}
	
      \vspace{0.1in}
\end{table*}

\textbf{Results on stonger data augmentations}~~~  Inspired by PaCo~\cite{cui2021parametric}, we further evaluate SADE training with stronger data augmentation  (\ie RandAugment~\cite{cubuk2020randaugment}) for 400   epochs. The results in Table~\ref{table_augmentation} further demonstrate the state-of-the-art performance of SADE.

\begin{table*}[h]  
      \vspace{0.1in}
     \caption{Accuracy of long-tailed methods with stronger augmentations, where the test class distribution is uniform. Here, $^{*}$ denotes training with RandAugment~\cite{cubuk2020randaugment} for 400   epochs. The baseline results    are directly copied from the work~\cite{cui2021parametric}.}  
     \label{table_augmentation} 
 \begin{center} 
 \begin{threeparttable} 
    \resizebox{0.98\textwidth}{!}{
 	\begin{tabular}{ccccccc}\toprule  
       Methods&   ImageNet-LT  &   CIFAR100-LT(IR10)  &  CIFAR100-LT(IR50) &  CIFAR100-LT(IR100) & Places-LT  &  iNaturalist 2018  \cr
        \midrule
        PaCo$^{*}$~\cite{cui2021parametric} 	 & 58.2 & 64.2 & 56.0 & 52.0 & 41.2 & 73.2	\\
         SADE$^{*}$ (ours)   & \textbf{61.2} &  \textbf{65.3} & \textbf{57.3} &  \textbf{52.2} & \textbf{41.3} & \textbf{74.5} \\
        \bottomrule

	\end{tabular}} 
	 \end{threeparttable}
	 \end{center}  
      \vspace{0.1in}
\end{table*}

\textbf{Results on more neural architectures}~~~ In addition to using the common practice of backbones as previous long-tailed studies~\cite{hong2020disentangling,wang2020long}, we further evaluate   SADE on more  neural architectures.   The results in Table~\ref{table_backbones}  demonstrate that SADE is able to train different   network backbones well.

\begin{table}[h]
  
      \vspace{0.1in}
      \caption{Accuracy of SADE with various network architectures.  Here, $^{*}$ denotes training with RandAugment~\cite{cubuk2020randaugment} for 400   epochs.}  \vspace{-0.1in}
     \label{table_backbones} 
 \begin{center} 
 \begin{threeparttable} 
    \resizebox{0.95\textwidth}{!}{
 		\begin{tabular}{lcccccclccccc}\toprule 
  	\multicolumn{6}{c}{ImageNet-LT}&&\multicolumn{6}{c}{iNaturalist 2018}  \cr\cmidrule{1-6}\cmidrule{8-13} 
       \multirow{1}{*}{Backbone}   & Methods & Many  & Med. & Few & All &&  \multirow{1}{*}{Backbone}   & Methods & Many  & Med. & Few & All \cr  
        \midrule
        \multirow{2}{*}{ResNeXt-50} & SADE & 66.5      & {57.0}	 	&  {43.5} &  {58.8} &&  \multirow{2}{*}{ResNet-50} & SADE & 74.5 & {72.5} & {73.0} & {72.9}  \\
        & SADE$^{*}$ & 67.3     & 60.4	 	&  46.4 &  61.2 && &  SADE$^{*}$ & 75.5     & 73.7	&  75.1 &  74.5 \\   \cmidrule{1-6}  \cmidrule{8-13} 
         \multirow{2}{*}{ResNeXt-101} & SADE &  66.8 	&  57.5 	& 43.1  & 59.1 & &\multirow{2}{*}{ResNet-152} &  SADE  &  76.2 	& 64.3 	& 65.1 & 74.8 \\
        & SADE$^{*}$ &  68.1	&  60.5 	& 45.5 & 61.4 &&&  SADE$^{*}$ &  \textbf{78.3}	&   \textbf{77.0}	&  \textbf{76.7} &  \textbf{77.0}\\   \cmidrule{1-6}  \cmidrule{8-13} 
        \multirow{2}{*}{ResNeXt-152}  & SADE & 67.2 		&  57.4	& 43.5  & 59.3 &&  \\
        & SADE$^{*}$ & \textbf{68.6} 		&  \textbf{61.2}	& \textbf{47.0} & \textbf{62.1} && \\
         
        \bottomrule

	\end{tabular}}
	 \end{threeparttable}
	 \end{center} 

      \vspace{0.1in}
\end{table}

\newpage
\textbf{Results on more datasets}~~~ We also conduct experiments on CIFAR10-LT with imbalance ratios of 10 and 100.  Promising results in Table~\ref{Result_cifar} further demonstrate the effectiveness and superiority of our proposed method.

\begin{table}[h]        \vspace{0.1in}
      \caption{Accuracy on CIFAR10-LT, where the test class distribution is uniform. Most results are directly copied from the work~\cite{zhong2021improving}.}  
      \label{Result_cifar}  
 \begin{center} 
 \begin{threeparttable} 
    \resizebox{0.7\textwidth}{!}{
 	\begin{tabular}{cccccc}\toprule  
       Imbalance Ratio&   Softmax  &   BBN   & MiSLAS & RIDE & SADE (ours)   \cr
        \midrule
        10 	 &86.4 & 88.4 & 90.0 & 89.7& \textbf{90.8} 	\\
        100 &  70.4 &  79.9& 82.1 & 81.6 & \textbf{83.8}\\
        \bottomrule

	\end{tabular}} 
	 \end{threeparttable}
	 \end{center}   
\end{table} 

\newpage
\subsection{More Results  on Test-agnostic   Long-tailed Recognition}\label{App_D4}
In the main paper, we have  provided the overall performance on four benchmark datasets with various   test class distributions (cf. Table~\ref{table_agnostic_imagenet}). In this appendix, we further verify the effectiveness of our method on     more dataset settings (\ie CIFAR100-IR10 and CIFAR100-IR50), as shown in Table~\ref{table_agnostic_imagenet_full}.

\begin{table}[h] \vspace{0.1in}
	\caption{Top-1 accuracy over all classes on various unknown test class distributions. ``Prior" indicates that the test class distribution is used as prior knowledge.  ``Uni." denotes the uniform distribution. ``IR" indicates the imbalance ratio. ``BS" denotes the balanced softmax~\cite{jiawei2020balanced}.}
	\label{table_agnostic_imagenet_full} 
 \begin{center}
 \begin{threeparttable} 
    \resizebox{1\textwidth}{!}{
 	\begin{tabular}{lcccccccccccccccccccccccccccc}\toprule  
   	     \multirow{4}{*}{Method} &\multirow{4}{*}{Prior}&  \multicolumn{13}{c}{(a) ImageNet-LT} && \multicolumn{13}{c}{(b) CIFAR100-LT  (IR10)}\cr \cmidrule{3-15}  \cmidrule{17-29}  
       && \multicolumn{5}{c}{Forward-LT} && Uni.  && \multicolumn{5}{c}{Backward-LT} && \multicolumn{5}{c}{Forward-LT} && Uni. && \multicolumn{5}{c}{Backward-LT} \cr  \cmidrule{3-7} \cmidrule{9-9}\cmidrule{11-15} \cmidrule{17-21} \cmidrule{23-23}\cmidrule{25-29} 
         && 50 &25 &10&5&2 && 1 && 2& 5& 10 &25 &50& &50 &25 &10&5&2 && 1 && 2& 5& 10 &25 &50 \cr  
        \midrule
        Softmax  & \xmark & 66.1 & 63.8 & 	  60.3& 56.6&52.0 && 48.0 && 43.9 & 38.6 & 34.9 & 30.9 & 27.6 && \textbf{72.0} & 69.6& 66.4 & 65.0 & 61.2 && 59.1 &&56.3 & 53.5 & 50.5 & 48.7& 46.5\\  
        BS    & \xmark & 63.2	 	& 61.9	& 59.5	& 57.2 & 54.4&&52.3 && 50.0 & 47.0 & 45.0 & 42.3 & 40.8 && 65.9 & 64.9 & 64.1 & 63.4 & 61.8 &&61.0 && 60.0 & 58.2& 57.5& 56.2 & 55.1  \\ 
        MiSLAS    & \xmark& 61.6 & 60.4 & 58.0 & 56.3 & 53.7 && 51.4 && 49.2 & 46.1& 44.0 & 41.5 & 39.5 && 67.0 & 66.1 & 65.5 & 64.4 & 63.2 && 62.5 && 61.2 & 60.4 & 59.3 & 58.5 & 57.7 \\ 
        LADE  &\xmark & 63.4 & 62.1 & 59.9 & 57.4 & 54.6 && 52.3 && 49.9 & 46.8 & 44.9 & 42.7 &40.7 && 67.5 & 65.8 &65.8 & 64.4 & 62.7 && 61.6 && 60.5 & 58.8 & 58.3 & 57.4 & 57.7 \\
        LADE  & \cmark&   65.8 & 63.8 & 60.6 & 57.5 & 54.5 && 52.3 &&   50.4 & 48.8 & 48.6 & 49.0 &49.2 &&  71.2  & 69.3& 67.1 & 64.6 & 62.4 && 61.6 && 60.4 & 61.4 &61.5 & \textbf{62.7} & \textbf{64.8}\\
        RIDE  & \xmark &67.6 & 66.3      & 64.0 & 61.7 & 58.9 && 56.3 && 54.0 & 51.0 & 48.7 & 46.2 & 44.0&& 67.1 & 65.3 & 63.6 & 62.1 & 60.9 && 61.8  &&   58.4 & 56.8 & 55.3 & 54.9 &53.4 \\\midrule
        SADE  & \xmark & \textbf{69.4} &  \textbf{67.4}    & \textbf{65.4} 	& \textbf{63.0}  & \textbf{60.6} && \textbf{58.8} && \textbf{57.1} & \textbf{55.5} & \textbf{54.5} & \textbf{53.7} & \textbf{53.1} && 71.2  & \textbf{69.4} & \textbf{67.6} & \textbf{66.3} & \textbf{64.4} && \textbf{63.6} && \textbf{62.9} & \textbf{62.4} & \textbf{61.7} & 62.1 & 63.0\\    
        \midrule\midrule

    \multirow{4}{*}{Method} &\multirow{4}{*}{Prior}&  \multicolumn{13}{c}{(c) CIFAR100-LT  (IR50)} && \multicolumn{13}{c}{(d) CIFAR100-LT  (IR100)}\cr \cmidrule{3-15}  \cmidrule{17-29}  
       && \multicolumn{5}{c}{Forward-LT} && Uni. && \multicolumn{5}{c}{Backward-LT} && \multicolumn{5}{c}{Forward-LT} && Uni. && \multicolumn{5}{c}{Backward-LT} \cr  \cmidrule{3-7} \cmidrule{9-9}\cmidrule{11-15} \cmidrule{17-21} \cmidrule{23-23}\cmidrule{25-29} 
         && 50 &25 &10&5&2 && 1 && 2& 5& 10 &25 &50& &50 &25 &10&5&2 && 1 && 2& 5& 10 &25 &50 \cr  
        \midrule
        Softmax  & \xmark & 64.8 & 62.7 & 58.5 & 55.0 & 49.9 &&45.6&&40.9 & 36.2 & 32.1 & 26.6 & 24.6 &&63.3&62.0&56.2 & 52.5 & 46.4 && 41.4 && 36.5 & 30.5 & 25.8 & 21.7 & 17.5  \\  
        BS    & \xmark & 61.6 & 60.2 & 58.4 & 55.9 & 53.7 && 50.9 && 48.5 & 45.7 & 43.9 & 42.5 & 40.6 && 57.8 & 55.5 & 54.2 & 52.0 & 48.7 && 46.1 && 43.6 & 40.8 & 38.4 & 36.3 & 33.7  \\ 
        MiSLAS  & \xmark&  60.1 & 58.9 & 57.7 & 56.2 & 53.7 && 51.5 &&  48.7 & 46.5 & 44.3   & 41.8 & 40.2 && 58.8 & 57.2 & 55.2 & 53.0 & 49.6 && 46.8 && 43.6 & 40.1 & 37.7 & 33.9 & 32.1 \\ 
        LADE  &\xmark & 61.3 & 60.2 & 56.9 & 54.3 & 52.3 && 50.1 && 47.8 & 45.7 & 44.0 & 41.8 & 40.5 && 56.0 & 55.5 &52.8 &  51.0 & 48.0 && 45.6 && 43.2 & 40.0 & 38.3 & 35.5 & 34.0  \\
        LADE  & \cmark& 65.9 & 62.1 & 58.8 & 56.0 & 52.3 && 50.1 && 48.3 & 45.5 & 46.5 & 46.8 & 47.8 && 62.6 & 60.2 & 55.6 & 52.7 & 48.2 && 45.6 && 43.8 & 41.1 & 41.5 & 40.7 & 41.6 \\
        RIDE & \xmark &  62.2 & 61.0 & 58.8 & 56.4 & 52.9 && 51.7 && 47.1 & 44.0 & 41.4 & 38.7 & 37.1 && 63.0 & 59.9 & 57.0 & 53.6 & 49.4 && 48.0 && 42.5 & 38.1 & 35.4 & 31.6 & 29.2  \\\midrule
       SADE    & \xmark &  \textbf{67.2} & \textbf{64.5} & \textbf{61.2} & \textbf{58.6} & \textbf{55.4} && \textbf{53.9} && \textbf{51.9} & \textbf{50.9} & \textbf{51.0} & \textbf{51.7} & \textbf{52.8} && \textbf{65.9} & \textbf{62.5} & \textbf{58.3} & \textbf{54.8} & \textbf{51.1} && \textbf{49.8} && \textbf{46.2} & \textbf{44.7} & \textbf{43.9} & \textbf{42.5} & \textbf{42.4}\\    
        \midrule\midrule

       \multirow{4}{*}{Method} &\multirow{4}{*}{Prior}&  \multicolumn{13}{c}{(e) Places-LT} && \multicolumn{13}{c}{(f) iNaturalist 2018}\cr \cmidrule{3-15}  \cmidrule{17-29}  
       && \multicolumn{5}{c}{Forward-LT} && Uni. && \multicolumn{5}{c}{Backward-LT} && \multicolumn{5}{c}{Forward-LT} && Uni. && \multicolumn{5}{c}{Backward-LT} \cr  \cmidrule{3-7} \cmidrule{9-9}\cmidrule{11-15} \cmidrule{17-21} \cmidrule{23-23}\cmidrule{25-29} 
         && 50 &25 &10&5&2 && 1 && 2& 5& 10 &25 &50& &  &3 & &2&  && 1 &&  &2&   &3 &  \cr  
        \midrule
        Softmax  & \xmark & 45.6 & 42.7 & 40.2 & 38.0 & 34.1 &&  31.4 && 28.4 & 25.4 & 23.4 & 20.8 & 19.4 &&  &65.4 &&    65.5 &&&64.7 &&&64.0 &&63.4 &  \\  
        BS   & \xmark & 42.7 & 41.7 & 41.3 & 41.0 & 40.0 & & 39.4 && 38.5 & 37.8 & 37.1 & 36.2 & 35.6 &&      &70.3 &&    70.5 &&& 70.6 &&& 70.6 &&70.8 &\\ 
        MiSLAS   & \xmark&  40.9 & 39.7 & 39.5 & 39.6 & 38.8 && 38.3 && 37.3 & 36.7 & 35.8 & 34.7 & 34.4 &&  & 70.8 &&   70.8 &&&70.7 &&& 70.7 &&70.2 & \\ 
        LADE &\xmark & 42.8 & 41.5 & 41.2 & 40.8 & 39.8 && 39.2 && 38.1 & 37.6 & 36.9 & 36.0 & 35.7 &&  & 68.4 &&   69.0 &&& 69.3 &&& 69.6 && 69.5 & \\
        LADE & \cmark&   46.3  & 44.2  & 42.2 & 41.2 & 39.7 && 39.4 && 39.2 & 39.9 &  40.9  & \textbf{42.4} & \textbf{43.0}  && & \xmark &&   69.1 &&& 69.3 &&& 70.2 && \xmark  &\\
        RIDE  & \xmark & 43.1 & 41.8 & 41.6 & 42.0 & 41.0 && 40.3 && 39.6 & 38.7 & 38.2 & 37.0 & 36.9 &&& 71.5 &&   71.9 &&& 71.8 &&& 71.9 && 71.8  & \\\midrule
         SADE   & \xmark &  \textbf{46.4} & \textbf{44.9} & \textbf{43.3} & \textbf{42.6} & \textbf{41.3} & & \textbf{40.9} && \textbf{40.6} & \textbf{41.1} & \textbf{41.4} & 42.0 & 41.6 &&     &\textbf{72.3}  &&  \textbf{72.5} &&& \textbf{72.9}  &&& \textbf{73.5} && \textbf{73.3} & \\    
 
        \bottomrule
    
	\end{tabular}}
	 \end{threeparttable}
	 \end{center}  

\vspace{0.1in}
\end{table}

\begin{figure*}  
  \begin{minipage}{0.48\linewidth}
   \centerline{\includegraphics[width=7.5cm, clip, trim={1.5cm 7.5cm 1cm 7.5cm}]{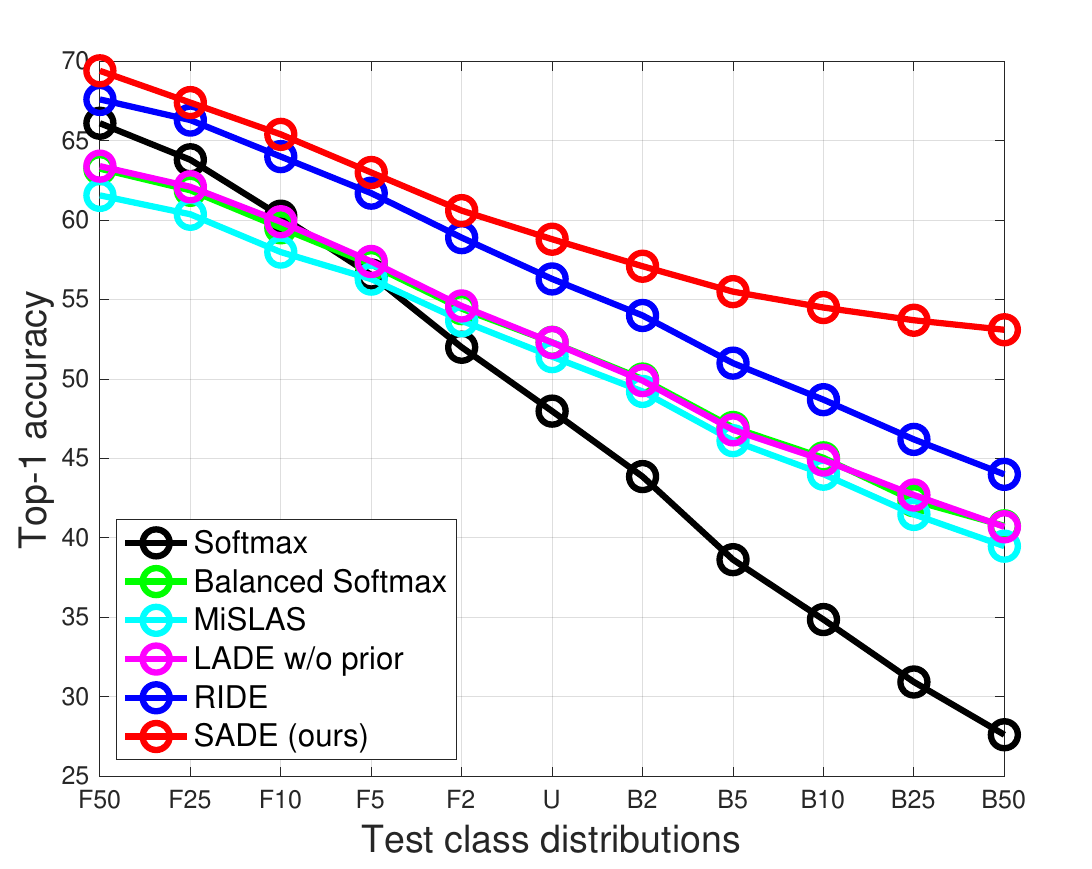}}
   \centerline{\small{(a) ImageNet-LT}}
  \end{minipage}
  \hfill
\begin{minipage}{0.48\linewidth}
   \centerline{\includegraphics[width=7.5cm, clip, trim={1.5cm 7.5cm 1cm 7.5cm}]{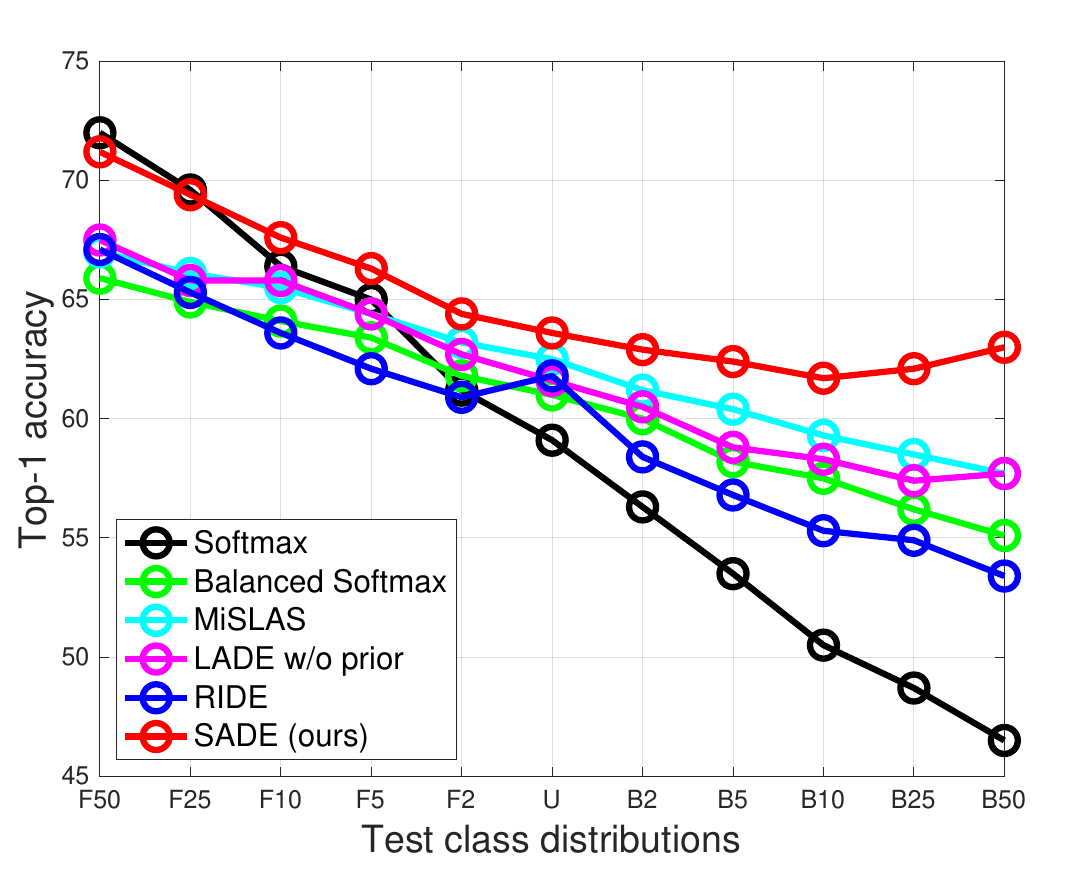}}
   \centerline{\small{(b) CIFAR100-LT(IR10)}}
  \end{minipage}
  \vfill
  \begin{minipage}{0.48\linewidth}
   \centerline{\includegraphics[width=7.5cm, clip, trim={1.5cm 7.5cm 1cm 7.5cm}]{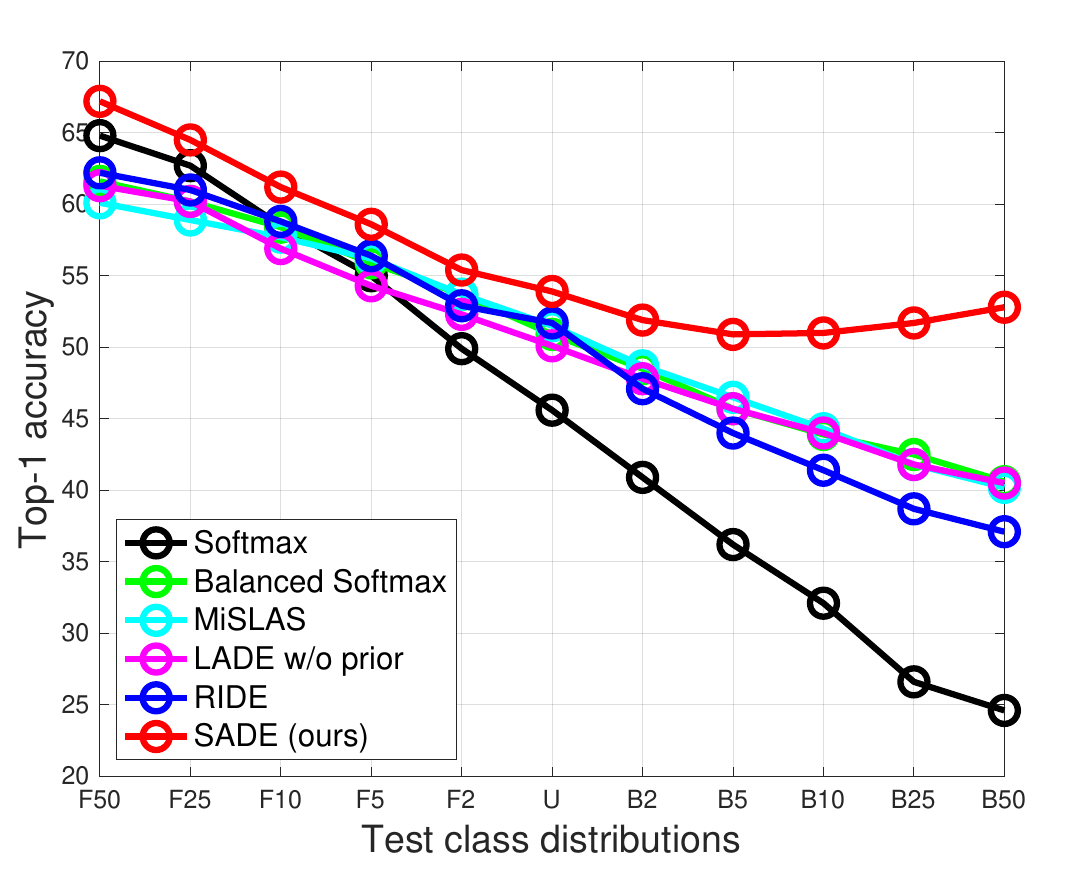}}
   \centerline{\small{(c) CIFAR100-LT(IR50)}}
  \end{minipage}
  \hfill    
  \begin{minipage}{0.48\linewidth}
   \centerline{\includegraphics[width=7.5cm, clip, trim={1.5cm 7.5cm 1cm 7.5cm}]{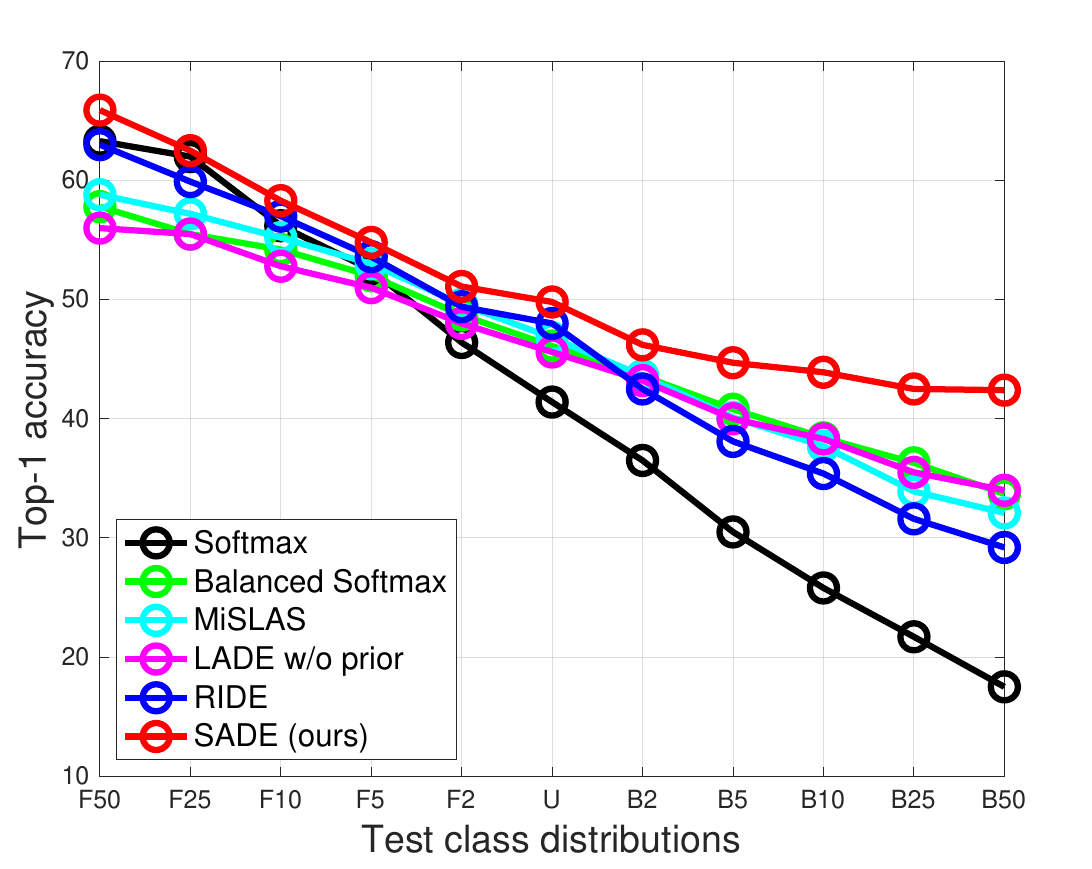}}
   \centerline{\small{(d) CIFAR100-LT(IR100)}}
  \end{minipage}
  \vfill
  \begin{minipage}{0.48\linewidth}
   \centerline{\includegraphics[width=7.5cm, clip, trim={1.5cm 7.5cm 1cm 7.5cm}]{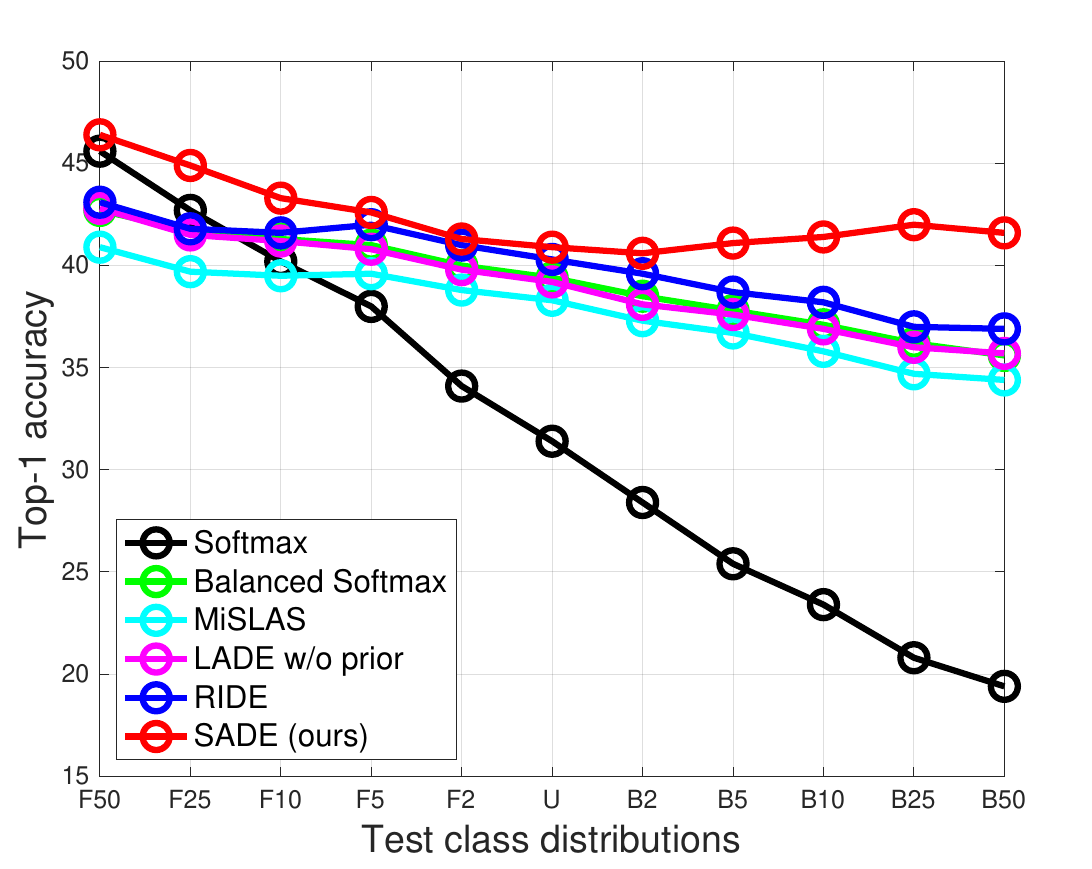}} 
   \centerline{\small{(e) Places-LT}}
  \end{minipage}
    \hfill    
  \begin{minipage}{0.48\linewidth}
   \centerline{\includegraphics[width=7.7cm, clip, trim={1cm 7.5cm 1cm 7.5cm}]{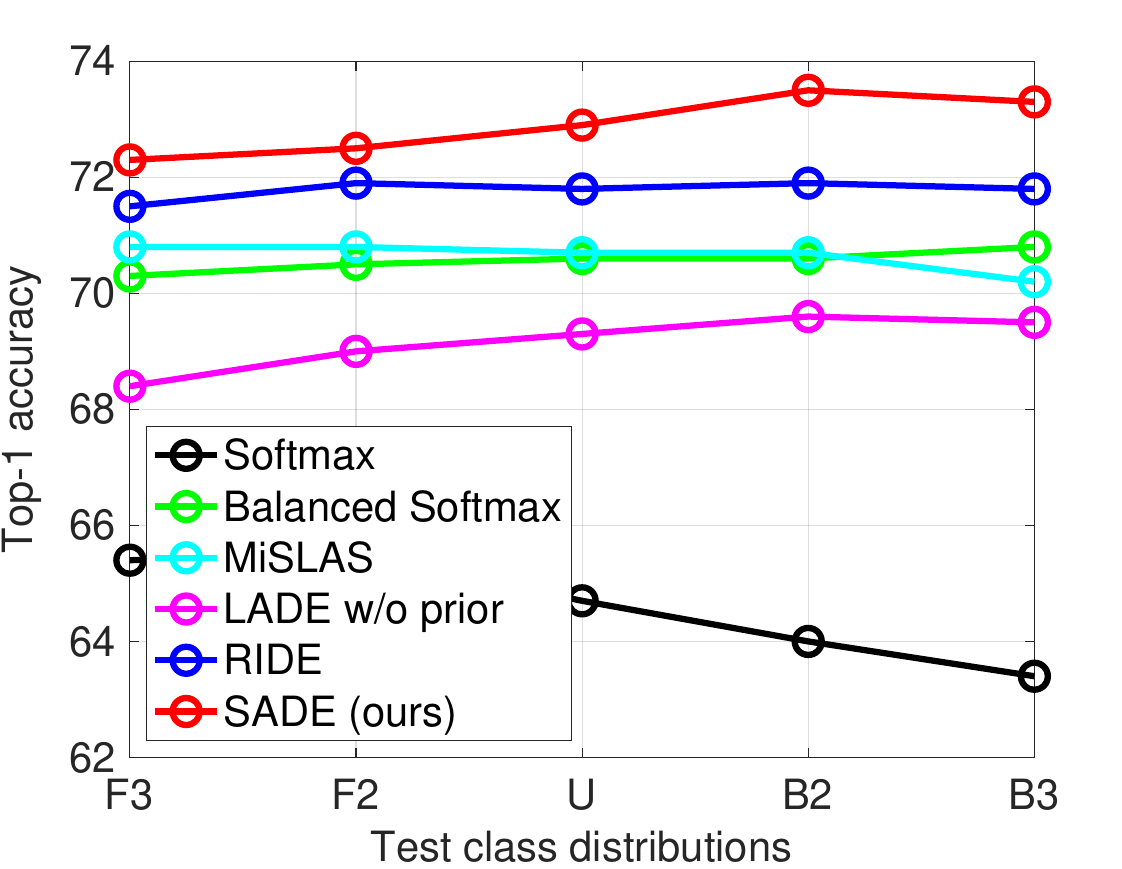}}
   \centerline{\small{(f) iNaturalist 2018}}
  \end{minipage}  
 \caption{Performance visualizations  on various unknown test class distributions, where ``F" indicates the forward long-tailed distributions as training data, ``B" indicates the backward long-tailed distributions to the training data, and ``U" denotes the uniform distribution.}\label{visualizations}  
\end{figure*}

Furthermore, we   plot the results of all methods under these benchmark datasets with various test class distributions in Figure~\ref{visualizations}.  To be specific, Softmax only performs well on   highly-imbalanced forward long-tailed class distributions. Existing   long-tailed   baselines outperform Softmax, but they cannot handle   backward test class distributions well.
In contrast, our  method consistently outperforms baselines on all benchmark datasets, particularly on the backward long-tailed test distributions with a relatively large imbalance ratio.

\clearpage
\subsection{More Results on Skill-diverse Expert Learning}
\label{cls_expert_learning_supp}

This appendix further evaluates the skill-diverse expert learning strategy on CIFAR100-LT, Places-LT and iNaturalist 2018 datasets. We report the results  in Table~\ref{table_expert_learning_supp}, from which we draw the following observations.
 RIDE~\cite{wang2020long} is one of the state-of-the-art ensemble-based long-tailed methods, which  tries to learn diverse distribution-aware experts by maximizing the divergence among expert predictions. However, such a method cannot  learn sufficiently diverse experts. As shown in Table~\ref{table_expert_learning_supp},   the three experts in RIDE perform very similarly on various groups of classes under all benchmark datasets,  and each expert has   similar overall performance  on each     dataset. Such results demonstrate that simply maximizing the KL divergence of different experts' predictions is not sufficient to learn visibly diverse  distribution-aware experts.

In contrast, our proposed method learns the skill-diverse experts by directly training each expert with their customized expertise-guided objective functions, respectively. To be specific, the forward expert $E_1$  seeks to learn the long-tailed  training distribution, so we directly train it with the cross-entropy loss. For the uniform expert $E_2$, we use the balanced softmax loss to simulate the uniform test distribution. For the backward expert $E_3$, we design a novel inverse softmax loss to train the expert, so that it   simulates the inversely long-tailed class distribution.  Table~\ref{table_expert_learning_supp} shows that the three experts trained by our method are visibly diverse  and skilled in handling different class distributions. Specifically, the forward expert is skilled in many-shot classes, the uniform expert is more balanced with higher overall  performance, and the backward expert is good at few-shot classes.  Because of such a novel design that enhances expert diversity, our method  achieves more promising ensemble performance compared to RIDE.

\begin{table*}[h] 
	\caption{Performance of each expert on the uniform test distribution. Here, the training imbalance ratio  of CIFAR100-LT is 100. The results show that our proposed method learns  more skill-diverse experts,  leading to better performance of ensemble aggregation.}
	\label{table_expert_learning_supp} 
 \begin{center}
 \begin{threeparttable} 
    \resizebox{0.99\textwidth}{!}{
 	\begin{tabular}{lccccccccccccccccccc}\toprule 
 	
        \multirow{4}{*}{Model}&\multicolumn{19}{c}{RIDE~\cite{wang2020long}}\cr\cmidrule{2-20}
        &\multicolumn{4}{c}{ImageNet-LT}&&\multicolumn{4}{c}{CIFAR100-LT} &&\multicolumn{4}{c}{Places-LT} &&\multicolumn{4}{c}{iNaturalist 2018}\cr\cmidrule{2-5}\cmidrule{7-10}\cmidrule{12-15}\cmidrule{17-20}
        & Many  & Med. & Few & All &&  Many  & Med. & Few & All &&  Many  & Med. & Few & All &&  Many  & Med. & Few & All\cr
        \midrule
         Expert  $E_1$   & 64.3	 	&  49.0	& 31.9 & 52.6 &&  63.5	 	& 44.8 & 20.3 & 44.0 & &  41.3 & 40.8 & 33.2& 40.1  &&  66.6 & 67.1 & 66.5 & 66.8 \\
         Expert  $E_2$  &  64.7	 	& 49.4	& 31.2	& 52.8 && 63.1	 	& 44.7	& 20.2 	& 43.8 & & 43.0& 40.9& 33.6& 40.3 &&  66.1 & 67.1 & 66.6 & 66.8 \\
         Expert  $E_3$    & 64.3       & 48.9	 	& 31.8 &	52.5 && 63.9      & 45.1	 	& 20.5 & 44.3 & & 42.8 &  41.0     &33.5   &  40.2  && 65.3 &67.3& 66.5& 66.7 \\ \midrule
         Ensemble & 68.0      & 52.9	 	& 35.1 &	56.3 && 67.4       & 49.5	 	&  23.7   & 48.0 & & 43.2& 41.1& 33.5& 40.3 && 71.5 & 72.0 & 71.6 & 71.8 \\
         
         \midrule \midrule
         \multirow{4}{*}{Model}&\multicolumn{19}{c}{SADE (ours)}\cr\cmidrule{2-20}
        &\multicolumn{4}{c}{ImageNet-LT}&&\multicolumn{4}{c}{CIFAR100-LT} &&\multicolumn{4}{c}{Places-LT} &&\multicolumn{4}{c}{iNaturalist 2018}\cr\cmidrule{2-5}\cmidrule{7-10}\cmidrule{12-15}\cmidrule{17-20}
        & Many  & Med. & Few & All &&  Many  & Med. & Few & All &&  Many  & Med. & Few & All &&  Many  & Med. & Few & All\cr
        \midrule 
         Expert  $E_1$   & \textbf{68.8}	 	&  43.7	& 17.2	& 49.8 &&  \textbf{67.6}	 	& 36.3	& 6.8  & 38.4 & & \textbf{47.6}& 27.1 & 10.3& 31.2  && \textbf{76.0} & 67.1 & 59.3 & 66.0 \\
         Expert  $E_2$  & 65.5	 	&  50.5	& 33.3	& \textbf{53.9}&& 61.2	 	& 44.7	& 23.5 & \textbf{44.2}   & & 42.6 &  42.3    & 32.3& \textbf{40.5}  && 69.2 & 70.7 & 69.8 & \textbf{70.2} \\
         Expert  $E_3$    & 43.4         & 48.6   	& \textbf{53.9}  &	 47.3  && 14.0         & 27.6	 	& \textbf{41.2} &	25.8    & & 22.6& 37.2& \textbf{45.6} & 33.6  && 55.6 & 61.5 & \textbf{72.1} & 65.1  \\ \midrule
         Ensemble & 67.0       & 56.7 	 	& 42.6 &	\textbf{58.8} && 61.6        & 50.5	 	&  33.9  & \textbf{49.4}  & & 40.4 & 43.2 &36.8 &\textbf{40.9}  && 74.4 & 72.5 & 73.1 & \textbf{72.9} \\       
        \bottomrule

	\end{tabular}}
	 \end{threeparttable}
	 \end{center} 

\end{table*}

\clearpage
\subsection{More Results on  Test-time Self-supervised Aggregation}
\label{cls_testtime_learning_supp}
This appendix provides more   results to examine the effectiveness of our test-time self-supervised aggregation strategy.  We report results in   Table~\ref{table_test_time_supp}, from which we draw several observations.  

First of all, our method is able to learn suitable expert aggregation weights for test-agnostic    class distributions, without relying on the true test class distribution. For  the forward long-tailed test distribution, where the test data number of many-shot classes is more than that of medium-shot and few-shot classes, our method learns a higher weight for  the forward expert $E_1$ who is skilled in many-shot classes, and learns relatively low weights for the expert $E_2$ and expert $E_3$ who are good at medium-shot and few-shot classes. 
Meanwhile, for the uniform test class distribution where all classes have the same number of test samples, our test-time expert aggregation strategy learns relatively balanced weights for the three experts. For example, on the uniform ImageNet-LT test data, the learned weights by our strategy are 0.33, 0.33 and 0.34 for the three experts, respectively. In addition, for the backward long-tailed test distributions, our method   learns a higher weight for the backward expert $E_3$ and a  relatively low weight for the forward expert $E_1$.
Note that when the class imbalance ratio becomes larger, our method is able to  learn more diverse expert weights adaptively for fitting  the actual   test class distributions.
 
Such results not only demonstrate the effectiveness of our proposed  strategy, but also verify the theoretical analysis that our method can simulate   the unknown test class distribution.
To our best knowledge, such an ability  is quite promising, since it is   difficult  to know the true test class distributions in real-world application. Therefore, our method opens the opportunity for tackling unknown class distribution shifts at test time, and can serve as  a better candidate to handle real-world long-tailed  learning applications.

\begin{table*}[h]   \vspace{-0.1in}
	\caption{The learned aggregation weights by our test-time self-supervised aggregation strategy on different test class distributions of ImageNet-LT, CIFAR100-LT, Places-LT and iNaturalist 2018.  The results show that our self-supervised strategy is able to learn suitable expert weights for various unknown  test class distributions. }
	\label{table_test_time_supp} 
 \begin{center} 
 \begin{threeparttable} 
    \resizebox{0.85\textwidth}{!}{
 	\begin{tabular}{lcccccccccccc}\toprule 
 	 
        \multirow{2}{*}{Test Dist.}
        &&\multicolumn{3}{c}{ImageNet-LT} &&\multicolumn{3}{c}{CIFAR100-LT(IR10)}
        &&\multicolumn{3}{c}{CIFAR100-LT(IR50)}  \cr \cmidrule{3-5}  \cmidrule{7-9} \cmidrule{11-13} 
        && E1 ($w_1$)  & E2 ($w_2$) & E3  ($w_3$)  && E1 ($w_1$)  & E2 ($w_2$) & E3  ($w_3$) && E1 ($w_1$)  & E2 ($w_2$) & E3  ($w_3$)   \cr
        \midrule
        Forward-LT-50   && 0.52	 	&  0.35	& 0.13	 &&  0.53 & 0.38 & 0.09 && 0.55  & 0.38 & 0.07\\
        Forward-LT-25   &&  0.50 & 0.35 & 0.15   && 0.52 & 0.37 & 0.11 &&  0.54  & 0.38 & 0.08\\
         Forward-LT-10  && 0.46	 	&  0.36	& 0.18   && 0.47 & 0.36 & 0.17 &  & 0.52 & 0.37 & 0.11\\
         Forward-LT-5   &&  0.43 & 0.34 & 0.23	  && 0.46 & 0.34 & 0.20 &  & 0.50 & 0.36 & 0.14 \\  
         Forward-LT-2   &&  0.37 & 0.35 & 0.28	 && 0.39  & 0.37 & 0.24   &  & 0.39 & 0.38 & 0.23\\
         Uniform    && 0.33       & 0.33 	 &	 0.34 && 0.38 & 0.32 & 0.3 &  & 0.35 & 0.33 & 0.33 \\
         Backward-LT-2  && 0.29 & 0.31 & 0.40 &&0.35 & 0.33 & 0.31 &  & 0.30 & 0.30 & 0.40\\ 
         Backward-LT-5  &&  0.24 & 0.31 & 0.45 && 0.31 & 0.32 & 0.37 &  &  0.21 & 0.29 & 0.50\\
         
          Backward-LT-10  && 0.21 	& 0.29	& 0.50  && 0.26 & 0.32 & 0.42 &  &   0.20 & 0.29 & 0.51\\
          Backward-LT-25  &&  0.18 & 0.29 & 0.53 &&  0.24 & 0.30 & 0.46 &  & 0.18 & 0.27 & 0.55\\
         Backward-LT-50    && 0.17       &0.27	 &	0.56 &&  0.23 & 0.28 & 0.49 &  & 0.14 & 0.24 & 0.62\\  
    \midrule \midrule
        \multirow{2}{*}{Test Dist.}
        &&\multicolumn{3}{c}{CIFAR100-LT(IR100)} &&\multicolumn{3}{c}{Places-LT}
        &&\multicolumn{3}{c}{iNaturalist 2018}  \cr \cmidrule{3-5}  \cmidrule{7-9} \cmidrule{11-13} 
        && E1 ($w_1$)  & E2 ($w_2$) & E3  ($w_3$)  && E1 ($w_1$)  & E2 ($w_2$) & E3  ($w_3$) && E1 ($w_1$)  & E2 ($w_2$) & E3  ($w_3$)   \cr
        \midrule
        Forward-LT-50   && 0.56 & 0.38 & 0.06 & &   0.50 & 0.20 & 0.20 & & -& - & - \\
        Forward-LT-25   && 0.55   & 0.38 & 0.07 &&  0.50 & 0.20 & 0.20  & & -& - & - \\
         Forward-LT-10  &&  0.52 & 0.39 &0.09 &&   0.50 & 0.20 & 0.20  & & -& - & - \\
         Forward-LT-5   &&  0.51 & 0.37 & 0.12 &&  0.46 & 0.32 & 0.22 & & -& - & - \\  
         Forward-LT-2   &&  0.49 & 0.35 &0.16 && 0.40 & 0.34 & 0.26 & &  0.41 & 0.34 &0.25\\
         Uniform     && 0.40 & 0.35 &0.24 &&  0.25 & 0.34 &0.41 & &  0.33 &0.33 & 0.34\\
         Backward-LT-2  && 0.33 & 0.31 & 0.36&&  0.18 & 0.30 &0.52 & &  0.28& 0.32 &0.40\\ 
         Backward-LT-5  && 0.28 & 0.30 & 0.42 &&  0.17 & 0.28 & 0.55 & & -& - & - \\
         
          Backward-LT-10  && 0.23 & 0.28&  0.49&&  0.17 & 0.27 & 0.56 & & -& - & - \\
          Backward-LT-25  && 0.21 & 0.26 & 0.53 &&  0.17 & 0.27 & 0.56 & & -& - & -  \\
         Backward-LT-50    && 0.16 & 0.28 & 0.56 &&  0.17 & 0.27 & 0.56  & &-& - & -  \\      
        \bottomrule

	\end{tabular}}
	 \end{threeparttable} 
	 \end{center}  
\end{table*} 

\begin{figure*}  
  \begin{minipage}{0.48\linewidth}
   \centerline{\includegraphics[width=7.5cm,clip]{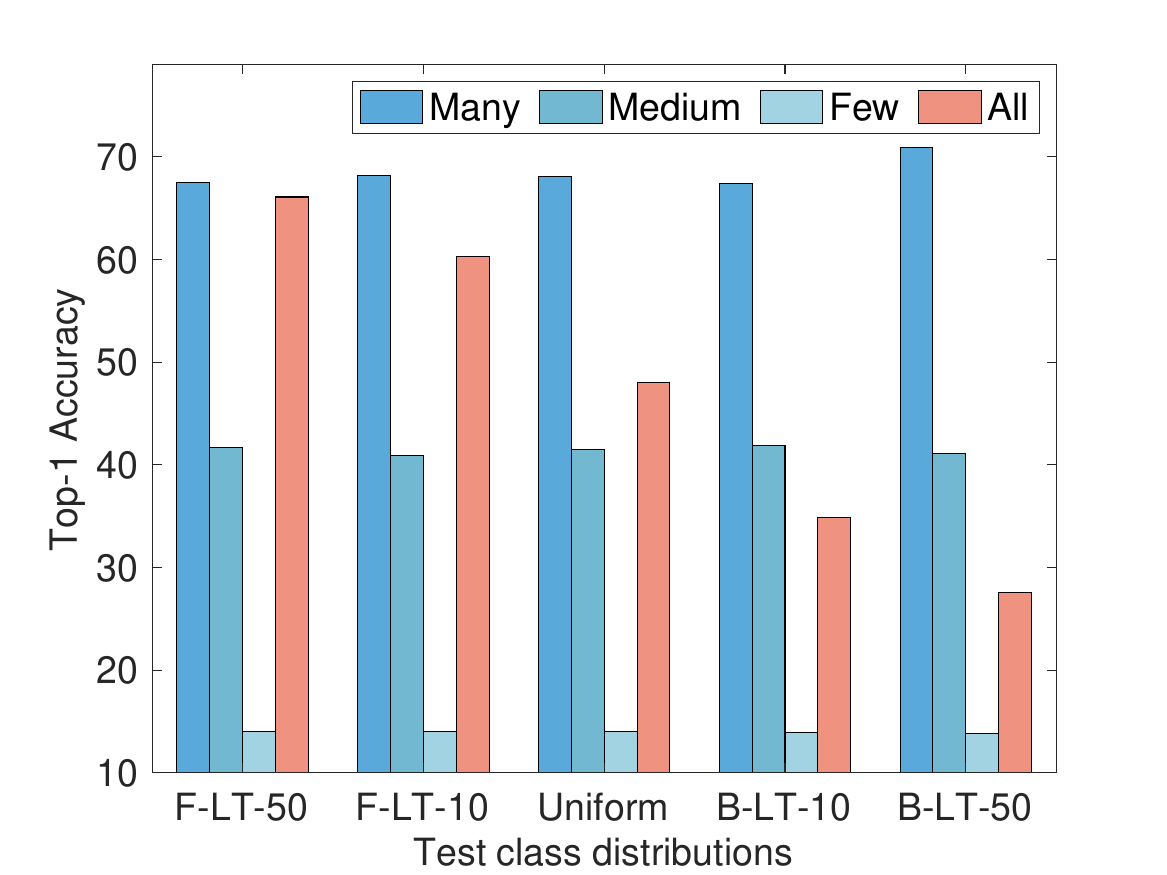}}
   \centerline{\small{(a) Softmax}}
  \end{minipage}
  \hfill
\begin{minipage}{0.48\linewidth}
   \centerline{\includegraphics[width=7.5cm,clip]{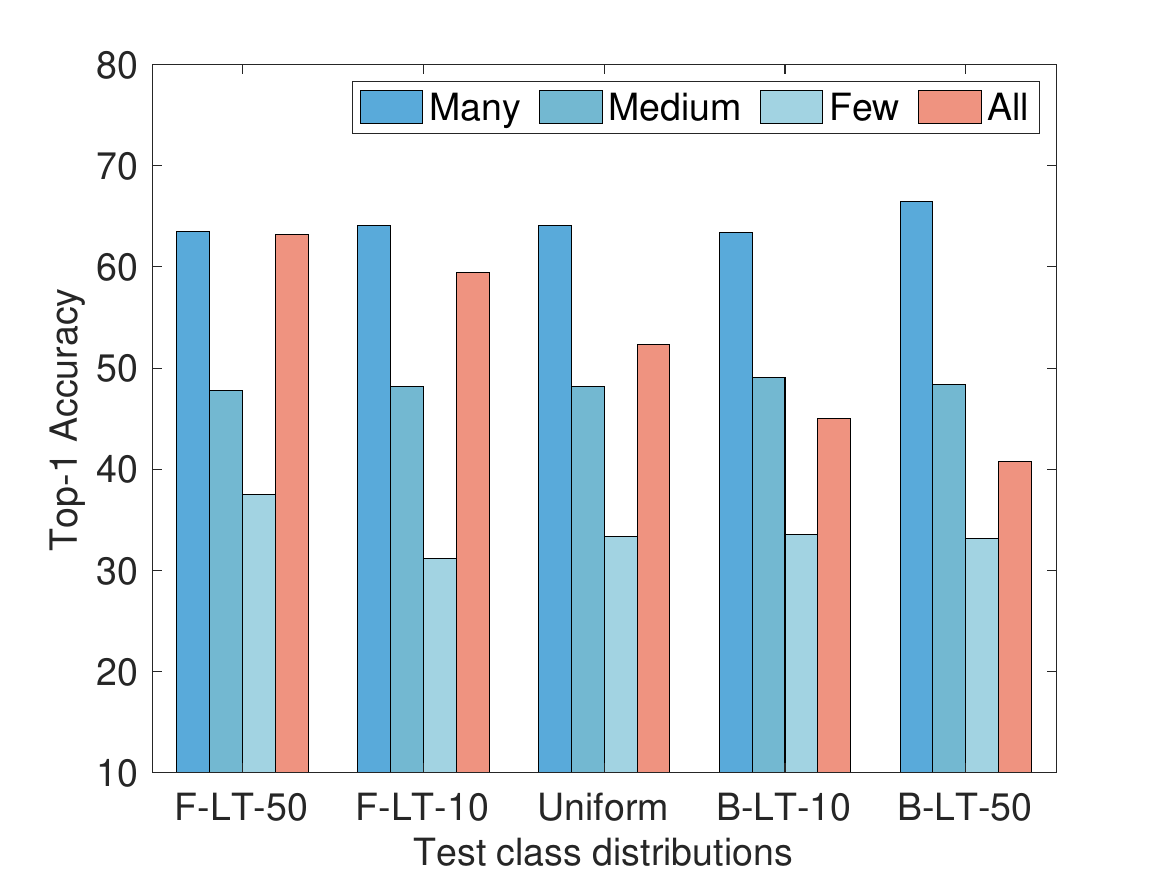}}
   \centerline{\small{(b) Balanced Softmax~\cite{jiawei2020balanced}}}
  \end{minipage}
  \vfill 
  \begin{minipage}{0.48\linewidth}
   \centerline{\includegraphics[width=7.5cm,clip]{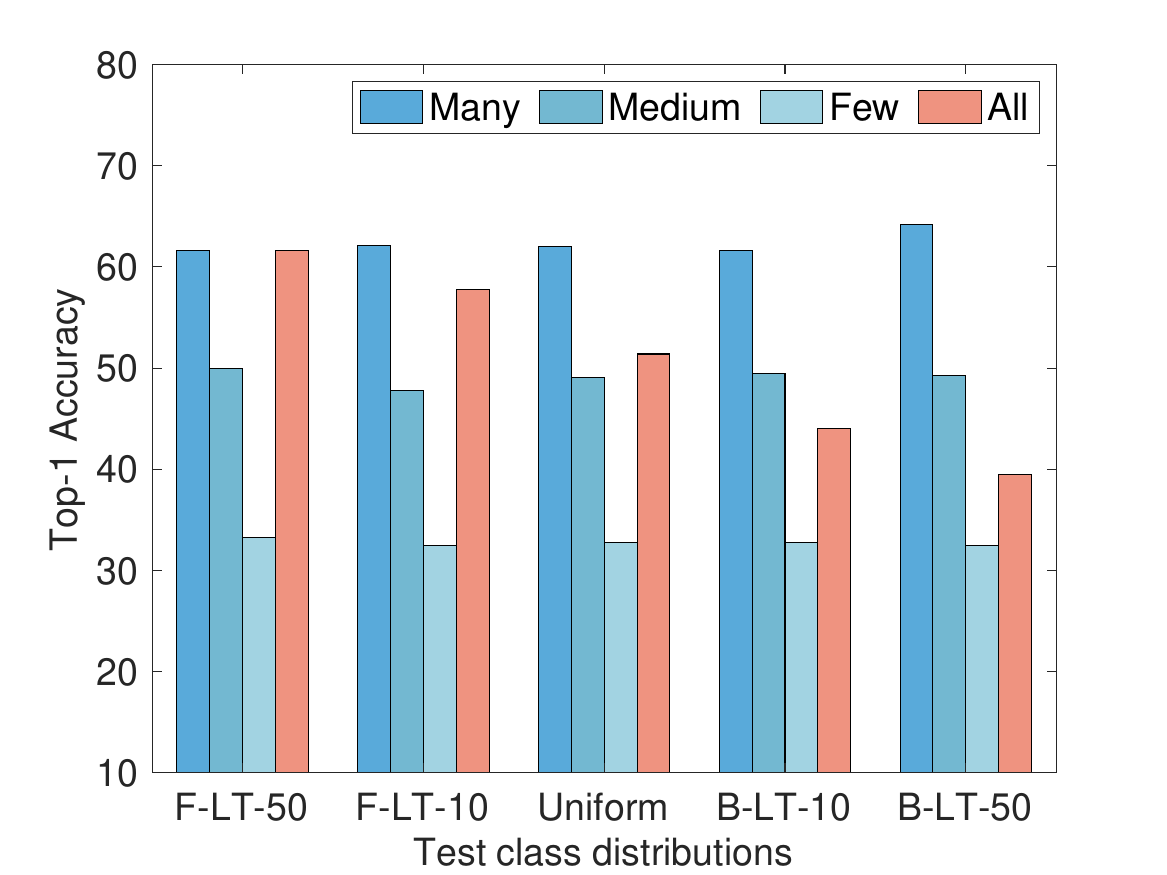}}
   \centerline{\small{(c) MiSLAS~\cite{zhong2021improving}}}
  \end{minipage}
   \hfill    
   \begin{minipage}{0.48\linewidth}
   \centerline{\includegraphics[width=7.5cm,clip]{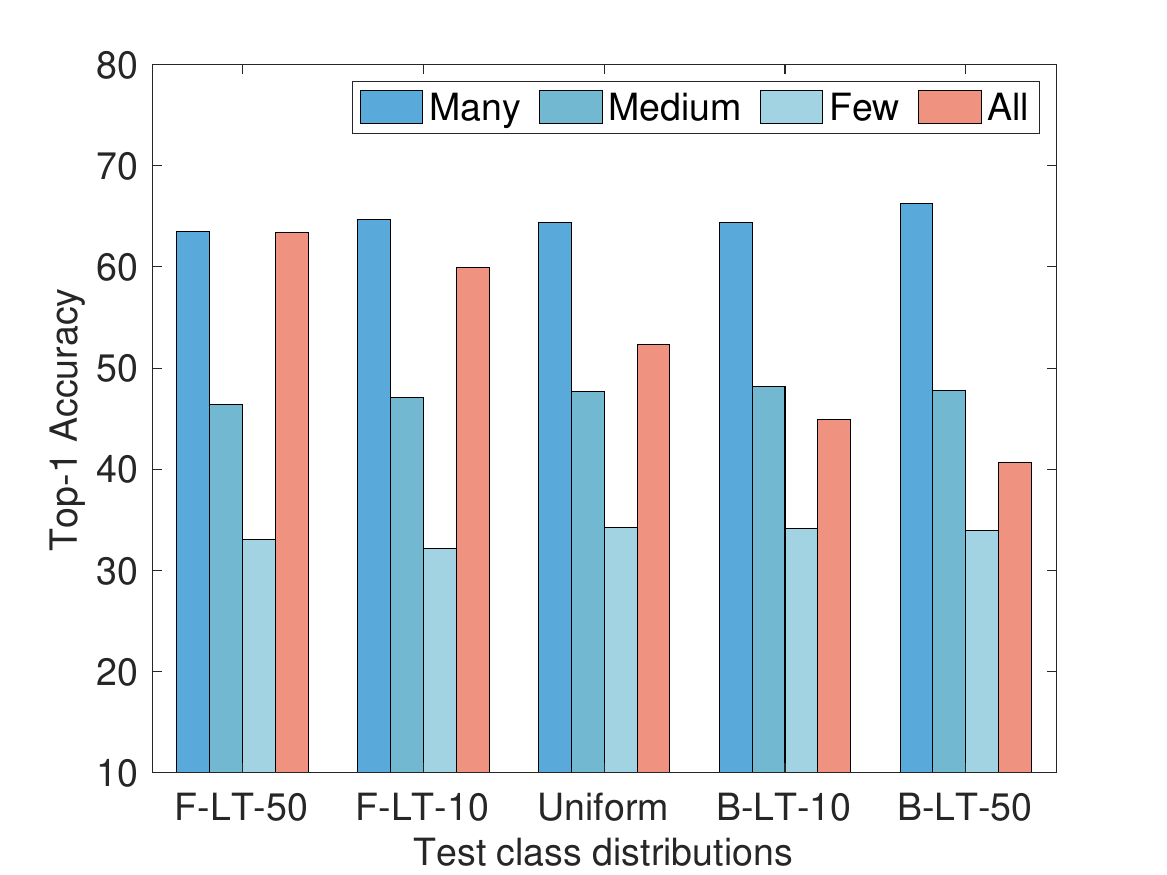}}
   \centerline{\small{(d) LADE w/o prior~\cite{hong2020disentangling}}}
   \end{minipage} 
     \vfill 
  \begin{minipage}{0.48\linewidth}
   \centerline{\includegraphics[width=7.5cm,clip]{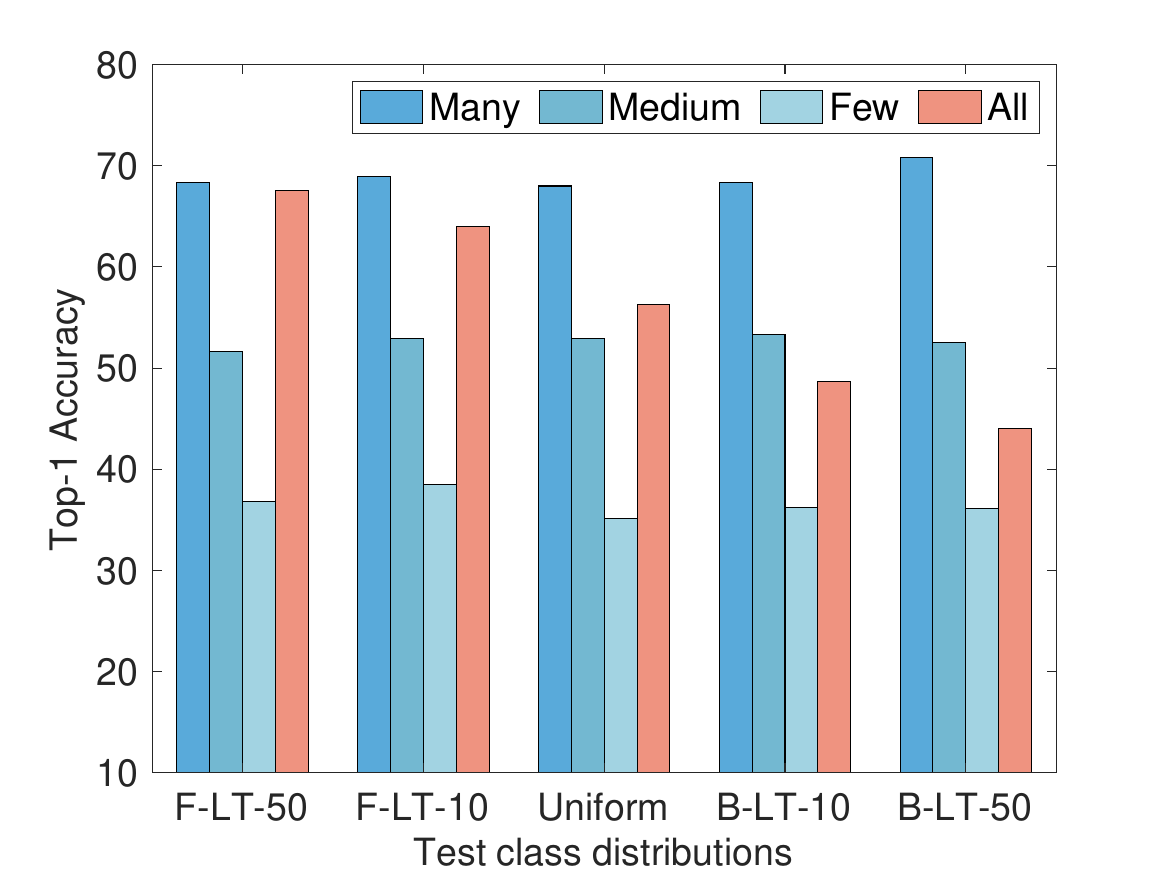}}
   \centerline{\small{(e) RIDE~\cite{wang2020long}}}
  \end{minipage}
   \hfill    
   \begin{minipage}{0.48\linewidth}
   \centerline{\includegraphics[width=7.5cm,clip]{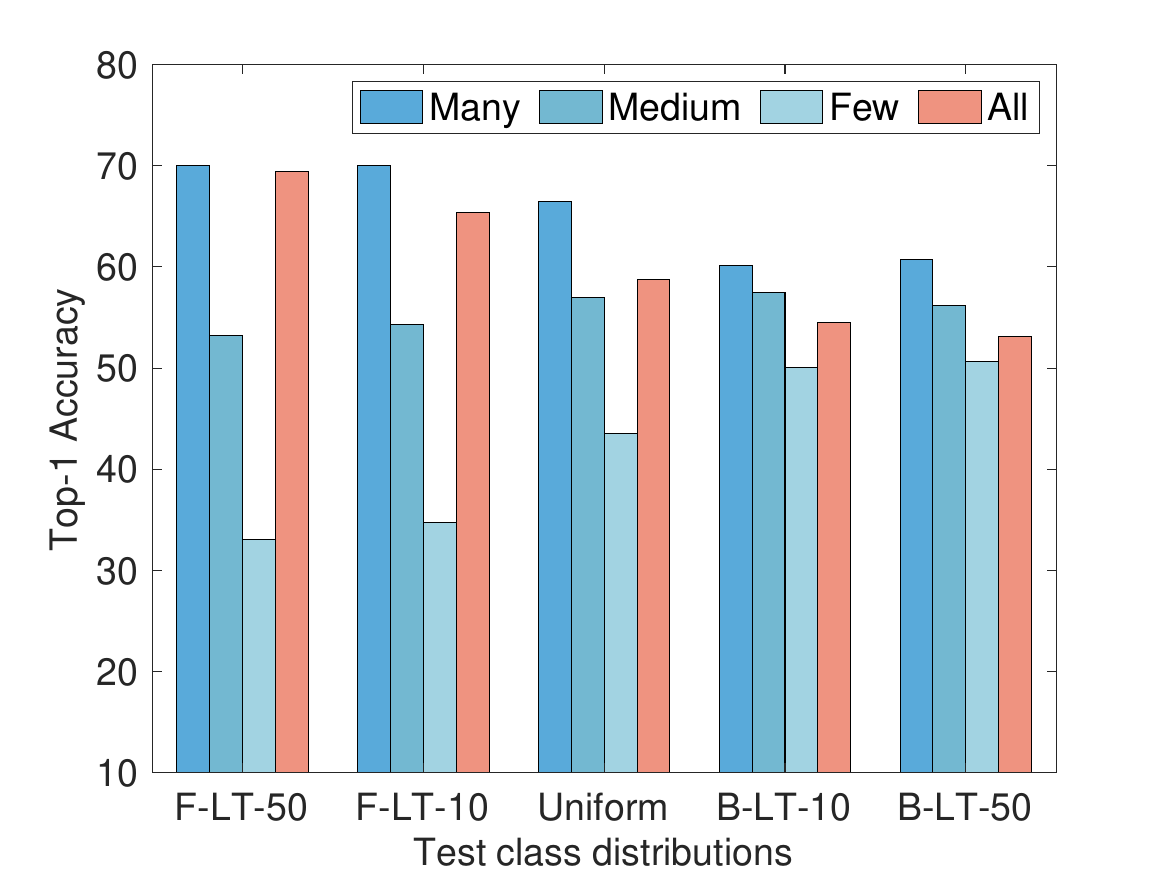}}
   \centerline{\small{(f) SADE (ours)}}
   \end{minipage}  
  \caption{Top-1 accuracy of existing long-tailed (LT) methods on ImageNet-LT  with various test class distributions, including uniform,  forward and backward long-tailed ones with imbalance ratios 10 and 50, respectively. Here, ``F-LT-$N$" and ``B-LT-$N$ indicate the cases where test samples follow the same long-tailed distribution as training data  and inversely long-tailed to the training data, with the imbalance ratio $N$, respectively.   The results show that \textbf{existing methods perform very similarly on various test class distributions in terms of their performance on many-shot, medium-shot and few-shot classes. In contrast, our proposed method is 
 capable of adaptingbi to various test class distributions in terms of many-shot, medium-shot and few-shot performance, thus leading to better overall performance on each test class distribution.}}
  \label{visualization_existing}  
\end{figure*}

 Relying on the learned expert weights, our method   aggregates the three experts appropriately and   achieves better performance on the  dominant test classes, thus obtaining    promising performance gains on various    test   distributions, as shown in Table~\ref{table_expert_learning_performance_supp}. Note that the performance gain  compared to existing methods gets larger as the test dataset gets more imbalanced. For example, on CIFAR100-LT with the imbalance ratio of 50, our test-time self-supervised   strategy  obtains a 7.7$\%$ performance gain   on the Forward-LT-50 distribution and obtains a 9.2$\%$ performance gain on the Backward-LT-50  distribution, both of which are non-trivial.  Such an observation   is also   supported by the visualization result  of  Figure~\ref{visualization_existing}, which plots the results of existing methods on ImageNet-LT with different test class distributions regarding the three class subsets. 

In addition, since the imbalance degrees of  the  test datasets are relatively low on  iNaturalist 2018,  the simulated test class distributions are thus relatively balanced. As a result, the obtained performance improvement is not that significant, compared to  other datasets. However, if there are   more iNaturalist test samples following highly imbalanced test class distributions   in real applications, our method would obtain more promising results.

  \begin{table*}[h]   \vspace{0.75in}
	\caption{The performance improvement  via  test-time self-supervised aggregation  on various test class distributions of ImageNet-LT, CIFAR100-LT, Places-LT and iNaturalist 2018.}
	\label{table_expert_learning_performance_supp} 
 \begin{center}
 \begin{threeparttable} 
    \resizebox{1\textwidth}{!}{
 	\begin{tabular}{lccccccccccccccccccc}\toprule  
        \multirow{4}{*}{Test Dist.}&\multicolumn{9}{c}{ImageNet-LT} &&\multicolumn{9}{c}{CIFAR100-LT(IR10)}\cr\cmidrule{2-10}\cmidrule{12-20}
        &\multicolumn{4}{c}{Ours w/o test-time   aggregation}&&\multicolumn{4}{c}{Ours w/ test-time   aggregation} && \multicolumn{4}{c}{Ours w/o test-time   aggregation}&&\multicolumn{4}{c}{Ours w/ test-time   aggregation}  \cr\cmidrule{2-5} \cmidrule{7-10}\cmidrule{12-15}\cmidrule{17-20}
        & Many  & Med. & Few & All &&  Many  & Med. & Few & All && Many  & Med. & Few & All &&  Many  & Med. & Few & All \cr
        \midrule
         Forward-LT-50   & 65.6	 	&  55.7	& 44.1 &   {65.5}  && 70.0 	& 53.2	& 33.1 &  {69.4 (\red{+3.9})}    && 66.3 & 58.3 & -&  {66.3}  &&  69.0  & 50.8 & - &  {71.2 (\red{+4.9})}  \\
         Forward-LT-25 & 65.3 & 56.9 &43.5 &  {64.4}   && 69.5 & 53.2 & 32.2 &  {67.4 (\red{+3.0})} &&  63.1 & 60.8 & - & {64.5} && 67.6 & 52.2 &- &  {69.4 (\red{+4.9})} \\ 
         Forward-LT-10  &  66.5	 	& 56.8 & 44.2	&  {63.6}  && 69.9	 	& 54.3	& 34.7 	&  {65.4 (\red{+1.8})} &&   64.1 & 58.8 & - &  {64.1} && 67.2 & 54.2 &-&  {67.6 (\red{+3.5})} \\
         Forward-LT-5 &65.9 & 56.5 & 43.3 &  {62.0} &&68.9 & 54.8  & 35.8 &  {63.0  (\red{+1.0})} &&  62.7 & 57.1 & - & {62.7} &&  66.9 & 54/3 & - &  {66.3 (\red{+3.6})}\\ 
         Forward-LT-2 & 66.2 & 56.5 & 42.1 &  {60.0}&& 68.2 & 56.0 & 40.1 &  {60.6 (\red{+0.6})} &&  62.8 & 56.3 &- &  {61.6}  && 66.1 & 56.6 &- &  {64.4 (\red{+2.8})}\\
         Uniform    &  67.0	 	& 56.7	& 42.6	& {58.8} && 66.5	 	& 57.0	& 43.5 	&  {58.8 (+0.0)}  &&  65.5 & 59.9 & - &  {63.6}  && 65.8 & 58.8 & - &  {63.6 (+0.0)} \\
         Backward-LT-2& 66.3 & 56.7 & 43.1 &  {56.8} && 65.3& 57.1 & 45.0 & {57.1 (\red{+0.3})} &&   62.7 & 56.9 &- &  {60.2} & & 65.6 & 59.5 & - &  {62.9 (\red{+2.7})} \\
          Backward-LT-5 & 66.6 & 56.9 & 43.0 &  {54.7} && 63.4 & 56.4 &47.5 &  {55.5 (\red{+0.8})} && 62.8 & 57.5 &- &  {59.7} & & 65.1 & 60.4 & - &  {62.4 (\red{+2.7})}   \\
         Backward-LT-10    & 65.0      & 57.6 	& 43.1 &  {53.1} && 60.9      & 57.5	 	& 50.1 &  {54.5 (\red{+1.4})}   &&    63.5 & 58.2 & - &  {59.8}& &  62.5 & 61.4 & - &  {61.7 (\red{+1.9})}\\ 
           Backward-LT-25 &  64.2 & 56.9 & 43.4 &  {51.1} &&  60.5 & 57.1 & 50.0 &  {53.7 (\red{+2.6})} && 63.4 & 57.7 &- &  {58.7}& & 61.9 & 62.0 & - &  {62.1 (\red{+3.4})}\\
          Backward-LT-50 & 69.1       & 57.0	 	& 42.9 &	 {49.8} && 60.7      & 56.2	 	&  50.7 &  {53.1 (\red{+3.3})}   && 62.0 & 57.8 & - &  {58.6}& &  62.6 & 62.6 & - &  {63.0 (\red{+3.8})}\\
     
        \midrule   \midrule
   \multirow{4}{*}{Test Dist.}&\multicolumn{9}{c}{CIFAR100-LT(IR50)} &&\multicolumn{9}{c}{CIFAR100-LT(IR100)}\cr\cmidrule{2-10}\cmidrule{12-20}
        &\multicolumn{4}{c}{Ours w/o test-time   aggregation}&&\multicolumn{4}{c}{Ours w/ test-time   aggregation} && \multicolumn{4}{c}{Ours w/o test-time   aggregation}&&\multicolumn{4}{c}{Ours w/ test-time   aggregation}  \cr\cmidrule{2-5} \cmidrule{7-10}\cmidrule{12-15}\cmidrule{17-20}
        & Many  & Med. & Few & All &&  Many  & Med. & Few & All && Many  & Med. & Few & All &&  Many  & Med. & Few & All \cr
        \midrule
         Forward-LT-50   & 59.7  & 53.3  & 26.9  &  {59.5}  & &  68.0   & 44.1 & 19.4 &  {67.2 (\red{+7.7})}   &  & 60.7 & 50.3 & 32.4 &  {58.4} &  & 69.9  & 48.8 & 14.2 &  {65.9 (\red{+7.5})}   \\
         Forward-LT-25 &  59.1  & 51.8  & 32.6  &  {58.6}  & & 67.3 & 46.2 & 19.5 &  {64.5 (\red{+6.9})} &  & 60.6 & 49.6 & 29.4 &  {57.0} &  & 68.9 & 46.5 & 15.1 &  {62.5 (\red{+5.5})}   \\ 
         Forward-LT-10  &  59.7  & 47.2  & 36.1  & {56.4}  & & 67.2 & 45.7 & 24.7 &  {61.2 (\red{+4.8})} &  & 60.1 & 48.6 & 28.4 &  {54.4} &  & 68.3 & 46.9 & 16.7 &  {58.3 (\red{+3.9})}  \\
         Forward-LT-5 &   59.7  & 46.9  & 36.9  &  {54.8}  & &  67.0 & 45.7 & 29.9 &  {58.6 (\red{+3.4})} &  & 60.3 & 50.3 & 29.5 &  {53.1} &  & 68.3 & 45.3 & 19.4 &  {54.8 (\red{+1.7})}  \\ 
         Forward-LT-2 &   59.2  & 48.4  & 41.9  &  {53.2}  & &  63.8 & 48.5 & 39.3 &  {55.4 (\red{+2.2})} &  & 60.6 & 48.8 & 31.3 &  {50.1} &  & 68.2 & 47.6 & 22.5 &  {51.1 (\red{+1.0})}   \\
         Uniform    &    61.0  &50.2  & 45.7  &  {53.8}  & &  61.5 & 50.2 & 45.0 &  {53.9 (\red{+0.1})} &  & 61.6 & 50.5 & 33.9 &  {49.4} &  & 65.4  & 49.3 & 29.3 &  {49.8 (\red{+0.4})} \\
         Backward-LT-2&  59.0  & 48.2  & 42.8  &  {50.1}  & & 57.5 & 49.7 & 49.4 &  {51.9 (\red{+1.8})} &  &  61.2 & 49.1 & 30.8 &  {45.2} &  &  63.1 & 49.4 & 31.7 &  {46.2 (\red{+1.0})}   \\
          Backward-LT-5 &    60.1  & 48.6  & 41.8  & {48.2}  & & 50.0 & 49.3 & 54.2 &  {50.9 (\red{+2.7})} &  & 62.0 & 48.9 & 32.0 &  {42.6} &  & 56.2 & 49.1 & 38.2 &  {44.7 (\red{+2.1})}   \\
         Backward-LT-10    &   58.6  & 46.9  & 42.6  &  {46.1}  & & 49.3 & 49.1 & 54.6 &  {51.0 (\red{+4.9})} &  & 60.6 & 48.2 & 31.7 &  {39.7} &  & 52.1 & 47.9 & 40.6 &  {43.9 (\red{+4.2})}   \\ 
           Backward-LT-25 &  55.1  & 48.9  & 41.2  &  {44.4}  & &  44.5 & 46.6 & 57.0 &  {51.7 (\red{+7.3})} &  & 58.2 & 47.9 & 32.2 &  {36.7} &  &  48.7 & 44.2 & 41.8 &  {42.5 (\red{+5.8})}    \\
          Backward-LT-50 &   57.0  & 48.8  & 41.6  &  {43.6}  & &  45.8 & 46.6 & 58.4 &   {52.8 (\red{+9.2})} &  & 66.9 & 48.6 & 30.4 &  {35.0}  &  & 49.0 & 42.7 & 42.5 &  {42.4 (\red{+7.4})}  \\
          
                  \midrule   \midrule
   \multirow{4}{*}{Test Dist.}&\multicolumn{9}{c}{Places-LT} &&\multicolumn{9}{c}{iNaturalist 2018}\cr\cmidrule{2-10}\cmidrule{12-20}
        &\multicolumn{4}{c}{Ours w/o test-time   aggregation}&&\multicolumn{4}{c}{Ours w/ test-time   aggregation} && \multicolumn{4}{c}{Ours w/o test-time   aggregation}&&\multicolumn{4}{c}{Ours w/ test-time   aggregation}  \cr\cmidrule{2-5} \cmidrule{7-10}\cmidrule{12-15}\cmidrule{17-20}
        & Many  & Med. & Few & All &&  Many  & Med. & Few & All && Many  & Med. & Few & All &&  Many  & Med. & Few & All \cr
        \midrule
         Forward-LT-50   &  43.5  & 42.5 & 65.9 &  {43.7}  &  & 46.8 & 39.3 & 30.5 &  {46.4 (\red{+2.7})} &  &  -& - & -& -&&  -& - & -& -\\
         Forward-LT-25 & 42.8 & 42.1 & 29.3 &  {42.7} &  & 46.3 & 38.9 & 23.6 &  {44.9 (\red{+2.3})} &  &   -& - & -& -&&   -& - & -& -\\ 
         Forward-LT-10  & 42.3 & 41.9 & 34.9 &  {42.3} &  & 45.4 & 39.0 & 27.0 &  {43.3 (\red{+1.0})} &  &   -& - & -& - &&   -& - & -& -  \\
         Forward-LT-5 &  43.0 & 44.0 & 33.1 &  {42.4} &  &  45.6 & 40.6  & 27.3 &  {42.6 (\red{+0.2})} &  &   -& - & -& - &&  -& - & -& -\\ 
         Forward-LT-2 &  43.4 & 42.4 & 32.6 &  {41.3} &  &  44.9 & 41.2 & 29.5 &  {41.3 (+0.0)} &  & 73.9 &72.4&72.0& {72.4}  && 75.5&72.5&70.7& {72.5 (\red{+0.1})}\\
         Uniform    &   43.1 & 42.4 & 33.2 &  {40.9} &  &  40.4 & 43.2 & 36.8 &  {40.9 (+0.0)} &  & 74.4&72.5&73.1 & {72.9} && 74.5& 72.5 &73.0 &  {72.9 (+0.0)} \\
         Backward-LT-2&   42.8 & 41.9 & 33.2 &  {39.9} &  &  37.1 & 42.9 & 40.0 &  {40.6 (\red{+0.7})} &  &  76.1&72.8&72.6& {73.1} && 74.9 & 72.6&73.7 &  {73.5 (\red{+0.4})}\\
          Backward-LT-5 &  43.1 & 42.0 & 33.6 &  {39.1} &  &  36.4 & 42.7 & 41.1 &  {41.1 (\red{+2.0})} &  &  -& - & -& - &&   -& - & -& -\\
         Backward-LT-10    &  43.5 & 42.9 & 33.7 &  {38.9} &  &  35.2 & 43.2 & 41.3 &  {41.4 (\red{+2.5})} &  &   -& - & -& - &&   -& - & -& -\\ 
           Backward-LT-25 & 44.6 & 42.4 & 33.6 &  {37.8} &  &  38.0 & 43.5 & 41.1 &  {42.0 (\red{+4.2})} &  &  -& - & -& - &&  -& - & -& - \\
          Backward-LT-50 &  42.2 & 43.4 & 33.3 &  {37.2} &  & 37.3 & 43.5 & 40.5 &  {41.6 (\red{+4.7})} &  & -& - & -& - && -& - & -& -\\
     \bottomrule
	\end{tabular}}
	 \end{threeparttable}
	 \end{center} 
    
\end{table*}

\clearpage 
\section{Ablation Studies on Skill-diverse Expert Learning}\label{App_E}
\subsection{Discussion on Expert Number}\label{App_E1}
In SADE, we consider three experts, where the ``forward" and ``backward" experts are necessary since they span a wide spectrum of possible test class distributions, while the ``uniform" expert ensures that we retain high accuracy on the uniform test class distributions. 
Nevertheless, our approach can be straightforwardly extended to more than three experts. 
For the models with  more  experts, we   adjust the  hyper-parameter $\lambda$ in Eq. (3) for the new experts and keep the hyper-parameters of the original three experts unchanged, so that different experts are skilled in different types of class distributions.
Following this, we further evaluate  the influence of the expert number on our method based on ImageNet. To be specific, when there are four experts, we set $\lambda=1$ for the new expert; while when there are five experts, we set $\lambda=0.5$ and $\lambda=1$ for the two newly-added experts, respectively.  As shown in Table~\ref{table_expert_number_supp}, with the increasing number of experts, the ensemble performance of our method  is improved on vanilla long-tailed recognition, \eg  four  experts   obtain  a 1.2$\%$ performance gain  compared to  three experts on ImageNet-LT. 
As a result, our method  with more experts obtains consistent performance improvement in test-agnostic long-tailed recognition on various test class distributions   compared to  three experts, as shown in Table~\ref{table_agnostic_expert_number_supp}.
Even so,  only three experts are sufficient to handle   varied test class distributions, and provide a good trade-off between  performance and efficiency.

\begin{table}[h] 
	\caption{Performance of our method with different numbers of experts   on ImageNet-LT with the uniform test distribution.}
	\label{table_expert_number_supp} 
 \begin{center}
 \begin{threeparttable} 
    \resizebox{0.95\textwidth}{!}{
 	\begin{tabular}{lccccccccc}\toprule

           \multirow{2}{*}{Model}
        &\multicolumn{4}{c}{4 experts} &&\multicolumn{4}{c}{5 experts}\cr\cmidrule{2-5}\cmidrule{7-10}
        & Many-shot  & Medium-shot & Few-shot & All classes  &&  Many-shot  & Medium-shot & Few-shot & All classes   \cr
        \midrule 
         Expert  $E_1$   &  69.4 	&  44.5 & 16.5 &50.3  && 69.8 & 44.9 & 17.0 & 50.7 \\
         Expert  $E_2$  & 66.2 & 51.5 & 32.9& 54.6 & & 68.8 & 48.4 & 23.9 & 52.9  \\
         Expert  $E_3$     & 55.7 & 52.7 & 46.8 & 53.4  & & 66.1 & 51.4 & 22.0 &54.5 \\   
         Expert  $E_4$   &  44.1 & 49.7 & 55.9 & 48.4  &  &  56.8 & 52.7 & 47.7 & 53.6 \\ 
         Expert  $E_5$ & - &- & - &- &  & 43.1 & 59.0 & 54.8 & 47.5 \\  \midrule
         Ensemble &  66.6 & 58.4 & 46.7 & 60.0 & &  68.8 & 58.5 & 43.2 & 60.4 \\   
         
        \bottomrule

	\end{tabular}}
	 \end{threeparttable}
	 \end{center} 
\end{table}

\begin{table}[h]  
	\caption{Performance of our method with different numbers of experts   on  various      test class distributions  of ImageNet-LT.} 
	\label{table_agnostic_expert_number_supp} 
 \begin{center}
 \begin{threeparttable} 
    \resizebox{0.95\textwidth}{!}{
	\begin{tabular}{lcccccccccccccc}\toprule  
   	     \multirow{4}{*}{Method} &\multirow{4}{*}{Experts}&  \multicolumn{13}{c}{ImageNet-LT} \cr \cmidrule{3-15}  
       && \multicolumn{5}{c}{Forward} && Uniform  && \multicolumn{5}{c}{Backward} \cr  \cmidrule{3-7} \cmidrule{9-9}\cmidrule{11-15} 
         && 50 &25 &10&5&2 && 1 && 2& 5& 10 &25 &50\cr  
        \midrule
         \multirow{3}{*}{SADE}  &   3 experts & {69.4} &  {67.4}    & {65.4} 	& {63.0}  & {60.6} && {58.8} && {57.1} & {55.5} & {54.5} & {53.7} & {53.1} \\      
         & 4 experts & 70.1 & 68.1 & 66.3 & 64.2 & 61.6 && 60.0 && 58.7 & 57.6 & 56.7 & 56.1 & 55.6\\         
         & 5 experts & 70.7 & 68.9 &  66.8 &  64.5 & 62.1 && 60.4 && 58.7 & 57.2 & 56.3 & 55.6 & 54.7  \\    
      
        \bottomrule
    
	\end{tabular}}
	 \end{threeparttable}
	 \end{center}  
\end{table}

 \newpage
\subsection{Hyper-parameters in Inverse Softmax Loss}
This appendix evaluates the influence of the hyper-parameter $\lambda$ in the   inverse softmax loss for the backward expert, where we fix all other hyper-parameters and only adjust the value of  $\lambda$. 
As shown in Table~\ref{table_lambda_supp}, with the increase of $\lambda$, the backward expert  simulates more inversely long-tailed distribution (to the training data),  and thus the ensemble performance  on \emph{few-shot classes} is better. Moreover, when $\lambda\in\{2,3\}$, our method achieves a better trade-off between  head classes  and tail classes, leading to  relatively better overall performance on ImageNet-LT.

\begin{table}[h]  
	\caption{Influence of the hyper-parameter $\lambda$ in the   inverse softmax loss on ImageNet-LT with the uniform test distribution.}
	\label{table_lambda_supp} 
 \begin{center} 
 \begin{threeparttable} 
    \resizebox{0.8\textwidth}{!}{
 	\begin{tabular}{lcccc}\toprule 

         \multirow{2}{*}{Model}
        &\multicolumn{4}{c}{$\lambda = 0.5$}\cr\cmidrule{2-5}
        & Many-shot classes  & Medium-shot classes & Few-shot classes & All long-tailed  classes \cr
        \midrule 
         Forward Expert  $E_1$   & 69.1 & 43.6 & 17.2 & 49.8  \\
         Uniform Expert  $E_2$  & 66.4 & 50.9 & 33.4 & 54.5   \\
         Backward Expert  $E_3$    & 61.9 & 51.9 & 40.3 & 54.2   \\ \midrule
         Ensemble &  71.0 & 54.6 & 33.4 & 58.0 \\ 
         
         \midrule \midrule 
         
    \multirow{2}{*}{Model}
        &\multicolumn{4}{c}{$\lambda = 1$}\cr\cmidrule{2-5}
        & Many-shot classes  & Medium-shot classes & Few-shot classes & All long-tailed  classes \cr
        \midrule 
         Forward Expert $E_1$    &  69.7 & 44.0 & 16.8 & 50.2   \\
         Uniform Expert $E_2$   &  65.5 & 51.1 & 32.4 & 54.4  \\
         Backward Expert $E_3$     & 56.5 & 52.3 & 47.1 & 53.2 \\ \midrule
         Ensemble &  77.2 & 55.7 & 36.2 & 58.6 \\
         
                  \midrule \midrule 
         
       \multirow{2}{*}{Model}
        &\multicolumn{4}{c}{$\lambda = 2$}\cr\cmidrule{2-5}
       & Many-shot classes  & Medium-shot classes & Few-shot classes & All long-tailed  classes \cr
        \midrule 
         Forward Expert $E_1$    &  68.8	 	&  43.7	& 17.2	& 49.8 \\
         Uniform Expert $E_2$   & 65.5	 	&  50.5	& 33.3	& 53.9\\
         Backward Expert $E_3$     & 43.4         & 48.6   	& 53.9  &	 47.3 \\ \midrule
         Ensemble & 67.0       & 56.7 	 	& 42.6 &	58.8\\

                  \midrule \midrule 
         
          \multirow{2}{*}{Model}
        &\multicolumn{4}{c}{$\lambda =3$}\cr\cmidrule{2-5}
        & Many-shot classes  & Medium-shot classes & Few-shot classes & All long-tailed  classes \cr
        \midrule 
        Forward  Expert $E_1$   & 69.6 & 43.8 & 17.4 & 50.2  \\
        Uniform Expert $E_2$   &66.2 & 50.7 & 33.1 & 54.2 \\
        Backward Expert $E_3$     &   43.4 & 48.6 & 53.9& 48.0 \\  \midrule
         Ensemble &  67.8 & 56.8 & 42.4 & 59.1 \\
         
            \midrule \midrule 
         
          \multirow{2}{*}{Model}
        &\multicolumn{4}{c}{$\lambda =4$}\cr\cmidrule{2-5}
        & Many-shot classes  & Medium-shot classes & Few-shot classes & All long-tailed  classes \cr
        \midrule 
        Forward Expert $E_1$    &   69.1 & 44.1 & 16.3 & 49.9  \\
        Uniform Expert $E_2$   &  65.7 & 50.8 & 32.6 & 54.1 \\
        Backward Expert $E_3$     &  21.9 & 38.1 & 58.9 & 34.7 \\ \midrule
         Ensemble & 60.2 & 57.5 & 50.4 & 57.6  \\
         
            \midrule \midrule 
         
          \multirow{2}{*}{Model}
        &\multicolumn{4}{c}{$\lambda =5$}\cr\cmidrule{2-5}
      & Many-shot classes  & Medium-shot classes & Few-shot classes & All long-tailed  classes \cr
        \midrule 
        Forward Expert $E_1$    & 69.7 & 43.7 & 16.5 & 50.0  \\
        Uniform Expert $E_2$   & 65.9 & 50.9 & 33.0 & 54.2 \\
        Backward Expert $E_3$    &  16.0 & 33.9 & 60.6  & 30.6  \\ \midrule
         Ensemble & 56.3 & 57.5 & 54.0 & 56.6 \\
        \bottomrule

	\end{tabular}}
	 \end{threeparttable}
	 \end{center} 
\end{table}

\clearpage
\section{Ablation Studies on Test-time Self-supervised Aggregation}\label{App_F}
\subsection{Influences of Training Epoch}
As illustrated in Section~\ref{exp_setup}, we set the training epoch of our test-time self-supervised aggregation strategy to 5   on all datasets. Here, we further evaluate the influence of the  epoch number, where we adjust the epoch number from 1 to 100. As shown in Table~\ref{table_test_time_epoch_supp}, when the training epoch number is larger than 5, the learned expert weights by our method  are converged on ImageNet-LT, which verifies that our method is robust enough. The corresponding performance on various test class distributions is reported in Table~\ref{table_epoch_performance_supp}.

\begin{table*}[h] 
	\caption{The influence of the epoch number on the learned expert weights by test-time self-supervised aggregation on ImageNet-LT.}
	\label{table_test_time_epoch_supp}  
 \begin{center}
 \begin{threeparttable} 
    \resizebox{0.85\textwidth}{!}{
 	\begin{tabular}{lcccccccccccc}\toprule 
 	 
        \multirow{2}{*}{Test Dist.}
        &&\multicolumn{3}{c}{Epoch 1} &&\multicolumn{3}{c}{Epoch 5}
        &&\multicolumn{3}{c}{Epoch 10}  \cr \cmidrule{3-5}  \cmidrule{7-9} \cmidrule{11-13} 
        && E1 ($w_1$)  & E2 ($w_2$) & E3  ($w_3$)  && E1 ($w_1$)  & E2 ($w_2$) & E3  ($w_3$) && E1 ($w_1$)  & E2 ($w_2$) & E3  ($w_3$)   \cr
        \midrule
        Forward-LT-50   && 0.44 & 0.33 & 0.23 &&  0.52	 	&  0.35	& 0.13	 &&  0.52 & 0.37 & 0.11 \\
        Forward-LT-25   && 0.43 & 0.34 & 0.23 &&  0.50 & 0.35 & 0.15   &&  0.50 & 0.37 &0.13\\
         Forward-LT-10  && 0.43 & 0.34 & 0.23 && 0.46	 	&  0.36	& 0.18   && 0.46 & 0.36 & 0.18 \\
         Forward-LT-5   && 0.41 & 0.34 & 0.25 &&  0.43 & 0.34 & 0.23	  && 0.43 & 0.35 & 0.22  \\  
         Forward-LT-2   && 0.37 & 0.33 & 0.30&&  0.37 & 0.35 & 0.28	 && 0.38 & 0.33 & 0.29 \\
         Uniform    && 0.34 & 0.31 &0.35 && 0.33       & 0.33 	 &	 0.34 &&  0.33 & 0.32 & 0.35 \\
         Backward-LT-2  && 0.30 & 0.32 & 0.38 && 0.29 & 0.31 & 0.40 &&  0.29 & 0.32 & 0.39\\ 
         Backward-LT-5  && 0.27 & 0.29 & 0.44 && 0.24 & 0.31 & 0.45 && 0.23 & 0.31 & 0.46 \\ 
          Backward-LT-10  && 0.24 & 0.29 & 0.47 && 0.21 	& 0.29	& 0.50  && 0.21 & 0.30 & 0.49 \\
          Backward-LT-25  &&   0.23 & 0.29 & 0.48 && 0.18 & 0.29 & 0.53 && 0.17 & 0.3 & 0.53 \\
         Backward-LT-50    &&  0.24 & 0.29 & 0.47 && 0.17       &0.27	 &	0.56 && 0.15 & 0.28 & 0.57 \\  
    \midrule \midrule
        \multirow{2}{*}{Test Dist.}
        &&\multicolumn{3}{c}{Epoch 20} &&\multicolumn{3}{c}{Epoch 50}
        &&\multicolumn{3}{c}{Epoch 100}  \cr \cmidrule{3-5}  \cmidrule{7-9} \cmidrule{11-13} 
        && E1 ($w_1$)  & E2 ($w_2$) & E3  ($w_3$)  && E1 ($w_1$)  & E2 ($w_2$) & E3  ($w_3$) && E1 ($w_1$)  & E2 ($w_2$) & E3  ($w_3$)   \cr
        \midrule
        Forward-LT-50   &&  0.53 & 0.38 & 0.09 && 0.53 & 0.38 & 0.09 &&  0.53 &0.38 & 0.09\\ 
        Forward-LT-25   && 0.51 & 0.37 &0.12 &&  0.52 & 0.37& 0.11 &&  0.50 & 0.38 & 0.12 \\
         Forward-LT-10  && 0.44 & 0.36 & 0.20 &&  0.45 & 0.37 & 0.18 && 0.46 & 0.36 & 0.18 \\
         Forward-LT-5   &&  0.42 & 0.35 & 0.23 && 0.42 & 0.35 & 0.23 && 0.42 & 0.35 & 0.23 \\
        Forward-LT-2   &&  0.38 & 0.33 &0.29 &&   0.39 & 0.33 & 0.28 && 0.38 & 0.32 & 0.30 \\
         Uniform     && 0.33 & 0.33 & 0.34 &&   0.34 & 0.32 & 0.34&&  0.32 & 0.33 &0.35\\
         Backward-LT-2  && 0.29 & 0.31 & 0.40 &&   0.30 & 0.32 &0.38 &&  0.29 & 0.30& 0.41 \\
         Backward-LT-5  &&  0.24 & 0.31 & 0.45&&   0.23 & 0.29 & 0.48 && 0.25 & 0.30 & 0.45 \\
         
          Backward-LT-10  && 0.20 & 0.30 & 0.50 && 0.21 & 0.31 & 0.48 && 0.21 & 0.30 & 0.49\\
          Backward-LT-25  && 0.16 & 0.30 & 0.54 && 0.17 & 0.29 & 0.54 &&   0.17 & 0.30 & 0.53\\
         Backward-LT-50    && 0.15 & 0.29 & 0.56 &&  0.14 & 0.29 &0.57 &&  0.14 & 0.29 & 0.57\\     
        \bottomrule

	\end{tabular}} 
	 \end{threeparttable} 
	 \end{center}  
\end{table*}

  \begin{table*}[h]  
	\caption{The influence of the epoch number on the performance of test-time self-supervised aggregation on ImageNet-LT.}
	\label{table_epoch_performance_supp} 
 \begin{center}	
 \begin{threeparttable} 
    \resizebox{0.85\textwidth}{!}{
 	\begin{tabular}{lcccccccccccccc}\toprule  
        \multirow{2}{*}{Test Dist.}  
        &\multicolumn{4}{c}{Epoch 1}&&\multicolumn{4}{c}{Epoch 5} && \multicolumn{4}{c}{Epoch 10} \cr\cmidrule{2-5} \cmidrule{7-10}\cmidrule{12-15}
        & Many  & Med. & Few & All &&  Many  & Med. & Few & All && Many  & Med. & Few & All \cr
        \midrule
      Forward-LT-50   &  68.8& 54.6& 37.5 &  {68.5} && 70.0 	& 53.2	& 33.1 &  {69.4}    &&  70.1 & 52.9 &  32.4 &   {69.5}\\
         Forward-LT-25 & 68.6 & 54.9 & 34.9 &  {66.9} &&  69.5 & 53.2 & 32.2 &  {67.4} && 69.7 &  52.5 & 32.5  &  {67.5} \\ 
         Forward-LT-10  & 60.3 & 55.3 & 37.6 &  {65.2} &&  69.9	 	& 54.3	& 34.7 	&  {65.4} && 69.9 & 54.5  & 35.0  &  {65.4} \\
         Forward-LT-5 & 68.4 & 55.3 & 37.3 &  {63.0} &&  68.9 & 54.8  & 35.8 &  {63.0} &&  68.8 & 54.9 & 36.0 &  {63.0} \\ 
     Forward-LT-2 &  67.9 & 56.2 & 40.8 &  {60.6} &&  68.2 & 56.0 & 40.1 &  {60.6} && 68.2  & 56.0 & 39.7 &  {60.5}  \\
         Uniform    & 66.7 & 56.9 & 43.1 &  {58.8} &&  66.5	 	& 57.0	& 43.5 	&  {58.8}  &&  66.4 & 56.9 & 43.4 &  {58.8} \\
        Backward-LT-2& 65.6 &57.1 & 44.7 &  {57.1} &&   65.3& 57.1 & 45.0 & {57.1} &&  65.3 & 57.1 & 45.0 &  {57.1} \\
          Backward-LT-5 & 63.9& 57.6 & 46.8 &  {55.5} &&  63.4 & 56.4 &47.5 &  {55.5} &&   63.3 & 57.4 & 47.8 &  {55.6} \\
         Backward-LT-10    &  62.1 & 57.6 & 47.9 &  {54.2} &&60.9      & 57.5	 	& 50.1 &  {54.5}   && 61.1  & 57.6 & 48.9 &  {54.5} \\ 
           Backward-LT-25 & 62.4 & 57.6 & 48.5 &  {53.4} && 60.5 & 57.1 & 50.0 &  {53.7} && 60.5 & 57.1 & 50.3 &  {53.8} \\
          Backward-LT-50 & 64.9 & 56.7 & 47.8 &  {51.9} && 60.7      & 56.2	 	&  50.7 &  {53.1}   && 60.1 & 55.9 & 51.2 &  {53.2} \\

        \midrule   \midrule
         \multirow{2}{*}{Test Dist.}     &\multicolumn{4}{c}{Epoch 20}&&\multicolumn{4}{c}{Epoch 50} && \multicolumn{4}{c}{Epoch 100} \cr\cmidrule{2-5} \cmidrule{7-10}\cmidrule{12-15}
        & Many  & Med. & Few & All &&  Many  & Med. & Few & All && Many  & Med. & Few & All \cr
        \midrule
         Forward-LT-50   &  70.3 & 52.2 & 32.4 &  {69.5} && 70.3 & 52.2 & 32.4 &  {69.5} &  & 70.0  & 52.2 & 32.4 &  {69.3}           \\
         Forward-LT-25 & 69.8 & 52.4 & 31.4 &  {67.5} & &    69.9 & 52.3 & 31.4 &  {67.6} &  & 69.7 & 52.6 & 32.6 &  {67.5}          \\
         Forward-LT-10  &  69.6 & 54.8 & 35.8 &  {65.3} &  &  69.8 & 54.6 & 35.2 &  {65.4} &  & 69.8 & 54.6 & 35.0 &  {65.4}        \\
         Forward-LT-5 &    68.7 & 55.0 & 36.4 &  {63.0} &  &  68. & 55.0 & 36.4 &  {63.0} &  &     68.7 & 54.7 & 36.7 &  {62.9}    \\ 
        Forward-LT-2 & 68.1 & 56.0 & 39.9 &  {60.5} &  &  68.3 & 55.9 & 39.6 &  {60.5} &  &  68.2 & 56.0 & 40.1 &  {60.6}            \\
         Uniform    & 66.7 & 56.9 & 43.2 &  {58.8} &  &66.9 & 56.8 & 42.8 &  {58.8} &  &      66.5 & 56.8 & 43.2 &  {58.7}         \\
         Backward-LT-2&   65.4 & 57.1 & 44.9 &  {57.1} &  & 65.6 & 57.0 & 44.7 &  {57.1} &  &  64.9 & 57.0 & 45.6 &  {57.0}          \\ 
          Backward-LT-5 & 63.4 & 57.4 & 47.6 &  {55.5} &  &  62.7 & 57.4 & 48.3 &  {55.6} &  &   63.4 & 57.5 & 47.0 &  {55.4}          \\
         Backward-LT-10    & 60.7 & 57.5 & 49.4 &  {54.6} &  &   61.1 & 57.6 & 48.8 &  {54.4} &  &   60.6 & 57.6 & 49.1 &  {54.5}        \\ 
           Backward-LT-25 &    60.4 & 57.1 & 50.4 &  {53.9} &  & 60.4 & 57.0 & 50.3 &  {53.8} &  &  60.9 & 56.8 & 50.2 &  {53.7}         \\ 
          Backward-LT-50 & 60.9 & 56.1 & 51.1 &  {53.2} &  &   60.6 & 55.9 & 51.1 &  {53.2} &  &   60.8 & 56.1 & 51.2 &  {53.2}          \\ 
      
     \bottomrule
	\end{tabular}}
	 \end{threeparttable}
	 \end{center}   
\end{table*}

\clearpage
\subsection{Influences of Batch Size}
In previous results, we set the batch size of   test-time self-supervised aggregation   to 128   on all datasets. In this appendix, we further evaluate the influence of the  batch size on our strategy, where we adjust the batch size from 64 to 256. As shown in Table~\ref{table_test_time_bs_supp}, with different batch sizes, the learned expert weights by our method  keep nearly the same, which shows that our method is insensitive to the batch size. The corresponding performance on various test class distributions is reported in Table~\ref{table_batch_performance_supp}, where the performance is also nearly the same when using different batch sizes.

\begin{table*}[h]
	\caption{The influence of the batch size on the learned expert weights by test-time self-supervised aggregation on ImageNet-LT.}
	\label{table_test_time_bs_supp}   
 \begin{center}
 \begin{threeparttable} 
    \resizebox{0.95\textwidth}{!}{
 	\begin{tabular}{lcccccccccccc}\toprule 
 	 
        \multirow{2}{*}{Test Dist.}
        &&\multicolumn{3}{c}{Batch size 64} &&\multicolumn{3}{c}{Batch size 128}
        &&\multicolumn{3}{c}{Batch size 256}  \cr \cmidrule{3-5}  \cmidrule{7-9} \cmidrule{11-13} 
        && E1 ($w_1$)  & E2 ($w_2$) & E3  ($w_3$)  && E1 ($w_1$)  & E2 ($w_2$) & E3  ($w_3$) && E1 ($w_1$)  & E2 ($w_2$) & E3  ($w_3$)   \cr
        \midrule
        Forward-LT-50   &&  0.52 & 0.37 &0.11 &&  0.52	 	&  0.35	& 0.13	 &&  0.50 & 0.33 & 0.17  \\
        Forward-LT-25   && 0.49 &.0.38 &0.13 &&  0.50 & 0.35 & 0.15   && 0.48 & 0.24 & 0.18  \\
         Forward-LT-10  && 0.46&0.36&0.18&& 0.46	 	&  0.36	& 0.18   &&  0.45 & 0.35 & 0.20 \\
         Forward-LT-5   && 0.44 &0.34&0.22 &&  0.43 & 0.34 & 0.23	  && 0.43 & 0.35 & 0.22  \\  
         Forward-LT-2   && 0.37&0.34&0.29 && 0.37 & 0.35 & 0.28	 && 0.38 & 0.33 &0.29 \\
         Uniform    && 0.34 &0.32&0.34 && 0.33       & 0.33 	 &	 0.34 &&  0.33 &0.32 &0.35 \\
         Backward-LT-2  && 0.28 &.032&0.40 &&  0.29 & 0.31 & 0.40 && 0.30 & 0.31 &0.39 \\ 
         Backward-LT-5  && 0.24 & 0.30 &0.46 &&  0.24 & 0.31 & 0.45 && 0.25 & 0.30 & 0.45 \\
         
          Backward-LT-10  && 0.21&0.30 &0.49&&  0.21 	& 0.29	& 0.50  &&  0.22 & 0.29 &0.49  \\
          Backward-LT-25  && 0.17 & 0.29 & 0.54 && 0.18 & 0.29 & 0.53 && 0.20 &0.28&0.52   \\
         Backward-LT-50    && 0.15 & 0.30 & 0.55 && 0.17       &0.27	 &	0.56 && 0.19 &0.27&0.54 \\  
 
        \bottomrule

	\end{tabular}}
	 \end{threeparttable}   \end{center}  

\end{table*} 

  \begin{table*}[h]   
	\caption{The influence of the batch size on the performance of test-time self-supervised aggregation on ImageNet-LT.}
	\label{table_batch_performance_supp} 
 \begin{center}
 \begin{threeparttable} 
    \resizebox{0.95\textwidth}{!}{
 	\begin{tabular}{lcccccccccccccc}\toprule  
        \multirow{2}{*}{Test Dist.}  
       & \multicolumn{4}{c}{Batch size 64} &&\multicolumn{4}{c}{Batch size 128}
        &&\multicolumn{4}{c}{Batch size 256}  \cr\cmidrule{2-5} \cmidrule{7-10}\cmidrule{12-15}
        & Many  & Med. & Few & All &&  Many  & Med. & Few & All && Many  & Med. & Few & All \cr
        \midrule
         Forward-LT-50   & 70.0&52.6 & 33.8& {69.3}&&  70.0 	& 53.2	& 33.1 &  {69.4}   && 69.7 & 53.8 & 34.6 &  {69.2}    \\
         Forward-LT-25 & 69.6 & 53.0&33.3 & {67.5} &&69.5 & 53.2 & 32.2 &  {67.4} &&      69.2 & 53.7 & 32.8 &  {67.2}   \\ 
         Forward-LT-10  &  69.9& 54.3 & 34.8&  {65.4} && 69.9	 	& 54.3	& 34.7 	&  {65.4}  && 69.5 & 55.0 & 35.9 &  {65.3}  \\
         Forward-LT-5 & 69.0 & 54.6 & 35.6 &  {63.0} && 68.9 & 54.8  & 35.8 &  {63.0}  &&  68.8 & 54.9 & 36.0 &  {63.0}   \\ 
         Forward-LT-2 & 68.2 & 56.0 & 40.0&  {60.6} &&  68.2 & 56.0 & 40.1 &  {60.6} &&   68.1 & 56.0 & 40.1 &  {60.5}  \\
         Uniform    &  66.9 & 56.6 & 42.4 &  {58.8} && 66.5	 	& 57.0	& 43.5 	&  {58.8}  && 66.5 & 56.9 & 43.3 &  {58.8}     \\
         Backward-LT-2& 64.9 & 57.0 & 45.7 &  {57.0} && 65.3& 57.1 & 45.0 & {57.1} &&   65.5 & 57.1 & 44.8 &  {57.1}  \\
          Backward-LT-5 & 63.1 & 57.4 & 47.3 &  {55.4} && 63.4 & 56.4 &47.5 &  {55.5}  &&  63.4 & 56.4 & 47.5 &  {55.5}      \\
         Backward-LT-10    & 60.9 & 57.7 & 48.6 &  {54.4} && 60.9      & 57.5	 	& 50.1 &  {54.5}  && 61.3 & 57.6 & 48.7&  {54.4}     \\ 
           Backward-LT-25 & 60.8 & 56.7& 50.1&  {53.6} &&  60.5 & 57.1 & 50.0 &  {53.7} &&    61.0 & 57.2 & 49.6  &  {53.6}  \\
          Backward-LT-50 & 61.1 & 56.2 & 50.8 &  {53.1} && 60.7      & 56.2	 	&  50.7 &  {53.1}   && 61.2  & 56.4 & 50.0 &  {52.9} \\

     \bottomrule
	\end{tabular}}
	 \end{threeparttable}
	 \end{center}  
\end{table*}

 \newpage
\subsection{Influences of Learning Rate}
 In this appendix, we  evaluate the influence of the  learning rate on our self-supervised strategy, where we adjust the learning rate from 0.001 to 0.5. As shown in Table~\ref{table_test_time_lr_supp}, with the increase of the learning rate, the learned expert weights by our method are   sharper and fit  the unknown test class distributions better. For example, when the learning rate is 0.001, the weight for expert $E_1$ is 0.36 on the Forward-LT-50 test distribution, while when the learning rate increases to 0.5, the weight for expert $E_1$ becomes 0.57 on the Forward-LT-50 test distribution.  Similar phenomenons 
are also   observed on backward long-tailed test class distributions.

  By  observing the corresponding model performance on various test class distributions in Table~\ref{table_lr_performance_supp}, we find that when the learning rate is too small (\eg 0.001), our test-time self-supervised aggregation strategy is unable to  converge, given a fixed training epoch number of 5. In contrast, given the same training epoch, our method can obtain better performance by reasonably increasing the learning rate.

\begin{table*}[h]   
	\caption{The influence of the learning rate on the learned expert weights by test-time self-supervised aggregation on ImageNet-LT, where the number of the training epoch is 5.}
	\label{table_test_time_lr_supp} 
 \begin{center}
 \begin{threeparttable} 
    \resizebox{0.85\textwidth}{!}{
 	\begin{tabular}{lcccccccccccc}\toprule 
 	 
        \multirow{2}{*}{Test Dist.}
        &&\multicolumn{3}{c}{Learning rate 0.001} &&\multicolumn{3}{c}{Learning rate 0.01}
        &&\multicolumn{3}{c}{Learning rate 0.025}  \cr \cmidrule{3-5}  \cmidrule{7-9} \cmidrule{11-13} 
        && E1 ($w_1$)  & E2 ($w_2$) & E3  ($w_3$)  && E1 ($w_1$)  & E2 ($w_2$) & E3  ($w_3$) && E1 ($w_1$)  & E2 ($w_2$) & E3  ($w_3$)   \cr
        \midrule
        Forward-LT-50   &&  0.36&0.34 	& 0.30 &&  0.49	& 0.33	& 0.18	& 	&  0.52	 	&  0.35	& 0.13	     \\
        Forward-LT-25   &&  0.36	& 0.34	& 0.30 	& 	&  0.48	& 0.34	& 0.18	& 	&  0.50 & 0.35 & 0.15    \\
         Forward-LT-10  && 0.36	& 0.34	& 0.30	& 	&  0.45	& 0.34	& 0.21	& 	&    0.46	 	&  0.36	& 0.18   \\
         Forward-LT-5   &&  0.36	& 0.33	& 0.31	& 	&  0.43	& 0.34	& 0.23	& 	&  0.43 & 0.34 & 0.23	   \\   
         Uniform    && 0.33	& 0.33	& 0.34	& 	& 0.34	& 0.33	& 0.33	& 	&  0.33       & 0.33 	 &	 0.34  \\ 
         Backward-LT-5  &&  0.31	& 0.32	& 0.37	& 	& 0.25	& 0.31	& 0.44	& 	&  0.24 & 0.31 & 0.45  \\
          Backward-LT-10  &&  0.31	& 0.32	& 0.37	& 	&  0.22	& 0.29	& 0.49	& 	&  0.21 	& 0.29	& 0.50  \\
          Backward-LT-25  &&  0.31	& 0.32	& 0.37	& 	& 0.21	& 0.28	& 0.51	& 	&   0.18 & 0.29 & 0.53  \\
         Backward-LT-50    && 0.31	& 0.32.	& 0.37	& 	& 0.20	& 0.28	& 0.52	& 	&    0.17       &0.27	 &	0.56  \\  
    \midrule \midrule
        \multirow{2}{*}{Test Dist.}
        &&\multicolumn{3}{c}{Learning rate 0.05} &&\multicolumn{3}{c}{Learning rate 0.1}
        &&\multicolumn{3}{c}{Learning rate 0.5}  \cr \cmidrule{3-5}  \cmidrule{7-9} \cmidrule{11-13} 
        && E1 ($w_1$)  & E2 ($w_2$) & E3  ($w_3$)  && E1 ($w_1$)  & E2 ($w_2$) & E3  ($w_3$) && E1 ($w_1$)  & E2 ($w_2$) & E3  ($w_3$)   \cr
        \midrule
        Forward-LT-50   && 0.53& 0.36& 0.11& &   0.53 & 0.37& 0.10& &   0.57& 0.34 & 0.09  \\ 
        Forward-LT-25   &&  0.51& 0.36& 0.13& &    0.52& 0.36& 0.12& &  0.57& 0.34& 0.09  \\ 
         Forward-LT-10  &&   0.45& 0.37& 0.18& &   0.47& 0.36& 0.18& &  0.44& 0.36& 0.20  \\ 
         Forward-LT-5   &&  0.42& 0.35& 0.23& &    0.47& 0.36& 0.18 & & 0.39& 0.36& 0.25    \\  
         Uniform     &&   0.33& 0.33& 0.34& &    0.31& 0.31& 0.38& &  0.33& 0.34& 0.33 \\  
         Backward-LT-5  &&  0.24& 0.31& 0.45& &0.24& 0.29& 0.47& &    0.21& 0.28& 0.51      \\  
          Backward-LT-10  &&  0.21& 0.30& 0.49& &  0.21& 0.31& 0.48& &    0.22& 0.32& 0.46  \\ 
          Backward-LT-25  &&  0.16& 0.28& 0.56& &      0.17& 0.31& 0.52& & 0.15& 0.30& 0.55 \\ 
         Backward-LT-50    &&   0.15& 0.28& 0.57& &0.14& 0.28& 0.58& &     0.12& 0.27& 0.61  \\      
        \bottomrule

	\end{tabular}} 
	 \end{threeparttable} 
	 \end{center}  \vspace{0.1in}
\end{table*}

  \begin{table*}[h]    
	\caption{The influence of   learning rates on  test-time self-supervised aggregation on ImageNet-LT, under training epoch   5.}
	\label{table_lr_performance_supp}  
 \begin{center}
 \begin{threeparttable} 
    \resizebox{0.85\textwidth}{!}{
 	\begin{tabular}{lcccccccccccccc}\toprule  
        \multirow{2}{*}{Test Dist.}  
        &\multicolumn{4}{c}{Learning rate 0.001}&&\multicolumn{4}{c}{Learning rate 0.01} && \multicolumn{4}{c}{Learning rate 0.025} \cr\cmidrule{2-5} \cmidrule{7-10}\cmidrule{12-15}
        & Many  & Med. & Few & All &&  Many  & Med. & Few & All && Many  & Med. & Few & All \cr
        \midrule
      Forward-LT-50   & 67.3& 56.1& 44.1&  {67.3}& & 69.5 & 54.0 & 34.6 &  {69.0} &  &   70.0 	& 53.2	& 33.1 &  {69.4}              \\
         Forward-LT-25 & 67.4& 56.2& 40.3&  {66.1}& & 69.2 & 53.8 & 33.2 &  {67.2} &  &    69.5 & 53.2 & 32.2 &  {67.4}           \\
         Forward-LT-10  &  67.7& 56.4& 41.9&  {64.5}& & 69.6 & 55.0 & 36.1 &  {65.4} &  &     69.9	 	& 54.3	& 34.7 	&  {65.4}           \\
         Forward-LT-5 &  67.2& 55.9& 40.8&  {62.6}& & 68.7 & 55.0 & 36.2 &  {63.0} &  &   68.9 & 54.8  & 35.8 &  {63.0}           \\ 
         Uniform    & 66.9& 56.6& 42.7&  {58.8}& & 67.0 & 56.8 & 42.7 &  {58.8} &  &   66.5	 	& 57.0	& 43.5 	&  {58.8}            \\ 
          Backward-LT-5 & 65.8& 57.5& 43.7&  {55.0}& & 63.9 & 57.5 & 46.9 &  {55.5} &  &  63.4 & 56.4 &47.5 &  {55.5}           \\
         Backward-LT-10    & 64.6& 57.5& 43.7&  {53.1}& & 61.3 & 57.6 & 48.6 &  {54.4} &  &   60.9      & 57.5	 	& 50.1 &  {54.5}            \\
           Backward-LT-25 & 66.0& 57.3& 44.1&  {51.5}& &  61.1  & 57.4 & 49.3 &  {53.5} &  &   60.5 & 57.1 & 50.0 &  {53.7}           \\
          Backward-LT-50 & 68.2& 56.8& 43.7&  {50.0}& & 63.1 & 56.5 & 49.5 &  {52.7} &  &   60.7      & 56.2	 	&  50.7 &  {53.1}             \\

        \midrule   \midrule
          \multirow{2}{*}{Test Dist.}    &\multicolumn{4}{c}{Learning rate 0.05}&&\multicolumn{4}{c}{Learning rate 0.1} && \multicolumn{4}{c}{Learning rate 0.5} \cr\cmidrule{2-5} \cmidrule{7-10}\cmidrule{12-15}
        & Many  & Med. & Few & All &&  Many  & Med. & Few & All && Many  & Med. & Few & All \cr
        \midrule
         Forward-LT-50   &   70.2 & 52.4 & 32.4 &  {69.5} &  &  70.3 & 52.3 & 32.4 &  {69.5} &  &  70.3   & 51.2 & 32.4 &  {69.5}   \\
         Forward-LT-25 &  69.7 & 52.5 & 32.5 &  {67.5} &  &  69.9 & 52.3 & 31.4 &  {67.6} &  &  69.9 & 51.1 & 29.5 &  {67.5}       \\
         Forward-LT-10  &  69.7 & 54.7 & 35.8 &  {65.4} &  &  69.9 & 54.3 & 34.8 &  {65.4} &  &  69.5 & 55.0 & 35.8 &  {65.3}        \\
         Forward-LT-5 &  68.8 & 54.9 & 36.2 &  {63.0} &  &  68.8 & 54.8 & 36.1 &  {63.0} &  &   68.3 & 55.3 & 37.6 &  {62.9}          \\ 
         Uniform    & 66.6 & 56.9 & 43.2 &  {58.8} &  &  65.6 & 57.1 & 44.7 &  {58.7} &  &  67.8 & 56.4 & 40.9 &  {58.7}        \\ 
          Backward-LT-5 & 63.6 & 57.5 & 48.9 &  {55.4} &  &   63.0 & 57.4 & 48.1 &  {55.6} &  &  61.4 & 57.4 & 49.2 &  {55.6}       \\
         Backward-LT-10    & 61.1 & 57.5 & 48.9 &  {54.4} &  &  61.3 & 57.6 & 48.6 &  {54.4}   &  & 62.0 & 57.5 & 47.9 &  {54.2}     \\
           Backward-LT-25 & 59.9 & 56.8 & 51.0 &  {53.9} &  & 60.9 & 57.2 & 49.9 &  {53.7} &  &  60.2 & 56.8 & 50.8 &  {53.9}          \\
          Backward-LT-50 & 60.1 & 56.0 & 51.2 &  {53.2} &  &   59.6 & 55.8 & 51.3 &  {53.2} &  &    58.2 & 55.6 & 52.2 &  {53.5}     \\
      
     \bottomrule
	\end{tabular}}
	 \end{threeparttable}
	 \end{center}    
\end{table*}

 \clearpage 
\subsection{Results of Prediction Confidence}\label{App_F4}

In our theoretical analysis (\ie Theorem~\ref{thm1}), we find that our test-time self-supervised aggregation strategy   not only  simulates   the test class distribution, but also  makes  the model predictions more confident.   In this appendix, we evaluate whether our strategy can really improve  the prediction confidence of models on various unknown test class distributions of ImageNet-LT. To this end, we compare the prediction confidence of our method without  and  with test-time self-supervised aggregation in terms of the hard mean of the highest prediction probability on all test samples. 

As shown in   Table~\ref{table_confidence_supp}, our test-time self-supervised aggregation strategy enables the deep model to have higher prediction confidence. For example, on the Forward-LT-50 test distribution, our strategy obtains 0.015 confidence improvement, which is non-trivial since it is an average value for a large number of samples (more than 10,000 samples).  In addition,   when the class imbalance ratio becomes larger, our method is able to obtain more apparent confidence improvement.
  
\begin{table*}[h]  
	\caption{Comparison of prediction confidence between our method without  and  with test-time self-supervised aggregation on ImageNet-LT, in terms of the hard mean of the highest prediction probability on each sample. The higher the highest prediction, the better the model.}
	\label{table_confidence_supp}   
 \begin{center}
 \begin{threeparttable} 
    \resizebox{0.95\textwidth}{!}{
 	\begin{tabular}{lccccccccccccccc}\toprule  
   	     \multirow{4}{*}{Method} & \multicolumn{13}{c}{Prediction confidence  on ImageNet-LT} \cr \cmidrule{2-14} 
      & \multicolumn{5}{c}{Forward-LT} && Uniform  && \multicolumn{5}{c}{Backward-LT}   \cr  \cmidrule{2-6} \cmidrule{8-8}\cmidrule{10-14} 
         & 50 &25 &10&5&2 && 1 && 2& 5& 10 &25 &50  \cr  
        \midrule
        Ours w/o test-time   aggregation  &  0.694 & 0.687  & 	   0.678 & 0.665 & 0.651 & &0.639 & &0.627 &  0.608  & 0.596  & 0.583 & 0.574 \\  
        Ours w  test-time   aggregation    & 0.711   & 0.704  & 0.689  & 0.674  & 0.654  &   & 0.639  &   & 0.625  & 0.609  & 0.599  & 0.589  & 0.583  \\

        \bottomrule
    
	\end{tabular}}
	 \end{threeparttable}
	 \end{center}  
\end{table*}

\subsection{Run-time Cost  of Test-time Aggregation}

One may be interested in the run-time cost of our   test-time self-supervised aggregation strategy, so we further report its running time  on Forward-LT-50 and Forward-LT-25 test class distributions for illustration. As shown in Table~\ref{table_time_ssl}, our  test-time self-supervised aggregation strategy is   fast in terms of  per-epoch time. The actual average additional time is only 0.009 seconds per sample at test time on  V100 GPUs. 
The result    is   easy to interpret since we freeze the model parameters and only learn the aggregation weights, which  is much more efficient than     training  the whole model. More importantly,  the goal of this paper is to handle  a   practical  yet challenging   test-agnostic long-tailed recognition task.  For solving this  challenging  problem, we believe it is acceptable  to allow models to be trained more, while the promising results  in previous experiments have demonstrated   the effectiveness of our proposed test-time self-supervised  learning strategy in  handling this problem. In the future, we will further extend the proposed method for better computational efficiency, e.g., exploring dynamic network routing. 
 
 \begin{table}[h]  
  \caption{Run-time cost of our   test-time self-supervised  aggregation   strategy   on ImageNet-LT, compared to the run-time cost of model training. Here, we   show two test class distributions for illustration, which have different numbers of test samples.}  
  \label{table_time_ssl}  
 \begin{center} 
 \begin{threeparttable} 
    \resizebox{0.7\textwidth}{!}{
 	\begin{tabular}{cccc}\toprule  
        \multirow{2}{*}{Dataset} &    \multirow{2}{*}{Model training}  &   \multicolumn{2}{c}{Test-time  weight learning}    \cr \cmidrule{3-4} 
        & & Forward-LT-50 & Forward-LT-25    \cr  
        \midrule
        Per-epoch time 	 & 713 s & 110 s  & 130 s	\\  
        \bottomrule

	\end{tabular}} 
	 \end{threeparttable}
	 \end{center}   
\end{table}

\clearpage

\subsection{Test-time Self-supervised Aggregation on Streaming Test Data} 
In the previous experiments, we conduct the test-time self-supervised aggregation strategy in an offline manner. However, as mentioned in Section~\ref{Test_Time}, our test-time strategy can also be conducted in an online manner and does not require access to all the test data in advance. To verify this, we further conduct our test-time strategy on steaming test data of ImageNet-LT.
As shown in Table~\ref{online_testime}, our test-time strategy performs well on the streaming test data. Even when the test data come in one by one, our test-time self-supervised strategy still outperforms the state-of-the-art baseline (i.e., offline Tent~\cite{wang2021tent})  by a large margin.   
 
\begin{table}[H]   
	\caption{Results of our test-time self-supervised aggregation strategy on streaming test data of ImageNet-LT, where all test-time strategies are used on the same  skill-diverse multi-expert model.  } 
	\label{online_testime} 
 \begin{center}
 \begin{threeparttable} 
    \resizebox{0.99\textwidth}{!}{
	\begin{tabular}{lcccccc}\toprule  
   	     \multirow{2}{*}{Backbone} &\multirow{2}{*}{Test-time strategy}&   \multicolumn{2}{c}{Forward-LT} && \multicolumn{2}{c}{Backward-LT} \cr  \cmidrule{3-4} \cmidrule{6-7} 
         && 50 &5  &&   5 &50\cr  
        \midrule
         \multirow{6}{*}{SADE}   &  No test-time adaptation & 65.5 &  62.0   && 54.7  &49.8  \\        
        & Offline Tent~\cite{wang2021tent} & 68.0    & 62.8   & &    53.2   &  45.7 \\    
     
          & Offline self-supervised aggregation (ours) & {69.4} &  {63.0}  &&  {55.5} &    {53.1} \\    
           & Online self-supervised aggregation with batch size 64  & {69.5} &  {63.6}   &&  {55.8}  & {53.1} \\      
          & Online self-supervised aggregation with batch size 8  & {69.8} &  {63.0}   &&  {55.4}  & {53.0} \\   
           & Online self-supervised aggregation with batch size 1  & {69.0} &  {62.8}   &&  {55.2}  & {52.8} \\   
 
        \bottomrule
    
	\end{tabular}}
	 \end{threeparttable}
	 \end{center}  
\end{table}

\clearpage

\section{More Discussions on Model Complexity}\label{App_G}

In this appendix, we  discuss the model complexity of our method in terms of the number of parameters, multiply-accumulate operations (MACs) and    top-1 accuracy on test-agnostic long-tailed recognition. 
As shown in Table~\ref{table_complexity_supp}, 
both SADE and RIDE belong to ensemble-based long-tailed learning methods, so they have more parameters (about 1.5x) and MACs (about 1.4x) than the original backbone model, where we do not use the efficient expert assignment trick in \cite{wang2020long} for both methods. Because of the ensemble effectiveness of the multi-expert scheme, both methods perform much better than non-ensemble methods (\eg Softmax and other long-tailed methods). 
In addition, since our method and RIDE use the same multi-expert framework, both  methods have the same number of parameters and MACs. Nevertheless, by using our proposed   skill-diverse expert learning and test-time self-supervised aggregation strategies, our method performs much better than RIDE with no increase in model parameters and computational costs.  

One may concern   the multi-expert scheme leads to more model parameters and higher  computational costs than the original backbone. However,  note that the main focus of this paper is to solve the challenging  test-agnostic long-tailed recognition, while   promising results have shown that our method addresses this problem well. In this sense, slightly increasing the model complexity is acceptable  for solving this practical yet challenging  problem. Moreover, since there have already been many  studies~\cite{havasi2020training,wen2019batchensemble} showing effectiveness in improving the efficiency of the multi-expert  scheme, we think the computation increment  is not a   severe issue and we leave it to the  future. 
 
\begin{table}[h]  
	\caption{Model complexity and performance of different methods in terms of the parameter number, Multiply–Accumulate Operations (MACs) and top-1 accuracy on test-agnostic long-tailed recognition. Here, we do not use the efficient expert assignment trick in \cite{wang2020long} for RIDE and our method.}\label{table_complexity_supp}  \vspace{-0.1in}
 \begin{center}
 \begin{threeparttable} 
    \resizebox{0.88\textwidth}{!}{
 	\begin{tabular}{lccccccccccccccccccc}\toprule  
   	     \multirow{4}{*}{Method} &  \multirow{4}{*}{Params (M)} & \multirow{4}{*}{MACs (G)} & &\multicolumn{13}{c}{ImageNet-LT (\textbf{ResNeXt-50)}} \cr \cmidrule{5-17} 
      &&&& \multicolumn{5}{c}{Forward-LT} && Uniform  && \multicolumn{5}{c}{Backward-LT}   \cr  \cmidrule{5-9} \cmidrule{11-11}\cmidrule{13-17} 
        &&& & 50 &25 &10&5&2 && 1 && 2& 5& 10 &25 &50  \cr  
        \midrule
        Softmax  & 25.03 (1.0x) & 4.26 (1.0x)&&  66.1 & 63.8 & 	  60.3& 56.6&52.0 && 48.0 && 43.9 & 38.6 & 34.9 & 30.9 & 27.6 \\  
        RIDE~\cite{wang2020long}       &38.28 (1.5x) &  6.08 (1.4x) & & 67.6 & 66.3      & 64.0 & 61.7 & 58.9 && 56.3 && 54.0 & 51.0 & 48.7 & 46.2 & 44.0  \\ 
        SADE (ours)   &38.28 (1.5x) &  6.08 (1.4x)  &&  \textbf{69.4} &  \textbf{67.4}    & \textbf{65.4} 	& \textbf{63.0}  & \textbf{60.6} && \textbf{58.8} && \textbf{57.1} & \textbf{55.5} & \textbf{54.5} & \textbf{53.7} & \textbf{53.1}   \\ 
        \midrule  \midrule 
 
   	     \multirow{4}{*}{Method} &  \multirow{4}{*}{Params (M)} & \multirow{4}{*}{MACs (G)} & &\multicolumn{13}{c}{CIFAR100-LT-IR100  (\textbf{ResNet-32)}} \cr \cmidrule{5-17} 
      &&&& \multicolumn{5}{c}{Forward-LT} && Uniform  && \multicolumn{5}{c}{Backward-LT}   \cr  \cmidrule{5-9} \cmidrule{11-11}\cmidrule{13-17} 
        &&& & 50 &25 &10&5&2 && 1 && 2& 5& 10 &25 &50  \cr  
        \midrule
        Softmax  & 0.46 (1.0x) & 0.07 (1.0x)&&  63.3&62.0&56.2 & 52.5 & 46.4 && 41.4 && 36.5 & 30.5 & 25.8 & 21.7 & 17.5  \\ 
        RIDE~\cite{wang2020long}     & 0.77 (1.5x) &  0.10 (1.4x) & &63.0 & 59.9 & 57.0 & 53.6 & 49.4 && 48.0 && 42.5 & 38.1 & 35.4 & 31.6 & 29.2  \\
        SADE (ours)   & 0.77 (1.5x) &  0.10 (1.4x)  &&  \textbf{65.9} & \textbf{62.5} & \textbf{58.3} & \textbf{54.8} & \textbf{51.1} && \textbf{49.8} && \textbf{46.2} & \textbf{44.7} & \textbf{43.9} & \textbf{42.5} & \textbf{42.4}\\   
        \midrule  \midrule 
        
	     \multirow{4}{*}{Method} &  \multirow{4}{*}{Params (M)} & \multirow{4}{*}{MACs (G)} & &\multicolumn{13}{c}{Places-LT  (\textbf{ResNet-152)}} \cr \cmidrule{5-17} 
      &&&& \multicolumn{5}{c}{Forward-LT} && Uniform  && \multicolumn{5}{c}{Backward-LT}   \cr  \cmidrule{5-9} \cmidrule{11-11}\cmidrule{13-17} 
        &&& & 50 &25 &10&5&2 && 1 && 2& 5& 10 &25 &50  \cr  
        \midrule
        Softmax  & 60.19 (1.0x) &11.56 (1.0x)&& 45.6 & 42.7 & 40.2 & 38.0 & 34.1 &&  31.4 && 28.4 & 25.4 & 23.4 & 20.8 & 19.4  \\ 
        RIDE~\cite{wang2020long}     & 88.07 (1.5x) &  13.18 (1.1x) & &43.1 & 41.8 & 41.6 & 42.0 & 41.0 && 40.3 && 39.6 & 38.7 & 38.2 & 37.0 & 36.9\\
        SADE (ours)   & 88.07 (1.5x) &  13.18 (1.1x)  &&  \textbf{46.4} & \textbf{44.9} & \textbf{43.3} & \textbf{42.6} & \textbf{41.3} & & \textbf{40.9} && \textbf{40.6} & \textbf{41.1} & \textbf{41.4} & \textbf{42.0} & \textbf{41.6}\\   
        \midrule  \midrule

	     \multirow{4}{*}{Method} &  \multirow{4}{*}{Params (M)} & \multirow{4}{*}{MACs (G)} & &\multicolumn{13}{c}{iNaturalist 2018  (\textbf{ResNet-50)}} \cr \cmidrule{5-17} 
      &&&& \multicolumn{5}{c}{Forward-LT} && Uniform  && \multicolumn{5}{c}{Backward-LT}   \cr  \cmidrule{5-9} \cmidrule{11-11}\cmidrule{13-17} 
        &&& &  &3 & &2&  && 1 &&  &2&   &3 &  \cr  
        \midrule
        Softmax  & 25.56 (1.0x) & 4.14 (1.0x)&&  &65.4 &&    65.5 &&&64.7 &&&64.0 &&63.4 &   \\ 
        RIDE~\cite{wang2020long}     & 39.07 (1.5x) &  5.80 (1.4x) & && 71.5 &&   71.9 &&& 71.8 &&& 71.9 && 71.8  &\\
        SADE (ours)   & 39.07 (1.5x) &  5.80 (1.4x) &&   &\textbf{72.3}  &&  \textbf{72.5} &&& \textbf{72.9}  &&& \textbf{73.5} && \textbf{73.3} & \\   
        
        \bottomrule
    
	\end{tabular}}
	 \end{threeparttable} 
	 \end{center} 
 	 \vspace{-0.15in}
	
\end{table}

\section{Potential Limitations}\label{App_H} 

One concern  is that this work only focuses on long-tailed classification problems. However, we believe this is enough for a new challenging task of test-agnostic long-tailed recognition, while how to extending to   object detection and instance segmentation will be explored in the future. 
Another potential concern is the model complexity of our method. However,   as discussed in Appendix~\ref{App_G},   the computation increment  is not a very severe issue, while  how to further accelerate our method will be explored   in   future. In addition, one may also expect to evaluate   the proposed method    on more test class distributions. However, as shown in Section~\ref{sec_5.5}, we have demonstrated the effectiveness of our method on the uniform class distribution, the forward and backward long-tailed class distributions with various imbalance ratios, and even partial   class distributions. Therefore, we believe   the  empirical verification  is sufficient for verifying   our method, and the  extension to   more complex test class distributions is left to the future.
\end{document}